\documentclass[pdflatex,sn-mathphys-num]{sn-jnl}

\usepackage{graphicx}%
\usepackage{multirow}%
\usepackage{amsmath,amssymb,amsfonts}%
\usepackage{amsthm}%
\usepackage{mathrsfs}%
\usepackage[title]{appendix}%
\usepackage{xcolor}%
\usepackage{textcomp}%
\usepackage{manyfoot}%
\usepackage{booktabs}%
\usepackage{algorithm}%
\usepackage{algorithmicx}%
\usepackage{algpseudocode}%
\usepackage{listings}%
\usepackage{array}
\usepackage{makecell}

\theoremstyle{thmstyleone}%
\theoremstyle{thmstyletwo}%

\theoremstyle{thmstylethree}%

\begin{document}

\title[Article Title]{A Reproducible Log-Driven AutoML Framework for Interpretable Pipeline Optimization in Healthcare Risk Prediction}


\author[1]{\fnm{Rui} \sur{Huang}}\email{ruihuang@znu.edu.cn}

\author*[2]{\fnm{Lican} \sur{Huang}}\email{ican.huang.hz@gmail.com}

\affil[1]{\orgdiv{School of Basic Medicine}, \orgname{Hangzhou Normal University}, \orgaddress{\street{No.2318, Yuhangtang Rd, Yuhang District}, \city{Hangzhou}, \postcode{311121}, \state{Zhejiang}, \country{China}}}

\affil[2]{\orgdiv{Research Department}, \orgname{Hangzhou Domain Zones Technology Co.Ltd.}, \orgaddress{\street{200 Zhenhua Rd, West District}, \city{Hangzhou}, \postcode{310030}, \state{Zhejiang}, \country{China}}}


\abstract{
 
 Accurate disease risk prediction is challenged by heterogeneous features, limited data, and class imbalance. This study presents yvsoucom-iterkit, a deterministic AutoML framework that models pipeline optimization as a configuration-level system with full reproducibility and traceable execution logs, enabling systematic analysis of component attribution, interactions, similarity, and cross-seed robustness.

Experiments on the Pima Indians Diabetes and Stroke datasets across more than 18,000 pipeline configurations reveal a structured yet partially redundant search space, where performance is dominated by a small subset of interacting components.

Ensemble models achieve stable performance, reaching a Weighted-F1 of 0.89 on Pima and 0.94 on Stroke. Macro-F1 reaches approximately  0.88 on Pima but drops to 0.6560 on Stroke due to severe imbalance. Cross-seed experiments show that ensembles reduce variance compared to single models. Friedman testing ($p < 0.05$) confirms significant ranking differences across configurations.

Based on analysis of component attribution, interaction, and similarity, optimal configuration design reveals dataset-dependent behavior. For the Pima dataset, computational efficiency benefits from simplified search spaces where redundant components can be removed, with split ratio playing a key role. In contrast, the Stroke dataset requires enhanced imbalance-aware strategies, where RandomOverSampler improves Macro-F1 from 0.6560 to 0.6766.

These findings demonstrate that effective AutoML optimization is achieved through optimal configuration design, where carefully constraining the search space to high-impact components can improve performance, stability, and interpretability while reducing unnecessary search complexity.
 }




\maketitle

 \section{Introduction}

Chronic diseases such as diabetes mellitus and cerebrovascular disorders represent major global health challenges, affecting hundreds of millions of individuals worldwide. Early identification of high-risk individuals is critical for enabling timely intervention and reducing severe complications, including cardiovascular and neurological conditions. With the increasing availability of structured clinical data, machine learning (ML) techniques have become essential tools for healthcare risk prediction, enabling the discovery of complex patterns from heterogeneous clinical attributes \cite{HaibeKains2020}.

Despite significant progress, developing reliable ML-based healthcare prediction models remains challenging. Clinical datasets are often small, imbalanced, and heterogeneous, and model performance is highly sensitive to preprocessing decisions, including feature selection, normalization, and data augmentation. Many existing studies rely on manually designed or weakly optimized pipelines, limiting reproducibility and generalizability across datasets and clinical settings.

Automated machine learning (AutoML) has emerged as a promising solution by enabling data-driven exploration of model architectures and hyperparameters. However, most existing AutoML frameworks, such as Auto-sklearn, TPOT, and AutoGluon, primarily focus on model selection and hyperparameter optimization, while treating preprocessing and data transformation steps as fixed or only loosely optimized components. As a result, complex interactions between preprocessing strategies and learning algorithms remain underexplored, particularly in small and imbalanced healthcare datasets where such interactions are critical.

To address these limitations, we propose \texttt{yvsoucom-iterkit}, a pipeline-centric AutoML framework that jointly optimizes preprocessing and modeling within a unified search space. The framework treats feature selection, normalization, data augmentation, class-imbalance handling, and classification models as first-class components, enabling systematic exploration of their combinations. It operates as a fully automated and reproducible experimentation engine, capable of generating, executing, and evaluating thousands of pipeline configurations without human intervention.

Compared with our previous work \cite{huang2026_ssrn}, \cite{huang2026automl} significantly extends the scope and rigor of the proposed system by incorporating additional datasets, expanding the pipeline search space, and introducing comprehensive cross-seed evaluation to assess robustness. Furthermore, the framework integrates automated logging and statistical analysis modules, enabling transparent and reproducible benchmarking across large-scale experiments. Importantly, this study corrects a normalization data leakage issue present in the previous version, ensuring that all preprocessing steps are properly confined to the training set and do not contaminate test evaluations. In addition, the current work introduces more models and further optimizes the pipeline based on detailed component-level analyses, leading to improved stroke prediction performance, particularly for Macro-F1, through targeted experimental exploration guided by these analyses.

To evaluate the effectiveness and generality of the proposed system, we conduct extensive experiments on two representative healthcare prediction tasks: diabetes risk prediction using the Pima Indians Diabetes dataset and stroke risk prediction using the Healthcare Stroke Dataset. These datasets differ substantially in feature composition, class distribution, and clinical context, providing a rigorous testbed for assessing robustness and transferability. Experimental results demonstrate that the proposed framework consistently identifies high-performing and stable pipeline configurations, highlighting the importance of joint optimization across preprocessing and modeling stages.

Although this study focuses on two representative healthcare tasks, the proposed framework is designed as a general-purpose AutoML system for structured data. It supports flexible combinations of preprocessing and modeling strategies and can be readily applied to other domains involving categorical prediction tasks.

  The main contributions and findings of this work are summarized as follows:

\begin{itemize}

\item We propose a pipeline-centric AutoML framework that jointly optimizes preprocessing, data augmentation, imbalance handling, and classification models within a unified and fully reproducible configuration space for healthcare data analysis.

\item We introduce a log-driven (LogDir) execution paradigm that enables transparent and traceable large-scale experimentation by linking each pipeline configuration to its corresponding performance outcomes.

\item We conduct a large-scale evaluation over 18{,}000+ pipeline configurations on the Pima Indians Diabetes and Stroke datasets, enabling multi-level analysis ranging from raw metric distributions to pipeline-level behavior, component-level attribution, cross-component interactions, and cross-seed robustness.

\item We demonstrate that the AutoML search space is highly structured and partially redundant, where many configurations yield near-equivalent performance and only a small subset of components drives most performance variation.

\item We identify strong dataset-dependent behavior and interaction effects across components. In particular, Pima exhibits a relatively stable performance landscape where simplified search spaces and redundancy removal improve computational efficiency without degrading performance, with split ratio playing a key role. In contrast, Stroke shows higher sensitivity due to severe class imbalance, where imbalance-aware strategies such as RandomOverSampler (ROS) significantly improve Macro-F1 from 0.6560 to 0.6766. These effects are further consistent with observations that ensemble models achieve the best balance between predictive performance and robustness, while high-capacity models such as SVM are more sensitive to stochastic variation across random seeds.

\end{itemize}

The remainder of this paper is organized as follows. Section~\ref{related works} reviews related work on healthcare prediction and AutoML. Section~\ref{sec:Methodology} presents the proposed methodology  and framework.   Section~\ref{sec:experiment} reports experimental results, followed by discussion and conclusions in Sections~\ref{sec:Discussion}  and~\ref{sec:Conclusion}.

 \section{Related Work}
\label{related works}

\subsection{Automated Machine Learning}

Automated Machine Learning (AutoML) has evolved into a central paradigm for reducing manual effort in model development by automating algorithm selection, hyperparameter tuning, and, to a limited extent, pipeline construction~\cite{hutter2019automated, HE2021106622}. Early systems such as Auto-WEKA~\cite{Thornton2013} formulated the Combined Algorithm Selection and Hyperparameter (CASH) problem and addressed it using Bayesian optimization. Subsequent approaches, including TPOT~\cite{Olson2016}, introduced evolutionary strategies to explore pipeline configurations, while Auto-sklearn~\cite{Feurer2015, Feurer2022} incorporated meta-learning and ensemble construction to improve efficiency and robustness. More recent frameworks such as AutoGluon~\cite{Erickson2020AutoGluon} and H2O AutoML~\cite{LeDell2020H2O} further emphasize scalability and ensemble learning in large-scale applications.

Despite these advances, existing AutoML systems largely retain a \emph{model-centric optimization paradigm}. Preprocessing operations—such as feature selection, normalization, and data augmentation—are typically treated as auxiliary or weakly parameterized components rather than first-class elements of the search space~\cite{Zoller2021Benchmarking}. Even when pipeline structures are explored (e.g., TPOT), the search process is often stochastic and opaque, limiting interpretability and reproducibility. Furthermore, most frameworks prioritize predictive performance as the primary objective, with limited consideration of robustness, variability, and statistical reliability across runs~\cite{Bischl2023Hyperparameter}.

These limitations are particularly restrictive in healthcare settings, where model behavior is highly sensitive to preprocessing choices and data characteristics. As a result, existing AutoML approaches provide only partial automation and insufficient transparency for systematic pipeline-level analysis.

\subsection{Preprocessing and Imbalanced Learning in Healthcare Data}

Preprocessing is a critical determinant of model performance in clinical machine learning due to the inherent characteristics of healthcare datasets, including small sample sizes, heterogeneous feature types, and severe class imbalance. Feature selection methods, particularly those based on information theory such as information gain and mutual information, have been widely adopted to improve generalization and reduce overfitting~\cite{Guyon2006, Chandrashekar2014, li2017feature}. Wrapper and embedded methods further extend these approaches but often introduce additional computational complexity and instability~\cite{Saeys2007Feature}.

Class imbalance further complicates predictive modeling. Techniques such as SMOTE~\cite{Chawla2002SMOTE}, Tomek Links~\cite{Tomek1976}, and hybrid resampling methods (e.g., SMOTEENN~\cite{Batista2004SMOTEENN}) aim to rebalance class distributions. More recent approaches, including ADASYN~\cite{He2008ADASYN} and MixUp~\cite{zhang2018mixup}, introduce adaptive or interpolation-based augmentation to improve generalization. Cost-sensitive learning and focal loss mechanisms have also been proposed to address imbalance at the algorithmic level~\cite{Lin2017FocalLoss}.

Although these methods have demonstrated effectiveness in specific scenarios, they are typically evaluated in isolation or within manually constructed pipelines. Consequently, their combined effects, interactions, and trade-offs remain insufficiently understood. Moreover, prior studies rarely investigate how preprocessing decisions influence not only predictive accuracy but also \emph{stability under stochastic variation}, which is critical for reliable deployment in clinical environments.

\subsection{Machine Learning for Healthcare Risk Prediction}

Machine learning techniques have been extensively applied to healthcare risk prediction tasks, including diabetes diagnosis and stroke prediction. Ensemble methods such as Random Forest~\cite{Breiman2001} and XGBoost~\cite{chen2016xgboost} are frequently reported as strong performers on structured clinical data due to their ability to model nonlinear relationships and handle feature interactions~\cite{sarwar2020diagnosis}. Gradient boosting frameworks have also demonstrated superior performance in tabular domains compared to deep learning approaches~\cite{Grinsztajn2022Tabular}.

Numerous studies on the Pima Indians Diabetes dataset and stroke prediction benchmarks report improved performance through tailored combinations of preprocessing and model tuning~\cite{pati2023review, patil2013performance, Singh2025StrokeML}. Deep learning approaches, including multilayer perceptrons and hybrid architectures, have also been explored, but often require extensive tuning and larger datasets to achieve competitive performance~\cite{Shickel2018DeepEHR}.

However, these approaches are typically dataset-specific and rely on manual or heuristic design choices. As a result, their conclusions are often difficult to generalize, and reproducibility is limited due to incomplete reporting of experimental configurations~\cite{Pineau2021ImprovingReproducibility}. More importantly, existing studies predominantly focus on optimizing single pipelines or a small set of configurations, rather than systematically exploring the broader pipeline space. This restricts the ability to identify generalizable patterns, quantify variability, or understand the relative importance of different pipeline components.

\subsection{Research Gap and Contribution}

The above analysis reveals a fundamental gap in current research: the absence of a unified, reproducible, and systematically analyzable framework for \emph{pipeline-centric} AutoML in structured healthcare data.

Existing AutoML systems emphasize model-level optimization but lack transparent and deterministic exploration of preprocessing–model interactions. Conversely, studies on preprocessing and healthcare prediction provide valuable insights into individual techniques but fail to integrate them into a scalable and reproducible experimental framework. In addition, robustness and stochastic sensitivity are rarely incorporated into evaluation protocols, despite their importance in real-world clinical deployment.

To address these limitations, we propose \texttt{yvsoucom-iterkit}, a pipeline-centric AutoML framework that explicitly models preprocessing operations, augmentation strategies, imbalance handling, and classification algorithms within a unified and fully enumerable search space. Unlike prior approaches, the proposed system adopts a \emph{log-driven and deterministic execution paradigm}, enabling:

\begin{itemize}
    \item systematic and exhaustive exploration of thousands of pipeline configurations,
    \item full reproducibility through structured logging of all experimental details,
    \item component-level and interaction-level statistical analysis,
    \item and robustness evaluation through multi-seed experimentation.
\end{itemize}

This design shifts AutoML from a black-box optimization problem to a transparent and analyzable experimental process, enabling deeper understanding of how pipeline components jointly influence performance, robustness, and generalization. In particular, by integrating multi-seed evaluation and statistical aggregation, the framework provides an implicit bridge between empirical AutoML practice and theoretical considerations such as bias--variance trade-offs and stochastic sensitivity, which are largely absent from existing pipeline optimization frameworks.

\section{Methodology and Framework}
\label{sec:Methodology}

This section presents the proposed pipeline-centric AutoML framework, integrating dataset design, pipeline modeling, system architecture, and automated experimentation into a unified methodology. The framework systematically explores a combinatorial space of pipeline configurations through deterministic enumeration, enabling large-scale, reproducible, and interaction-aware benchmarking for structured healthcare data.

 \subsection{Datasets}

Experiments are conducted on two publicly available datasets representing heterogeneous clinical settings and feature characteristics:

\textbf{Pima Indians Diabetes Dataset.}
This dataset contains 768 instances with eight clinical features (e.g., glucose, blood pressure, BMI, age) and a binary diabetes outcome. All features are numerical, and the dataset exhibits moderate class imbalance.

\textbf{Healthcare Stroke Dataset.}
This dataset includes both numerical and categorical variables, such as age, average glucose level, BMI, hypertension status, heart disease, and smoking status, with a binary stroke label. Compared to the diabetes dataset, it presents higher class imbalance and increased feature heterogeneity due to mixed data types.

These datasets serve as complementary case studies, enabling evaluation across purely numerical and mixed-type feature spaces, thereby assessing the robustness and transferability of the proposed AutoML framework under varying data characteristics.

\subsection{Pipeline Design and Search Space}

The proposed framework defines a unified pipeline configuration space:

\begin{equation}
\mathcal{P} = F \times K \times N \times O \times A \times B \times M \times S \times T \times R
\end{equation}

where each dimension corresponds to a configurable component:
\begin{itemize}
\item $F$ — feature selection strategy  
\item $K$ — number of selected features  
\item $N$ — normalization method  
\item $O$ — normalization order (before or after augmentation and imbalance handling)  
\item $A$ — data augmentation method  
\item $B$ — imbalance handling method  
\item $M$ — classification model  
\item $S$ — train/test split ratio  
\item $T$ — decision threshold  
\item $R$ — random seed  
\end{itemize}
 
Each configuration $\mathcal{P}$ defines a complete machine learning pipeline, where preprocessing, modeling, and experimental factors are jointly specified. The framework captures interactions among these components within a unified combinatorial search space.

Unlike conventional AutoML approaches that rely on stochastic or heuristic search strategies, the proposed framework adopts deterministic enumeration of $\mathcal{P}$. This enables systematic and exhaustive (or controlled) exploration of the configuration space, ensuring full transparency, reproducibility, and comparability across pipeline variants.

By explicitly integrating preprocessing operations, execution order, model selection, and experimental parameters into a unified formulation, the proposed search space substantially extends conventional AutoML designs. This unified representation provides a structured and interpretable foundation for interaction-aware analysis of pipeline behavior, enabling systematic investigation of how individual components jointly influence performance.

Consequently, the proposed formulation shifts the focus of AutoML from isolated model-centric optimization toward holistic, pipeline-centric system exploration, where performance is understood as an emergent property of coordinated component interactions rather than single-stage decisions.

\subsection{Framework Architecture and Automated Execution}

\begin{figure*}[t]
\centering
\includegraphics[width=\textwidth]{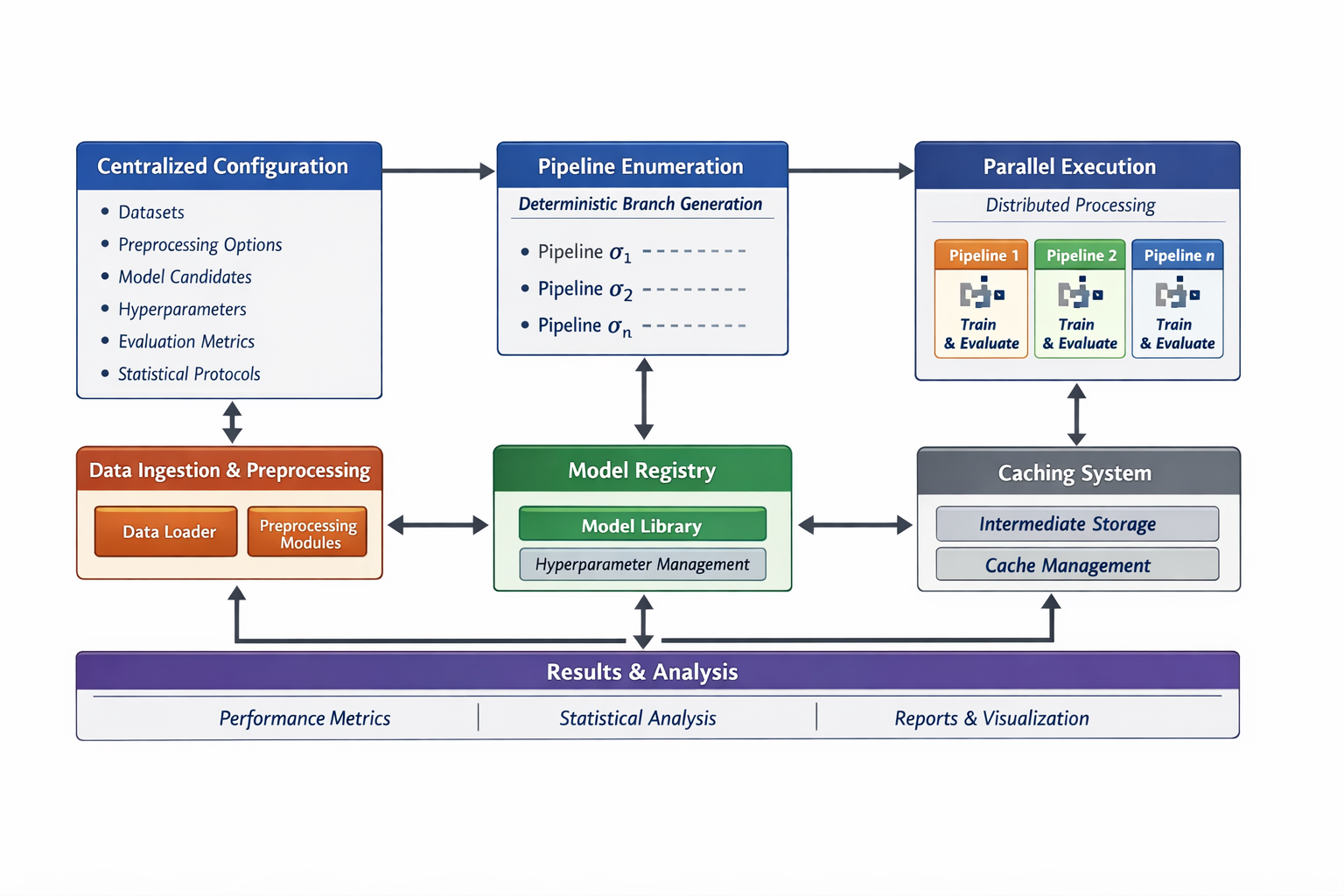}
\caption{System architecture of the proposed AutoML framework.
The framework follows a centralized configuration and deterministic
pipeline enumeration strategy, enabling parallel execution and
reproducible large-scale experimentation.}
\label{fig:framework_architecture}
\end{figure*}

As illustrated in Fig.~\ref{fig:framework_architecture}, the proposed AutoML framework is a modular and extensible system for pipeline-centric optimization on structured healthcare datasets. It adopts a deterministic, pre-enumerated execution model that enables systematic exploration of preprocessing, modeling, and experimental configurations within a unified search space.

The framework consists of five core components: (i) data ingestion, (ii) preprocessing and pipeline optimization, (iii) model management, (iv) experiment execution, and (v) result aggregation and analysis. These components interact through standardized interfaces and are coordinated by a centralized configuration module, ensuring consistency across all experimental stages.

Each experimental configuration is represented as an independent pipeline instance generated prior to execution. This flat execution structure eliminates recursive dependencies and ensures that all configurations are explicitly defined, directly comparable, and fully reproducible.

Deterministic enumeration enables complete or controlled coverage of the pipeline search space, in contrast to stochastic AutoML approaches based on heuristic or probabilistic search. Combined with parallel execution and structured logging, this design supports large-scale experimentation with strong guarantees of traceability, consistency, and reproducibility.

\paragraph{Centralized Configuration and Branch Enumeration.}
A centralized configuration defines the full experimental design, including datasets, preprocessing options, pipeline components, model candidates, hyperparameter ranges, evaluation metrics, and statistical protocols. Based on this configuration, the framework deterministically enumerates all valid pipeline configurations (referred to as \emph{branches}), each representing a unique combination of preprocessing strategies, model choices, and experimental parameters.

\paragraph{Model Integration and Registry.}
Models are integrated through a unified interface that standardizes training, prediction, and evaluation. A centralized registry maintains metadata for each model, including supported input formats, preprocessing compatibility, and configurable hyperparameters.  

\paragraph{Parallel Execution and Caching.}
Each pipeline configuration is executed independently and can be dispatched in parallel. Execution order does not affect correctness due to pre-defined branch generation. To improve efficiency, the framework supports caching of intermediate results (e.g., preprocessed data), indexed using deterministic configuration signatures for consistent reuse.

\paragraph{Logging, Output, and Reproducibility.}

All experiments are executed within the \texttt{yvsoucom-iterkit} system, which automates pipeline generation, execution, and evaluation under a unified workflow. Controlled random seeds ensure deterministic behavior across runs. A structured logging mechanism records complete experimental metadata, including pipeline configurations, preprocessing steps, model parameters, intermediate outputs, raw predictions, and evaluation metrics. Each execution is associated with a unique branch identifier, enabling full traceability and systematic aggregation of results.

The log-centric design preserves complete execution histories, supporting reproducible analysis and independent verification of pipeline behavior. By maintaining access to both intermediate and final outputs, the framework enables fine-grained inspection of model performance beyond aggregate metrics.

 \subsection{Preprocessing and Pipeline Components}

The proposed framework treats preprocessing operations as first-class components within the optimization space, enabling their systematic integration and joint evaluation with model selection. The following preprocessing strategies are considered:

\textbf{Missing Value Handling.}
Implausible values (e.g., zero entries in physiological measurements) are treated as missing and imputed using mean substitution:
\begin{equation}
x_i =
\begin{cases}
\mu, & \text{if } x_i \text{ is missing or invalid} \\
x_i, & \text{otherwise}
\end{cases}
\end{equation}

\textbf{Feature Selection.}
Features are ranked using Information Gain (IG):
\begin{equation}
IG(Y, A) = H(Y) - H(Y \mid A)
\end{equation}
The top-$k$ features are selected based on ranking. To capture different relevance criteria, three variants are considered:
(i) standard IG,  
(ii) maximum binary IG, and  
(iii) mean binary IG.

\textbf{Normalization.}
Feature scaling is performed using two standard strategies:

\begin{align}
x' &= \frac{x - \mu}{\sigma} \quad (\text{StandardScaler}) \\
x' &= \frac{x - x_{\min}}{x_{\max} - x_{\min}} \quad (\text{MinMaxScaler})
\end{align}

These methods are evaluated in conjunction with different execution orders within the pipeline.

\textbf{Data Augmentation.}
To mitigate limited sample sizes and improve generalization, augmentation is applied exclusively to the training data. The framework supports multiple strategies, including CTGAN-based synthetic generation, Gaussian noise injection, and MixUp interpolation, enabling systematic comparison of generative and perturbation-based approaches.

\textbf{Imbalance Handling.}
To address class imbalance, the framework incorporates a diverse set of resampling techniques, including SMOTE, ADASYN, SMOTEENN, Tomek Links, and random over/under-sampling. Rather than evaluating these methods in isolation, the framework integrates them within the full pipeline, allowing analysis of their interactions with feature selection, normalization, and model choice.

\subsection{Learning Models}

The framework evaluates a diverse set of classifiers:

\begin{itemize}
\item \textbf{Logistic Regression (LR)} — linear baseline model.
\item \textbf{Support Vector Machine (SVM)} — margin-based classifier with kernel support.
\item \textbf{Decision Tree (DT)} — interpretable tree-based model.
\item \textbf{Random Forest (RF)} — ensemble of decision trees with bootstrap aggregation.
\item \textbf{Gradient Boosting (GB)} — sequential ensemble minimizing loss gradients.
\item \textbf{XGBoost (XGB)} — regularized and efficient gradient boosting implementation.
\end{itemize}

Additionally, two neural network architectures are considered:
a deep model with multiple hidden layers and regularization, and a shallow model for lightweight learning. Both are trained using Adam optimization, early stopping, and binary cross-entropy loss.

\subsection{Evaluation Metrics}

Performance is evaluated using accuracy, precision, recall, and F1-score to capture both overall predictive performance and class-specific behavior. For multi-class aggregation, three standard strategies are employed:

\begin{itemize}
\item \textbf{Macro averaging} — unweighted mean across classes
\item \textbf{Weighted averaging} — weighted by class support
\item \textbf{Micro averaging} — global computation over all samples
\end{itemize}

For binary classification, the metrics are defined as:
\begin{align}
\text{Precision} &= \frac{TP}{TP + FP}, \\
\text{Recall} &= \frac{TP}{TP + FN}, \\
\text{F1} &= \frac{2PR}{P + R}.
\end{align}

This multi-metric evaluation provides a comprehensive assessment of model performance, particularly under class imbalance.

\subsection{Multi-Level Analysis Framework}

To systematically analyze the pipeline search space, we adopt a multi-level analytical framework that captures performance behavior from global distribution to fine-grained component interactions.

\textbf{(1) Branch-Level Distribution Analysis.}  
Let $\mathcal{P}$ denote the set of all pipeline configurations. We analyze the distribution of performance metrics $\{M(p) \mid p \in \mathcal{P}\}$ to characterize global properties of the search space, including clustering, dispersion, and structural patterns. This analysis captures overall stability and sensitivity across configurations.

\textbf{(2) Pipeline-Level Performance Analysis.}  
We identify top- and bottom-performing configurations based on evaluation metrics such as weighted precision, recall, F1-score, and integrated performance score. Formally, we rank configurations $p \in \mathcal{P}$ according to $M(p)$ and analyze the corresponding component combinations. This enables identification of effective pipeline designs.

\textbf{(3) Component-Level Statistical Analysis.}  
For each component $C_i$, we aggregate performance over all configurations sharing the same value $v \in \mathcal{V}_i$. This yields statistics $\mu_{i,v}$ and $\sigma_{i,v}$, representing the mean performance and variability associated with each component value. This analysis isolates the general effect of individual components independent of specific pipeline compositions.

\textbf{(4) Cross-Component Interaction Analysis.}  
To capture dependencies between components, we analyze joint distributions of performance conditioned on multiple components, i.e., $M(p) \mid (C_i = v_i, C_j = v_j)$. This reveals interaction effects that cannot be observed through marginal analysis, demonstrating that pipeline performance is governed by combinations of components rather than isolated factors.

\textbf{(5) Robustness and Variability Analysis.}  
To evaluate reliability, we analyze performance across multiple random seeds. For each configuration, we compute mean and variance over repeated runs. Additionally, non-parametric statistical tests (e.g., Friedman and Wilcoxon tests) are applied to assess the significance of differences between configurations. This ensures that observed performance patterns reflect consistent behavior rather than stochastic variation.

\subsection{Component Importance and Search Space Analysis}

This subsection formalizes the analysis of pipeline component importance and search space structure. The goal is to identify which components significantly influence performance, how stable their effects are, and how configuration choices shape the optimization landscape.

\paragraph{Pipeline Search Space.}
Let $\mathcal{P}$ denote the pipeline search space, defined as the Cartesian product of $k$ components:
\begin{equation}
\mathcal{P} = \mathcal{V}_1 \times \mathcal{V}_2 \times \cdots \times \mathcal{V}_k,
\end{equation}
where each component $C_i$ takes values from a finite set $\mathcal{V}_i$. 
A pipeline configuration is denoted as $p = (v_1, v_2, \dots, v_k) \in \mathcal{P}$.
Let $M(p)$ denote the performance metric (e.g., Accuracy, F1-score) associated with configuration $p$.

\paragraph{Component-Level Aggregation.}
To evaluate the global effect of a component $C_i$, performance is aggregated over all configurations sharing the same value $v \in \mathcal{V}_i$:
\begin{equation}
\mu_{i,v} = \mathbb{E}[M(p) \mid C_i = v], \quad
\sigma_{i,v} = \mathrm{Std}(M(p) \mid C_i = v).
\end{equation}

Component-level statistics:
\begin{align}
\Delta_i &= \max_{v \in \mathcal{V}_i} \mu_{i,v} - \min_{v \in \mathcal{V}_i} \mu_{i,v}, \\
\sigma_i &= \mathbb{E}_{v \in \mathcal{V}_i} [\sigma_{i,v}], \\
S_i &= \mathrm{Var}_{v \in \mathcal{V}_i} (\mu_{i,v}),
\end{align}
where:
\begin{itemize}
    \item $\Delta_i$ measures the performance impact of component $C_i$,
    \item $\sigma_i$ measures stability across configurations,
    \item $S_i$ captures sensitivity to value changes, defined as
\end{itemize}
\begin{equation}
S_i = \frac{1}{|\mathcal{V}_i|} \sum_{v \in \mathcal{V}_i} \left( \mu_{i,v} - \bar{\mu}_i \right)^2, 
\quad
\bar{\mu}_i = \frac{1}{|\mathcal{V}_i|} \sum_{v \in \mathcal{V}_i} \mu_{i,v}.
\end{equation}
 
 \paragraph{Random Forest–Based Attribution.}

A Random Forest model predicts performance metrics from pipeline configurations.
The resulting feature importance provides a model-based estimate of component contribution:

\[
R_i = \mathrm{RFImportance}(C_i)
\]

where \(C_i\) denotes pipeline component \(i\), and \(R_i\) represents its importance score.

In implementation, this importance is computed using the built-in impurity-based feature importance of each tree and averaged across the ensemble:

\[
R_i = \frac{1}{T}\sum_{t=1}^{T} I_{t,i}
\]

where \(T\) is the number of trees and \(I_{t,i}\) is the feature importance of component \(i\) in tree \(t\), as returned by the Random Forest estimator.

Each tree-level importance \(I_{t,i}\) is derived from cumulative impurity reduction over all splits involving feature \(i\).

Importantly, this attribution reflects predictive relevance rather than causal direction; it identifies components that are informative for predicting performance variability across pipeline configurations, but does not indicate whether a component has a positive or negative effect on performance.

\paragraph{Integrated Definition of Component Importance.}
A potential direction for future work is the development of an integrated component importance measure that combines multiple factors, including performance impact, sensitivity, model-based attribution, and stability. Such a unified metric could provide a more comprehensive assessment of component influence across pipeline configurations. Component importance is defined as:
\begin{equation}
\mathcal{I}(C_i) = f\big( \Delta_i, \ \sigma_i, \ S_i, \ R_i \big),
\end{equation}
where importance is high if any of the following is true:
\begin{itemize}
    \item High performance impact ($\Delta_i$),
    \item High sensitivity across values ($S_i$),
    \item High Random Forest attribution ($R_i$),
    \item Acceptable stability (moderate $\sigma_i$).
\end{itemize}

\subsection{Value-Level Component Similarity Analysis}

To quantify functional similarity between different values of a pipeline component, we perform a value-level RMS-difference analysis across all pipeline branches.
This approach identifies configurations that yield nearly identical performance, enabling search space pruning and more efficient AutoML exploration.

\paragraph{Setup.}
Let $C_i$ denote a pipeline component with value set
$\mathcal{V}i = {v_1, v_2, \dots, v{|\mathcal{V}_i|}}$.
Let $\mathcal{B}$ denote the set of all pipeline branches (configurations).
For each branch $b \in \mathcal{B}$ and each value $v \in \mathcal{V}_i$, let
$M_m^b(v)$ denote the performance on metric $m$ when $C_i = v$ and other component values are fixed as in branch $b$.

\paragraph{Per-Branch RMS Computation.}
For each branch $b$ and each pair of values $(v_a, v_b)$ of component $C_i$, compute the root-mean-square difference across metrics:

\begin{equation}
\text{RMS}^b(v_a, v_b) = \sqrt{ \frac{1}{|\text{Metrics}|} \sum_{m \in \text{Metrics}} \big( M_m^b(v_a) - M_m^b(v_b) \big)^2 }.
\end{equation}

If metric normalization is desired, divide each squared difference by the corresponding metric variance $\mathrm{Var}_b(M_m^b)$ before taking the square root.

\paragraph{Aggregate Across Branches.}
To summarize the value-pair difference across all branches, compute the mean RMS:

\begin{equation}
\text{RMS}\text{total}(v_a, v_b) = \frac{1}{|\mathcal{B}|} \sum{b \in \mathcal{B}} \text{RMS}^b(v_a, v_b).
\end{equation}

This provides a single, interpretable measure of functional difference between $v_a$ and $v_b$ across the entire search space.

\paragraph{Interpretation.}
\begin{itemize}
\item Small $\text{RMS}\text{total}(v_a, v_b)$ indicates functional equivalence between $v_a$ and $v_b$.
\item Large $\text{RMS}\text{total}(v_a, v_b)$ indicates meaningful performance differences.
\end{itemize}

\paragraph{Operationalization.}
\begin{enumerate}
\item For each component $C_i$, iterate over all value pairs $(v_a, v_b)$.
\item Compute per-branch RMS differences $\text{RMS}^b(v_a, v_b)$ while keeping other component values fixed.
\item Aggregate over branches to compute $\text{RMS}_\text{total}(v_a, v_b)$.
\item Identify clusters of functionally equivalent values for search space reduction.
\end{enumerate}

\paragraph{Benefits.}
This RMS-based analysis provides a fine-grained, quantitative measure of redundancy and similarity in the pipeline search space, complementing component-level and branch-level analyses while offering an interpretable scale of performance difference.

\subsection{Cross-Seed Robustness Analysis}

Each pipeline configuration is evaluated across multiple random seeds. For each metric $m$, compute mean and standard deviation:
\begin{align}
\bar{m} = \frac{1}{R} \sum_{r=1}^{R} m_r, \quad
\sigma_m = \sqrt{\frac{1}{R} \sum_{r=1}^{R} (m_r - \bar{m})^2}.
\end{align}

Statistical significance is assessed using the Friedman test followed by Wilcoxon signed-rank tests.

The \textbf{Normalized Robust Rank Score (NRRS)} combines performance and stability:
\begin{equation}
\text{NRRS} = \bar{r} + \lambda \cdot \sigma_r,
\end{equation}
where $\bar{r}$ is mean rank and $\sigma_r$ is rank variance across runs. Lower NRRS indicates robust, high-performing pipelines.

\section{Experiments and Results}
\label{sec:experiment}

This section presents the experimental setup and results used to evaluate the proposed framework on the Pima and Stroke datasets, focusing on predictive performance and robustness across diverse pipeline configurations.  All experiments were conducted on a MacBook Air with Apple M4 chip, 16~GB RAM, running macOS 15.6.1.

\subsection{Datasets and Experimental Configuration}

Experiments were conducted on two representative healthcare datasets to evaluate the effectiveness of the proposed framework under heterogeneous conditions.

The \textbf{Pima Indians Diabetes dataset} consists exclusively of numerical features and represents a small-scale, moderately imbalanced binary classification task. In contrast, the \textbf{Healthcare Stroke dataset} includes both numerical and categorical features, requiring automatic encoding and presenting a more complex and highly imbalanced prediction scenario.

A unified automated configuration system was used to generate all experiments. The framework systematically explores combinations of the following components:

\begin{itemize}
    \item \textbf{Normalization order:} \texttt{Norm\_First} $\in \{\text{True}, \text{False}\}$
    \item \textbf{Train--test splits:} $\{0.1, 0.2\}$
    \item \textbf{Probability thresholds:} $\{0.35, 0.5\}$
    \item \textbf{Models:} SVM, Decision Tree, Logistic Regression, Random Forest, Gradient Boosting, XGBoost
    \item \textbf{Augmentation:} None, Gaussian Noise, MixUp
    \item \textbf{Imbalance handling:} None, SMOTE, ADASYN, Random Under-Sampling, Tomek Links
    \item \textbf{Feature selection:} \texttt{infgain}, \texttt{biMeanInfgain}, \texttt{biMaxInfgain}, or none
    \item \textbf{Feature count:} upper-range feature counts per dataset (e.g., 4--8 for total 8 features, 5--9 for total 9 features).
\end{itemize}

Four augmentation–imbalance modes were evaluated:  
(1) none, (2) augmentation only, (3) imbalance handling only, and (4) combined strategies.

Performance was evaluated using Accuracy, Macro/Micro/Weighted Precision, Recall, F1-score, and a composite Integrated Score.

To assess robustness, experiments on the Pima dataset were repeated across multiple random seeds. These seeds affect data splitting, augmentation sampling, and model initialization, enabling evaluation of performance stability and sensitivity to stochastic variation. This multi-seed design provides a more reliable estimate of generalization performance, particularly for small and imbalanced datasets.

\subsection{Branch-Based Execution and Experimental Scale}

Each unique configuration is defined as a \emph{branch}, representing a fully specified machine-learning pipeline. Branches are generated through deterministic enumeration and executed independently, enabling structured comparison across pipeline configurations.

In total, 18{,}720 branches were executed and grouped into 3{,}120 data-branch collections, forming a large-scale yet controlled experimental environment. This scale allows systematic exploration of interactions between preprocessing, augmentation, imbalance handling, and model selection.
 
 \subsection{Log-Driven Execution and Data Aggregation}

A key feature of the framework is its log-driven architecture. Each pipeline execution produces structured logs capturing configuration parameters, execution traces, prediction outputs, and evaluation metrics. These logs are organized in a hierarchical directory structure, where each experiment is associated with a dedicated \texttt{LogDir}, enabling full traceability from configuration to results.

For the Pima dataset, six independent runs were conducted, while the Stroke dataset includes five runs. Each run generates a corresponding \texttt{LogDir}, enabling systematic multi-run analysis.

The \texttt{staticsanalysis()} module serves as the central aggregation mechanism, transforming raw logs into unified analytical datasets. Its behavior is controlled by the \texttt{runid} parameter:

\begin{itemize}
    \item A specific \texttt{runid} enables within-run analysis.
    \item Multiple \texttt{runid}s enable cross-run aggregation.
    \item No \texttt{runid} aggregates all runs for global analysis.
\end{itemize}

This design decouples execution from analysis, allowing reproducible post hoc evaluation without re-running pipelines.

\paragraph{Evaluation Protocol.}
The framework adopts a two-level evaluation protocol: \emph{within-run} and \emph{cross-run} analysis.

\textbf{Within-run analysis} evaluates all pipeline configurations within a single \texttt{LogDir}, where executions are deterministic given a fixed seed. This enables controlled comparison of preprocessing strategies and model configurations.

\textbf{Cross-run analysis} aggregates results across independent runs with different random seeds, capturing variability from stochastic factors such as data splitting and model initialization. Performance is summarized using mean, standard deviation, and ranking-based measures (e.g., NRRS).

Statistical significance is assessed using non-parametric tests, including the Friedman test and Wilcoxon signed-rank test, ensuring robust comparison across pipeline configurations.

\subsection{Statistical Analysis}

All statistical analyses in this study are derived from the aggregated logs and merged CSV datasets generated by the \texttt{yvsoucom-iterkit} framework. The availability of both raw per-branch metrics and aggregated summaries enables comprehensive evaluation from multiple analytical perspectives.

All experimental data, tables, figures, and generated artifacts are publicly available via the GitHub repository\footnote{\url{https://github.com/yvsoucom/itekit-examples}}. These resources support full reproducibility and allow direct inspection of both high-performing and low-performing pipeline instances.

The statistical evaluation is organized into five complementary analytical components:

\textbf{(1) Branch-level distribution and trend analysis.}   
This analysis examines the full distribution of performance across all pipeline branches, revealing clustering behavior, dispersion patterns, and structural properties of the search space. It provides insight into stability and sensitivity across configurations.

\textbf{(2) Pipeline-level performance analysis.}  
Top- and bottom-performing configurations are identified based on key evaluation metrics, including weighted precision, recall, F1-score, and the integrated performance score. This analysis highlights effective combinations of preprocessing strategies and models, providing direct guidance for optimal pipeline design.

\textbf{(3) Component-level statistical analysis.}  
This analysis evaluates the average contribution of individual pipeline components—such as feature selection methods, normalization strategies, data augmentation techniques, imbalance-handling approaches, and classifiers—across all configurations. By aggregating results across branches, it isolates the general effect of each component independent of specific pipeline compositions.

\textbf{(4) Cross-component interaction analysis.}  
This component investigates how different pipeline elements jointly influence performance. Using merged datasets, interaction patterns between models, preprocessing strategies, and imbalance-handling methods are analyzed to reveal dependencies that are not observable through isolated component evaluation. This demonstrates that predictive performance is governed by interactions among components rather than individual factors alone.

\textbf{(5) Robustness and variability analysis.}  
This analysis evaluates the stability of pipeline performance across repeated runs and different random seeds. Statistical measures such as mean and variance are used to assess consistency, while non-parametric statistical tests further validate observed differences between pipeline configurations. This ensures that reported results reflect reliable and reproducible behavior rather than stochastic variation.

Representative figures and summary tables are included in the main text, while complete experimental results, detailed statistical outputs, and full merged datasets are provided in the Appendix to ensure transparency and enable further investigation.

 \subsubsection{Raw Metric Distributions and Branch-Level Trends}
\label{sec:branch_trends}
 
This section provides a fine-grained analysis of pipeline behavior by examining raw metric distributions and performance trends across all evaluated branches. Unlike aggregated summaries, this analysis exposes the full variability and structural characteristics of the explored configuration space.

Each branch corresponds to a fully specified pipeline instance. By analyzing metric distributions across all branches, it is possible to observe clustering behavior, dispersion patterns, and the presence of outlier configurations. These properties provide insight into both the stability of the framework and the sensitivity of performance to configuration choices.

 \paragraph{Raw Metric Analysis.}

For both the Pima and Stroke datasets, the highest F1 scores are consistently observed on the majority class (Class 0), which is expected due to class imbalance.

Table~\ref{tab:pima_top5_class_wise_F1} reports class-wise F1 scores for the Pima dataset using the abbreviated notation defined in the Pima Supplementary Materials. Detailed results for the Stroke dataset are also provided in the Stroke Supplementary Materials.

For the Pima dataset (Table~\ref{tab:pima_top5_class_wise_F1}), Class 0 F1 ranges from 0.905 to 0.909, Weighted-F1 from 0.865 to 0.885, and Macro-F1 from 0.848 to 0.876, indicating relatively balanced predictive performance across classes.

In contrast, the Stroke dataset shows very high Class 0 F1 (0.977–0.978) and Weighted-F1 (0.936–0.943), but substantially lower Macro-F1 (0.562–0.622), indicating degraded performance on minority classes and stronger class imbalance effects.

\begin{table}[htbp]
\centering
\caption{Top 5 Class-wise F1 Scores on Pima Dataset }

\footnotesize
\renewcommand{\arraystretch}{0.85}
\setlength{\tabcolsep}{3pt}

\begin{tabular}{l l l l l l l l l l l l l l l l}
\hline
F1 & C & RunID & Feat & FS & Sc & Aug & Imb & Model & Acc & W-F1 & M-F1 & Norm & Split & Prob & Seed \\
\hline
0.909 & 0 & 0754 & 4 & bMax & Std & MX & nIMB & SVM & 0.883 & 0.884 & 0.873 & L & 0.1 & 0.35 & 126 \\
0.909 & 0 & 0754 & 4 & bMean & Std & MX & nIMB & SVM & 0.883 & 0.884 & 0.873 & F & 0.1 & 0.35 & 126 \\
0.906 & 0 & 1919 & 5 & bMean & Std & MX & nIMB & RF & 0.870 & 0.866 & 0.849 & L & 0.1 & 0.50 & 7 \\
0.906 & 0 & 0754 & 4 & bMean & Std & MX & nIMB & GB & 0.870 & 0.866 & 0.849 & L & 0.1 & 0.50 & 126 \\
0.905 & 0 & 1411 & 6 & bMax & Std & NA & Tomek & XGB & 0.883 & 0.885 & 0.876 & L & 0.1 & 0.35 & 126 \\
\hline
\end{tabular}
\label{tab:pima_top5_class_wise_F1}
\end{table}

\paragraph{Trend and Geometric Analysis.}

Branch-level distributions reveal structured, quasi-cyclic variation in both datasets when projected along the branch axis, reflecting systematic differences across pipeline configurations.

For the Pima dataset, performance exhibits relatively regular fluctuations with alternating levels of dispersion. Branches form sequences of moderate and higher deviation, indicating phased variation across configuration groups. This suggests a stable and low-sensitivity landscape, where changes in certain components introduce limited and predictable variation.

In contrast, the Stroke dataset shows a bounded concentration--dispersion pattern. Metric values remain within relatively stable upper and lower bounds, while intermediate configurations exhibit higher dispersion before reconverging toward these boundaries. This behavior reflects stronger sensitivity to pipeline design, particularly under class imbalance, with alternating phases of divergence and stabilization.

Overall, Pima demonstrates a more uniform and stable distribution, whereas Stroke exhibits boundary-constrained and metric-sensitive variability. The recurrence of near-identical metric values across branches further indicates the presence of functionally equivalent configurations, suggesting opportunities for search space reduction in future experiments.

Full visualizations supporting these observations are provided in the Pima Supplementary Materials and Stroke Supplementary Materials .

\subsubsection{Pipeline-Level and Aggregate Performance Analysis}

Analysis of the top-ranked pipelines on the Pima dataset (Table~\ref{tab:pima_rank_metric_vertical}) reveals distinct optimization characteristics and structural properties of the search space. Full visualizations for both the Pima and Stroke datasets are provided in the Pima and Stroke Supplementary Materials.

\paragraph{Pima Dataset.}
Top-performing pipelines exhibit strong convergence across metrics, with Accuracy, Micro-F1, and Weighted-F1 consistently around $0.8831$, and Macro-F1 up to $0.8764$.

High-performing configurations are dominated by \textbf{XGBoost} and \textbf{SVM}. The best Macro Precision ($0.9032$) is achieved by an SVM pipeline (\texttt{4+ infgain + gaussian\_noise + noImbl + standard}, $\texttt{Split}=0.1$, $\texttt{P\_th}=0.35$), while the best Macro Recall ($0.8930$) and Macro-F1 ($0.8764$) are obtained by XGBoost pipelines using \texttt{6+biMean/biMax Infgain + noAug + TomekLinks + standard} with the same $\texttt{Split}$ and $\texttt{P\_th}$. Micro and Weighted metrics ($\approx 0.8831$--$0.8850$) are also dominated by these XGBoost configurations, though SVM pipelines achieve identical values in several cases.

Across top-ranked pipelines, consistent patterns emerge: $\texttt{Split}=0.1$ and $\texttt{P\_th}=0.35$ are dominant, \texttt{biMean/biMax Infgain} is frequently used, and imbalance handling (\texttt{noImbl}, \texttt{TomekLinks}) shows limited impact. Augmentation differs by model, with \texttt{noAug} common in XGBoost and \texttt{mixup} in SVM.

 \paragraph{Stroke Dataset.}

Top-ranked pipelines on the Stroke dataset achieve high majority-class performance, with Accuracy and Micro-F1 of $0.9534$–$0.9537$ and Weighted-F1 up to $0.9424$, but substantially lower Macro-F1 ($0.6511$–$0.6560$).

XGBoost pipelines (e.g., \texttt{6+infgain + noAUG + noIMB + Standard}) achieve the best Macro Precision ($0.9765$), typically with \texttt{Split}=0.1, \texttt{P\_th}=0.5. In contrast, Logistic Regression with imbalance handling (e.g., \texttt{SMOTE}, \texttt{ADASYN}) achieves higher Macro Recall (0.7800) and Macro-F1 (0.6560), usually with \texttt{Split}=0.2, \texttt{P\_th}=0.5.

Micro-F1 and Accuracy ($\approx 0.9537$) are consistently obtained by Gradient Boosting or Logistic Regression pipelines with \texttt{mixup + noIMB}, often using \texttt{P\_th}=0.35. Weighted-F1 ($0.9399$–$0.9424$) is dominated by XGBoost without imbalance handling.

No single optimal configuration emerges: \texttt{Split}=0.1–0.2 and \texttt{P\_th}=0.35–0.5 vary by metric.

Aggregate results in the  Pima and Stroke Supplementary Materials show clear differences across the same metrics on two  datasets. For Macro-F1, Pima achieves $0.6550$ (Std $0.0920$) versus Stroke $0.5116$ (Std $0.0482$). For Weighted-F1, Pima reaches $0.6707$ (Std $0.1011$), while Stroke is substantially higher at $0.8511$ (Std $0.0964$). Accuracy is also higher but more variable in Stroke ($0.8181$, Std $0.1450$) compared to Pima ($0.6736$, Std $0.0897$), indicating stronger majority-class dominance and greater variability.

These differences are consistent with the behavior of top-ranked pipelines. In Pima, top configurations achieve high Macro-F1 (up to $0.8764$), creating a wider performance gap relative to the overall mean and resulting in higher variance (Std $0.0920$). In contrast, Stroke shows consistently lower Macro-F1 (mean $0.5116$, best $\approx 0.6560$), leading to a more compressed distribution and lower variance (Std $0.0482$). This suggests that pipeline configurations have a stronger impact on class-balanced performance in Pima, whereas in Stroke, performance is more constrained by class imbalance.

\begin{table*}[htbp]
\caption{Top-ranked pipelines under different evaluation metrics on the Pima Indians Diabetes dataset.   S\_R represents Split Ratio  P\_th represents probability threshold. M\_ for Macro, m\_ for Micro, W\_ for Weighted metrics,  R\_S for Random Seed.}
\label{tab:pima_rank_metric_vertical}
\centering
\footnotesize
\renewcommand{\arraystretch}{0.85}
\setlength{\tabcolsep}{2pt}
\begin{tabular}{|c|l|c|>{\ttfamily\footnotesize\arraybackslash}p{8cm}|c|c|c|}
\hline
Rank & Metric & Value & LogDir & S\_R & P\_th &R\_S\\
\hline
\multirow{10}{*}{1} & M\_P & 0.9032 & \texttt{4/infgain/gaussian\_noise\_\_noImbl\_\_standard/sklearn\_SVM} & 0.1 & 0.35  & 23 \\ 
 & M\_R & 0.8930 & \texttt{6/biMeanInfgain/noAug\_\_TomekLinks\_\_standard/XGBmodel} & 0.1 & 0.35  & 126 \\ 
 & M\_F1 & 0.8764 & \texttt{6/biMaxInfgain/noAug\_\_TomekLinks\_\_standard/XGBmodel} & 0.1 & 0.35  & 126 \\ 
 & m\_P & 0.8831 & \texttt{6/biMaxInfgain/noAug\_\_TomekLinks\_\_standard/XGBmodel} & 0.1 & 0.35  & 126 \\ 
 & m\_R & 0.8831 & \texttt{6/biMaxInfgain/noAug\_\_TomekLinks\_\_standard/XGBmodel} & 0.1 & 0.35  & 126 \\ 
 & m\_F1 & 0.8831 & \texttt{4/biMeanInfgain/standard\_\_mixup\_\_noImbl/sklearn\_SVM} & 0.1 & 0.35  & 126 \\ 
 & W\_P & 0.8944 & \texttt{6/biMaxInfgain/noAug\_\_TomekLinks\_\_standard/XGBmodel} & 0.1 & 0.35  & 126 \\ 
 & W\_R & 0.8831 & \texttt{6/biMaxInfgain/noAug\_\_TomekLinks\_\_standard/XGBmodel} & 0.1 & 0.35  & 126 \\ 
 & W\_F1 & 0.8850 & \texttt{6/biMeanInfgain/noAug\_\_TomekLinks\_\_standard/XGBmodel} & 0.1 & 0.35  & 126 \\ 
 & Acc & 0.8831 & \texttt{6/biMaxInfgain/noAug\_\_TomekLinks\_\_standard/XGBmodel} & 0.1 & 0.35  & 126 \\ 
\hline
\multirow{10}{*}{2} & M\_P & 0.8906 & \makecell[l]{ \texttt{6/biMeanInfgain/standard\_\_gaussian\_noise\_\_noImbl/}\\\texttt{LogisticRegression}} & 0.1 & 0.5  & 7 \\ 
 & M\_R & 0.8930 & \texttt{6/biMaxInfgain/noAug\_\_TomekLinks\_\_standard/XGBmodel} & 0.1 & 0.35  & 126 \\ 
 & M\_F1 & 0.8764 & \texttt{6/biMeanInfgain/noAug\_\_TomekLinks\_\_standard/XGBmodel} & 0.1 & 0.35  & 126 \\ 
 & m\_P & 0.8831 & \texttt{4/biMaxInfgain/mixup\_\_noImbl\_\_standard/sklearn\_SVM} & 0.1 & 0.35  & 126 \\ 
 & m\_R & 0.8831 & \texttt{4/biMaxInfgain/mixup\_\_noImbl\_\_standard/sklearn\_SVM} & 0.1 & 0.35  & 126 \\ 
 & m\_F1 & 0.8831 & \texttt{6/biMaxInfgain/noAug\_\_TomekLinks\_\_standard/XGBmodel} & 0.1 & 0.35  & 126 \\ 
 & W\_P & 0.8944 & \texttt{6/biMeanInfgain/noAug\_\_TomekLinks\_\_standard/XGBmodel} & 0.1 & 0.35  & 126 \\ 
 & W\_R & 0.8831 & \texttt{4/biMaxInfgain/mixup\_\_noImbl\_\_standard/sklearn\_SVM} & 0.1 & 0.35  & 126 \\ 
 & W\_F1 & 0.8850 & \texttt{6/biMaxInfgain/noAug\_\_TomekLinks\_\_standard/XGBmodel} & 0.1 & 0.35  & 126 \\ 
 & Acc & 0.8831 & \texttt{4/biMaxInfgain/mixup\_\_noImbl\_\_standard/sklearn\_SVM} & 0.1 & 0.35  & 126 \\ 
\hline
\multirow{10}{*}{3} & M\_P & 0.8906 & \texttt{5/biMeanInfgain/minmax\_\_gaussian\_noise\_\_noImbl/sklearn\_SVM} & 0.1 & 0.5  & 126 \\ 
 & M\_R & 0.8830 & \texttt{5/biMaxInfgain/noAug\_\_noImbl\_\_standard/XGBmodel} & 0.1 & 0.35  & 126 \\ 
 & M\_F1 & 0.8727 & \texttt{4/biMaxInfgain/mixup\_\_noImbl\_\_standard/sklearn\_SVM} & 0.1 & 0.35  & 126 \\ 
 & m\_P & 0.8831 & \texttt{6/biMeanInfgain/noAug\_\_TomekLinks\_\_standard/XGBmodel} & 0.1 & 0.35  & 126 \\ 
 & m\_R & 0.8831 & \texttt{6/biMeanInfgain/noAug\_\_TomekLinks\_\_standard/XGBmodel} & 0.1 & 0.35  & 126 \\ 
 & m\_F1 & 0.8831 & \texttt{4/biMaxInfgain/mixup\_\_noImbl\_\_standard/sklearn\_SVM} & 0.1 & 0.35  & 126 \\ 
 & W\_P & 0.8868 & \texttt{7/biMaxInfgain/standard\_\_mixup\_\_TomekLinks/random\_forest} & 0.1 & 0.35  & 126 \\ 
 & W\_R & 0.8831 & \texttt{6/biMeanInfgain/noAug\_\_TomekLinks\_\_standard/XGBmodel} & 0.1 & 0.35  & 126 \\ 
 & W\_F1 & 0.8836 & \texttt{4/biMeanInfgain/standard\_\_mixup\_\_noImbl/sklearn\_SVM} & 0.1 & 0.35  & 126 \\ 
 & Acc & 0.8831 & \texttt{6/biMeanInfgain/noAug\_\_TomekLinks\_\_standard/XGBmodel} & 0.1 & 0.35  & 126 \\ 
\hline
\multirow{10}{*}{4} & M\_P & 0.8846 & \texttt{6/infgain/standard\_\_gaussian\_noise\_\_noImbl/LogisticRegression} & 0.1 & 0.5  & 126 \\ 
 & M\_R & 0.8830 & \texttt{5/biMeanInfgain/noAug\_\_noImbl\_\_standard/XGBmodel} & 0.1 & 0.35  & 126 \\ 
 & M\_F1 & 0.8727 & \texttt{4/biMeanInfgain/standard\_\_mixup\_\_noImbl/sklearn\_SVM} & 0.1 & 0.35  & 126 \\ 
 & m\_P & 0.8831 & \texttt{4/biMeanInfgain/standard\_\_mixup\_\_noImbl/sklearn\_SVM} & 0.1 & 0.35  & 126 \\ 
 & m\_R & 0.8831 & \texttt{4/biMeanInfgain/standard\_\_mixup\_\_noImbl/sklearn\_SVM} & 0.1 & 0.35  & 126 \\ 
 & m\_F1 & 0.8831 & \texttt{6/biMeanInfgain/noAug\_\_TomekLinks\_\_standard/XGBmodel} & 0.1 & 0.35  & 126 \\ 
 & W\_P & 0.8868 & \texttt{7/infgain/standard\_\_noAug\_\_TomekLinks/XGBmodel} & 0.1 & 0.35  & 126 \\ 
 & W\_R & 0.8831 & \texttt{4/biMeanInfgain/standard\_\_mixup\_\_noImbl/sklearn\_SVM} & 0.1 & 0.35  & 126 \\ 
 & W\_F1 & 0.8836 & \texttt{4/biMaxInfgain/mixup\_\_noImbl\_\_standard/sklearn\_SVM} & 0.1 & 0.35  & 126 \\ 
 & Acc & 0.8831 & \texttt{4/biMeanInfgain/standard\_\_mixup\_\_noImbl/sklearn\_SVM} & 0.1 & 0.35  & 126 \\ 
\hline
\multirow{10}{*}{5} & M\_P & 0.8846 &\makecell[l]{  \texttt{5/biMeanInfgain/gaussian\_noise\_\_noImbl\_\_standard/}\\ \texttt{LogisticRegression}} & 0.1 & 0.5  & 7 \\ 
 & M\_R & 0.8830 & \texttt{5/biMeanInfgain/standard\_\_noAug\_\_noImbl/XGBmodel} & 0.1 & 0.35  & 126 \\ 
 & M\_F1 & 0.8635 & \texttt{5/biMaxInfgain/noAug\_\_noImbl\_\_standard/XGBmodel} & 0.1 & 0.35  & 126 \\ 
 & m\_P & 0.8701 & \texttt{4/biMaxInfgain/standard\_\_noAug\_\_noImbl/XGBmodel} & 0.1 & 0.5  & 126 \\ 
 & m\_R & 0.8701 & \texttt{4/biMaxInfgain/standard\_\_noAug\_\_noImbl/XGBmodel} & 0.1 & 0.5  & 126 \\ 
 & m\_F1 & 0.8701 & \texttt{4/biMeanInfgain/minmax\_\_mixup\_\_noImbl/sklearn\_SVM} & 0.1 & 0.5  & 126 \\ 
 & W\_P & 0.8855 & \texttt{5/biMeanInfgain/noAug\_\_noImbl\_\_standard/XGBmodel} & 0.1 & 0.35  & 126 \\ 
 & W\_R & 0.8701 & \texttt{4/biMaxInfgain/standard\_\_noAug\_\_noImbl/XGBmodel} & 0.1 & 0.5  & 126 \\ 
 & W\_F1 & 0.8725 & \texttt{5/biMaxInfgain/standard\_\_noAug\_\_noImbl/XGBmodel} & 0.1 & 0.35  & 126 \\ 
 & Acc & 0.8701 & \texttt{4/biMaxInfgain/standard\_\_noAug\_\_noImbl/XGBmodel} & 0.1 & 0.5  & 126 \\ 
\hline
\end{tabular}
\end{table*}

 \begin{table*}[htbp]
\centering
\caption{Aggregate performance statistics across all pipeline configurations on the Pima dataset. Results are reported as mean and standard deviation over all branches.}
\label{tab:pima_aggregate}
 
\small
\setlength{\tabcolsep}{4pt}
\begin{tabular}{|c|c|c|}
\hline
Metric & Mean & Std \\
\hline
Accuracy & 0.673649383024383 & 0.0897363795392836 \\
Macro\_Precision & 0.6838601341830406 & 0.0691823082366285 \\
Macro\_Recall & 0.6826892501253036 & 0.0755243216727676 \\
Macro\_F1 & 0.6550064165406746 & 0.0919794462164404 \\
Weighted\_Precision & 0.7246319526049205 & 0.0672222636465707 \\
Weighted\_Recall & 0.673649383024383 & 0.0897363795392836 \\
Weighted\_F1 & 0.6706759109693264 & 0.1011450473334413 \\
Micro\_Precision & 0.673649383024383 & 0.0897363795392836 \\
Micro\_Recall & 0.673649383024383 & 0.0897363795392836 \\
Micro\_F1 & 0.673649383024383 & 0.0897363795392836 \\
Integrated\_Score & 0.6774302360275797 & 0.0818401652216246 \\
\hline
\end{tabular}
\end{table*}

 \subsubsection{Pipeline Component Analysis: RF-Based Importance and Component-Specific Analysis}
\label{sec:Pipeline_Component_Analysis}

\paragraph{RF-Based Component Importance (Component Level).}

Random Forest importance results (Figure~\ref{fig:pima_rf_importance} and the Pima and Stroke Supplementary Materials) quantitatively confirm the component-level trends identified in the metric analysis.

\textbf{Pima dataset:}  
Macro-F1: AugMethod ($0.454$), Model ($0.198$), imblMethod ($0.101$).  
Macro-Precision: AugMethod ($0.400$), Model ($0.344$), imblMethod ($0.101$).  
Macro-Recall: Model ($0.431$), AugMethod ($0.272$), imblMethod ($0.062$).
 
Overall, AugMethod emerges as the most consistently influential component across Macro-F1 and Macro-Precision, while Model contributes strongly across all metrics, particularly Macro-Recall.

\textbf{Stroke dataset:}  
Macro-F1: imblMethod ($0.406$), NormOrder ($0.153$), Model ($0.138$), AugMethod ($0.110$).  
Macro-Precision: Model ($0.215$), AugMethod ($0.141$), imblMethod ($0.100$), InfoGain ($0.095$).  
Macro-Recall: imblMethod ($0.685$), Model ($0.157$).
   
\begin{figure*}[!t]
\centering
\includegraphics[width=\textwidth]{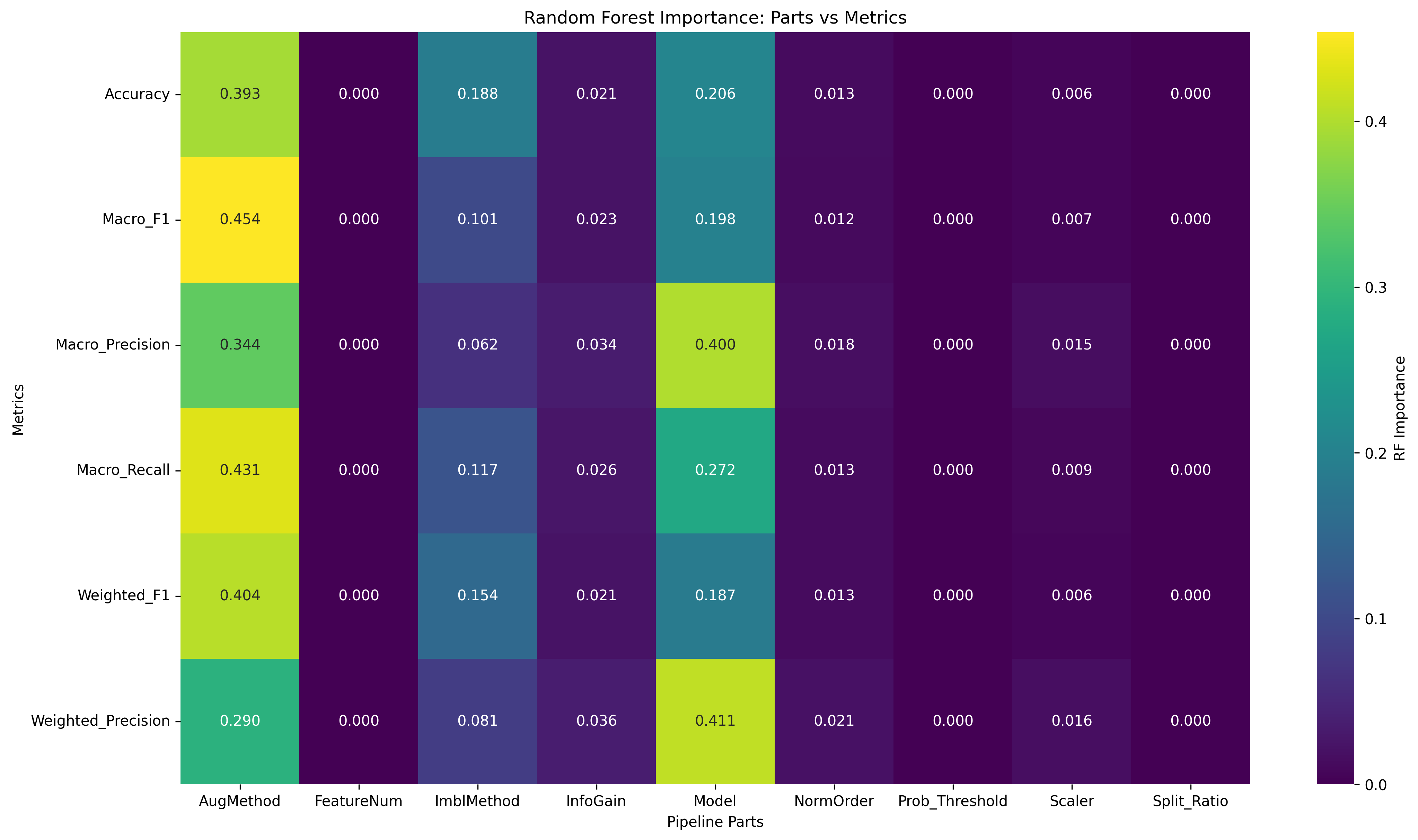}
\caption{Random Forest-based component importance for the Pima dataset.}
\label{fig:pima_rf_importance}
\end{figure*}

\paragraph{RF Importance (Detailed Parts Level).}

Detailed importance analysis (Figure~\ref{fig:pima_rf_importance_detail} and  the Pima and Stroke Supplementary Materials) reveals fine-grained component effects beyond aggregated metrics.

In the Pima dataset, augmentation-related methods dominate, with \texttt{Gaussian noise} emerging as the most influential component, while \texttt{mixup} and \texttt{noAug} show negligible influence. Model variants, such as \texttt{DTmodel}, also contribute substantially.

In contrast, the Stroke dataset is dominated by imbalance handling, with \texttt{RandomUnderSampler} as the most influential component, while other components contribute minimally.

Exact importance values are provided in the experimental logs.  Random Forest importance reflects sensitivity of performance to component perturbations rather than direct performance improvement or degradation.

\begin{figure*}[!t]
\centering
\includegraphics[width=\textwidth]{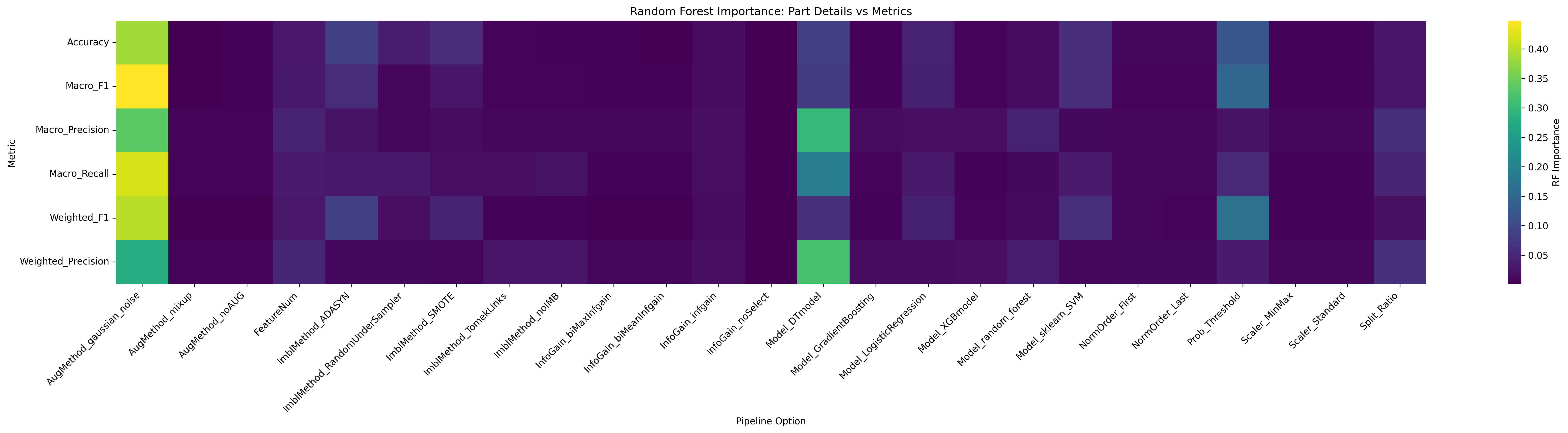}
\caption{Detailed Random Forest importance analysis for the Pima dataset.}
\label{fig:pima_rf_importance_detail}
\end{figure*}

\paragraph{Component-Specific Analysis.}

The unified heatmaps (Figure~\ref{fig:pima_unified_per_value_mega_heatmap} and  the Pima and Stroke Supplementary Materials) provide a component-level view of pipeline behavior across datasets.

In the Pima dataset, performance is influenced by multiple components in a relatively balanced manner. Augmentation (e.g., \texttt{gaussian\_noise}) consistently degrades performance across Macro-F1 ($0.58$), while configurations such as \texttt{mixup} and \texttt{noAug} achieve higher Macro-F1 (up to $0.71$), indicating that no augmentation is often more effective. Model choice introduces moderate variation in Macro-F1 (e.g., \texttt{sklearn\_SVM} $0.70$ vs.\ \texttt{DTmodel} $0.62$), while threshold tuning and split ratio also affect performance (Macro-F1 $0.66$--$0.68$). In contrast, scaling methods such as \texttt{standard} and \texttt{minmax} show minimal variation.

In the Stroke dataset, \texttt{TomekLinks} and \texttt{SMOTE} achieve the highest Macro-F1 (up to $\sim 0.53$), while model choice (e.g., \texttt{logistic regression}) exhibits the lowest Macro-F1 (around $\sim 0.49$).

 \begin{figure*} [!t]
  \centering
 
    \includegraphics[width=\textwidth,height=0.95\textheight,keepaspectratio]{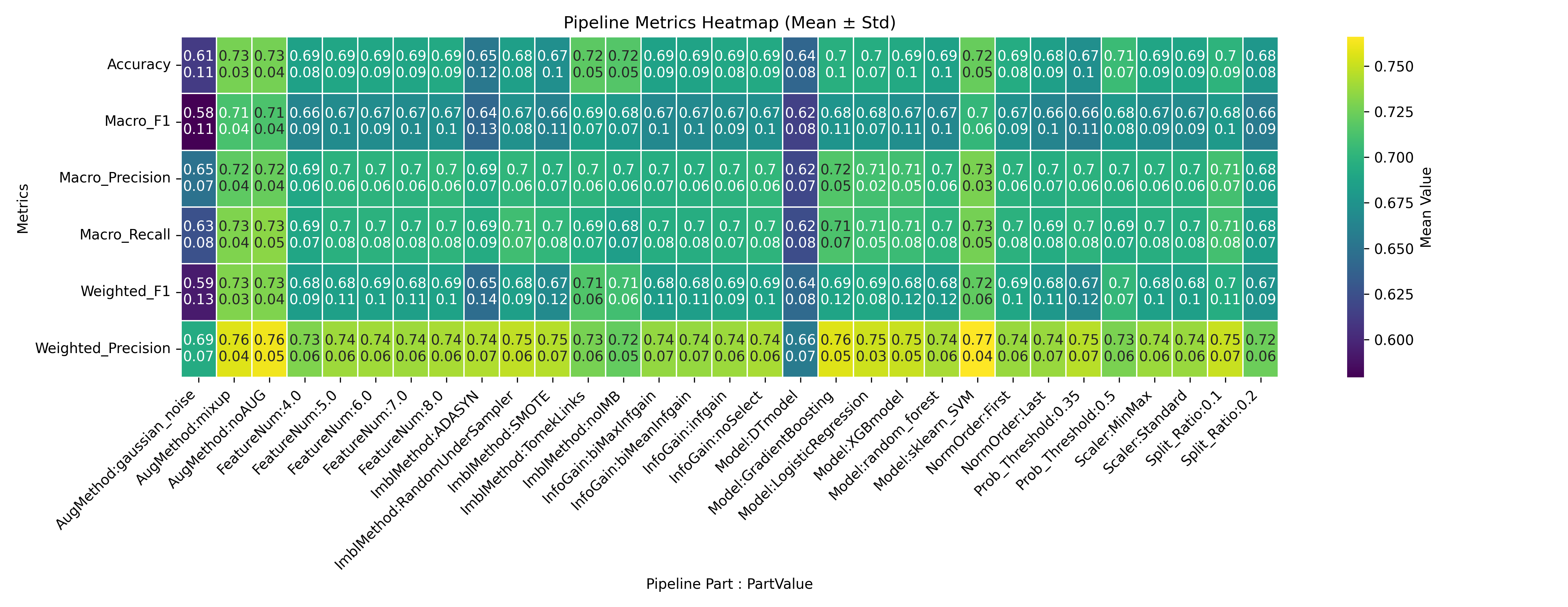}

    \caption{pima unified metric mega heat map for branches}
              \label{fig:pima_unified_per_value_mega_heatmap}

\end{figure*}

 \subsubsection{Value-Level Component Similarity Analysis}

Tables~\ref{tab:pima_value_similarity_all} summarize the value-level similarity analysis for the Pima dataset. Ranking value-level RMS similarities reveals structured redundancy patterns across pipeline components.

\paragraph{Similarity-Based Insights.}
 
\textit{Feature Dimensionality:} Configurations with 4–8 selected features form a compact similarity group, indicating reduced sensitivity to the exact number of selected features within this range (RMS: 0.0342--0.0414).

\textit{Feature Selection:} FS-BMax and FS-BMean exhibit highly similar behavior (RMS: 0.0252), suggesting interchangeable functionality.

\textit{Data Transformation:} Mixup shows high similarity to no augmentation (RMS: 0.0279), indicating limited additional effect in the evaluated setting, whereas Gaussian noise produces larger deviations.

\textit{Imbalance Handling:} TomekLinks and noIMB form a similarity pair with low RMS distance (0.0325).

\textit{Model Level:} Boosting-based methods (GB and XGB) exhibit strong similarity (RMS: 0.0319), whereas tree-based and linear models show larger divergence, reflecting fundamentally different decision boundaries.

\textit{Other Components:} Normalization order, probability thresholds, and data split ratios yield relatively small RMS differences, indicating limited sensitivity within the explored search space.

\paragraph{Integrated Future Experiment Configuration Strategy.}
We integrate component importance analysis with value-level similarity and performance-based evaluation to support principled AutoML search space reduction.

At the component level, we assess each pipeline component’s importance based on its overall impact on performance. Components with consistently low importance are considered for simplification.

At the value level, two complementary criteria are applied:  
(1) \textit{Similarity-based pruning:} groups values with low RMS differences, allowing representative configurations to be selected.  
(2) \textit{Performance-aware selection:} prioritizes values that contribute to high-performing configurations while deprioritizing consistently weak performers.

This combined strategy reduces redundancy while preserving diversity among high-impact components, enabling more efficient and effective AutoML search space exploration.

\begin{table*}[htbp]
\caption{Value-Level Similarity (RMS) across all pipeline components on Pima dataset}
\label{tab:pima_value_similarity_all}

\raggedright
\scriptsize

\textbf{Abbreviations:}  
F denotes the number of selected features.  
FS denotes feature selection methods: BMax (biMaxInfgain), BMean (biMeanInfgain), IG (information gain), None (no selection).  
Sc denotes scaling: MinMax (min–max normalization), Std (standardization).  
Aug denotes data augmentation: GN (Gaussian noise), Mix (mixup), None.  
Imb denotes imbalance handling: ADA (ADASYN), RUS (random undersampling), SMOTE, TL (Tomek links), None.  
Models include DT (decision tree), GB (gradient boosting), LR (logistic regression), XGB, RF (random forest), and SVM.  
Norm denotes normalization order (F: first, L: last).  
Thr denotes probability threshold, and Split denotes test split ratio.  

 \vspace{0.5em}
 
 \centering
\footnotesize
\renewcommand{\arraystretch}{0.85}
\setlength{\tabcolsep}{2pt}

\textbf{(a) Feature and Feature Selection Components}

\begin{tabular}{lrrrrrrrrr}
\toprule
 & F:4 & F:5 & F:6 & F:7 & F:8 & FS:biMax & FS:biMean & FS:inf & FS:none \\
\midrule
F:4 &  & 0.0381 & 0.0397 & 0.0414 & 0.0405 \\
F:5 & 0.0381 &  & 0.0342 & 0.0364 & 0.0388 \\
F:6 & 0.0397 & 0.0342 &  & 0.0345 & 0.0373 \\
F:7 & 0.0414 & 0.0364 & 0.0345 &  & 0.0355 \\
F:8 & 0.0405 & 0.0388 & 0.0373 & 0.0355 &  \\

FS:biMax &  &  &  &  &  &  & 0.0252 & 0.0394 & 0.0405 \\
FS:biMean &  &  &  &  &  & 0.0252 &  & 0.0392 & 0.0411 \\
FS:inf &  &  &  &  &  & 0.0394 & 0.0392 &  & 0.0458 \\
FS:none &  &  &  &  &  & 0.0405 & 0.0411 & 0.0458 &  \\
\bottomrule
\end{tabular}

 \vspace{0.5em}

\textbf{(b) Data Transformation Components}

\begin{tabular}{lrrrrrrrrrrrr}
\toprule
 & Sc:minmax & Sc:std 
 & Aug:noise & Aug:mix & Aug:none 
 & Imb:ADA & Imb:RUS & Imb:SMOTE & Imb:Tomek & Imb:none \\
\midrule

Sc:minmax &  & 0.0438 \\
Sc:std & 0.0438 &  \\

Aug:noise &  &  &  & 0.1045 & 0.1028 \\
Aug:mix &  &  & 0.1045 &  & 0.0279 \\
Aug:none &  &  & 0.1028 & 0.0279 &  \\

Imb:ADA &  &  &  &  &  &  & 0.0435 & 0.0383 & 0.0648 & 0.0672 \\
Imb:RUS &  &  &  &  &  & 0.0435 &  & 0.0395 & 0.0567 & 0.0595 \\
Imb:SMOTE &  &  &  &  &  & 0.0383 & 0.0395 &  & 0.0548 & 0.0571 \\
Imb:Tomek &  &  &  &  &  & 0.0648 & 0.0567 & 0.0548 &  & 0.0325 \\
Imb:none &  &  &  &  &  & 0.0672 & 0.0595 & 0.0571 & 0.0325 &  \\

\bottomrule
\end{tabular}

\vspace{0.5em}

\textbf{(c) Model and Decision Components}

\begin{tabular}{lrrrrrrrrrrrr}
\toprule
 & M:DT & M:GB & M:LR & M:XGB & M:RF & M:SVM 
 & Norm:first & Norm:last 
 & Thr:.35 & Thr:.5 
 & Split:.1 & Split:.2 \\
\midrule

M:DT &  & 0.0799 & 0.0970 & 0.0792 & 0.0736 & 0.0992 \\
M:GB & 0.0799 &  & 0.0510 & 0.0319 & 0.0422 & 0.0510 \\
M:LR & 0.0970 & 0.0510 &  & 0.0523 & 0.0581 & 0.0384 \\
M:XGB & 0.0792 & 0.0319 & 0.0523 &  & 0.0419 & 0.0514 \\
M:RF & 0.0736 & 0.0422 & 0.0581 & 0.0419 &  & 0.0596 \\
M:SVM & 0.0992 & 0.0510 & 0.0384 & 0.0514 & 0.0596 &  \\

Norm:first &  &  &  &  &  &  &  & 0.0349 \\
Norm:last &  &  &  &  &  &  & 0.0349 &  \\

Thr:.35 &  &  &  &  &  &  &  &  &  & 0.0566 \\
Thr:.5 &  &  &  &  &  &  &  & 0.0566 &  \\

Split:.1 &  &  &  &  &  &  &  &  &  &  &  & 0.0472 \\
Split:.2 &  &  &  &  &  &  &  &  &  &  & 0.0472 &  \\

\bottomrule
\end{tabular}
\end{table*}

 \subsubsection{Cross-Component Interaction Analysis}
\label{sec:part_part_correlation}

To characterize the structure of the AutoML search space, we analyze part--part cross-component correlation matrices on both the Pima and Stroke datasets. These matrices quantify statistical dependencies between pipeline components by measuring the co-variation of their induced performance across all evaluated configurations.

Given the large number of components in the pipelines, analyzing all pairwise correlations can be overwhelming and less informative. Therefore, we focus on selected subsets of components to highlight meaningful interactions, including those with high mean correlation, high standard deviation of correlation, and clustered groups identified from the full correlation structure. Within-component correlations are zero by construction due to the branch-based pipeline design, where only one value per component is active in each configuration. As a result, correlation analysis is meaningful only across different components.

\begin{figure*} [!t]
  \centering
 
    \includegraphics[width=\textwidth,height=0.95\textheight,keepaspectratio]{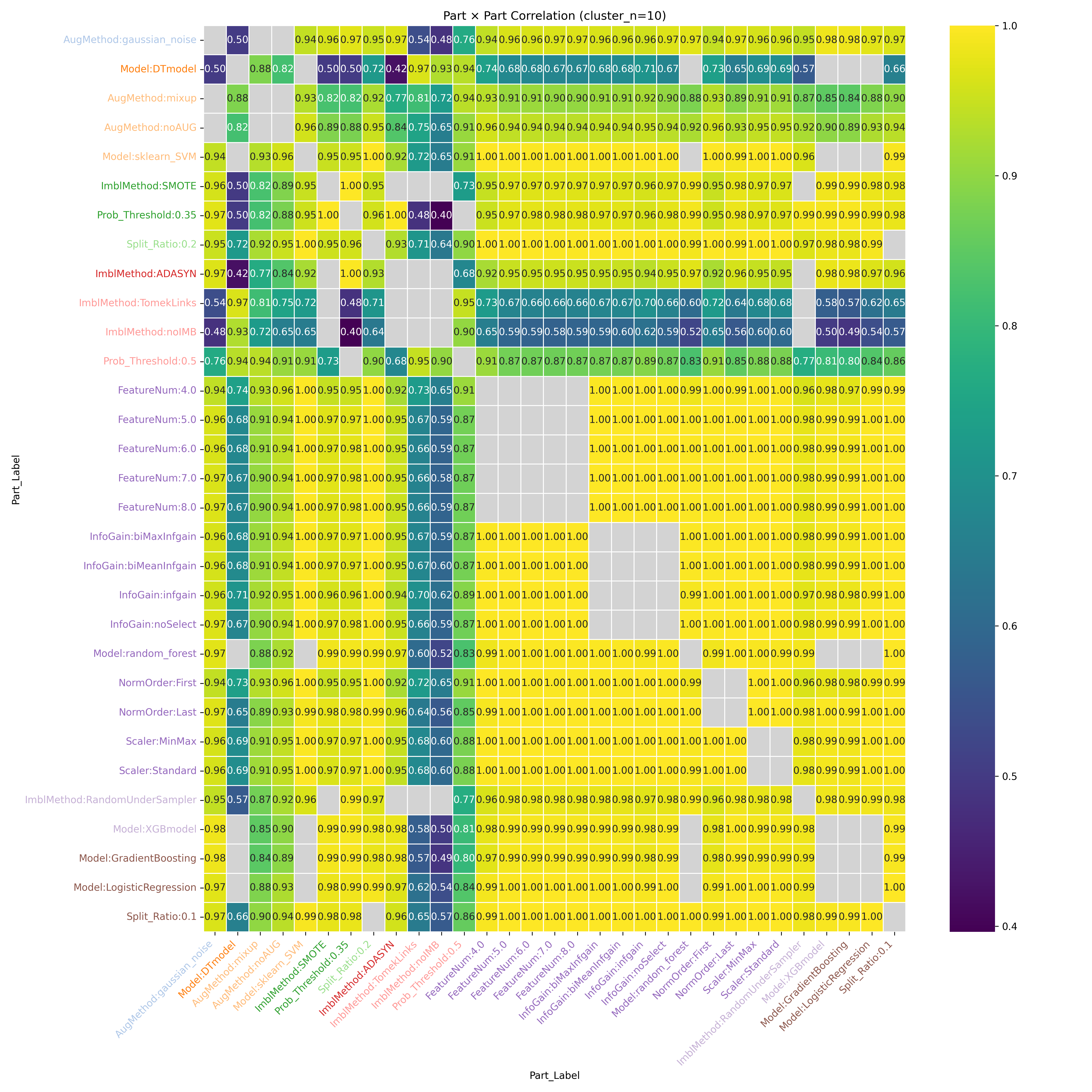}

    \caption{pima   cluster10  part vs part heat map}
              \label{fig:pimclusteredPart × Part Correlation}

\end{figure*}

For the Pima dataset, Figure~\ref{fig:pimclusteredPart × Part Correlation} presents clustered part-to-part correlations,  while additional correlation analyses are provided in the Pima supplementary material.
 
Across components, interaction strengths vary by method group. Decision tree-based models (\texttt{DTModel}) show moderate-to-high correlations (0.42–0.88), with typical values around 0.68. Imbalance-handling methods such as \texttt{TomekLinks} exhibit a wider range (0.54–0.97), with most correlations centered around 0.66. The no-imbalance configuration (\texttt{noIbml}) shows similar variability (0.40–0.93), with an average level near 0.65. Overall, all components demonstrate consistently moderate to high cross-component correlation, indicating strong structural coupling within the pipeline design space.

For the Stroke dataset, the corresponding correlation analyses—including top-10 mean correlation patterns, high-variance correlation patterns, and clustered structures—are provided in the Stroke supplementary material.

The RandomUnderSampler exhibits correlation values ranging from 0.74 to 0.91, with most values concentrated around 0.81, indicating relatively strong but more variable interactions compared to the Pima dataset.

These analyses help identify groups of preprocessing, augmentation, normalization, and modeling components that tend to co-vary in their effects. By focusing on subsets of high-impact or highly variable components, we reduce complexity while highlighting the most influential interactions. This approach provides insight into synergistic and redundant pipeline structures and supports future optimization strategies, including higher-order component interactions.

\subsubsection{Cross-Seed Robustness Analysis }

 To rigorously evaluate the stability and reliability of pipeline performance under stochastic conditions, each configuration was executed across multiple random seeds, affecting data splitting, augmentation and imbalance sampling, as well as model initialization. This experimental design enables a comprehensive robustness assessment by jointly considering predictive performance and variability. All results in this subsection are reported for the \textbf{Pima dataset} only and are further detailed in the Pima Supplementary Materials.
 
\paragraph{Performance and Stability Analysis Under Stochastic Variation.}
 
 Figure~\ref{fig:pima_cross_seed_deviation} illustrates the joint distribution of mean Macro-F1 and standard deviation across models, highlighting the trade-off between predictive performance and stability. SVM achieves the highest Macro-F1 (0.689) but also exhibits the highest variability ($\sigma \approx 0.029$), indicating strong sensitivity to stochastic perturbations. In contrast, ensemble methods (XGBoost, Gradient Boosting, Random Forest) provide a more balanced trade-off, achieving competitive Macro-F1 (0.645--0.659) with moderate variability ($\sigma \approx 0.023$--$0.026$). Decision Trees exhibit the lowest variability ($\sigma \approx 0.012$), reflecting high stability but limited predictive performance (Macro-F1 $\approx 0.652$), while Logistic Regression lies in an intermediate regime (Macro-F1 $\approx 0.671$,  variability $\sigma \approx 0.022$), achieving slightly higher mean performance than ensemble methods, although the difference remains within the range of cross-seed variability.
 
\paragraph{Statistical Significance Under Stochastic Variation.}
Friedman tests (Table~\ref{tab:friedman_results}) show statistically significant differences across models for all evaluation metrics ($\chi^2 \approx 26.38$--$27.90$, $p \ll 0.05$), with a constant critical difference (CD = 2.53), confirming that the observed ranking structure is consistent under stochastic variation. The stability of test statistics across metrics further indicates a coherent and reproducible ranking pattern. However, statistical significance does not imply equal robustness, as cross-seed variability differs substantially across models. Overall, ensemble methods provide the most reliable balance between performance and stability, while SVM prioritizes accuracy at the cost of robustness.

\begin{figure*}
\centering
\includegraphics[width=0.8\linewidth]{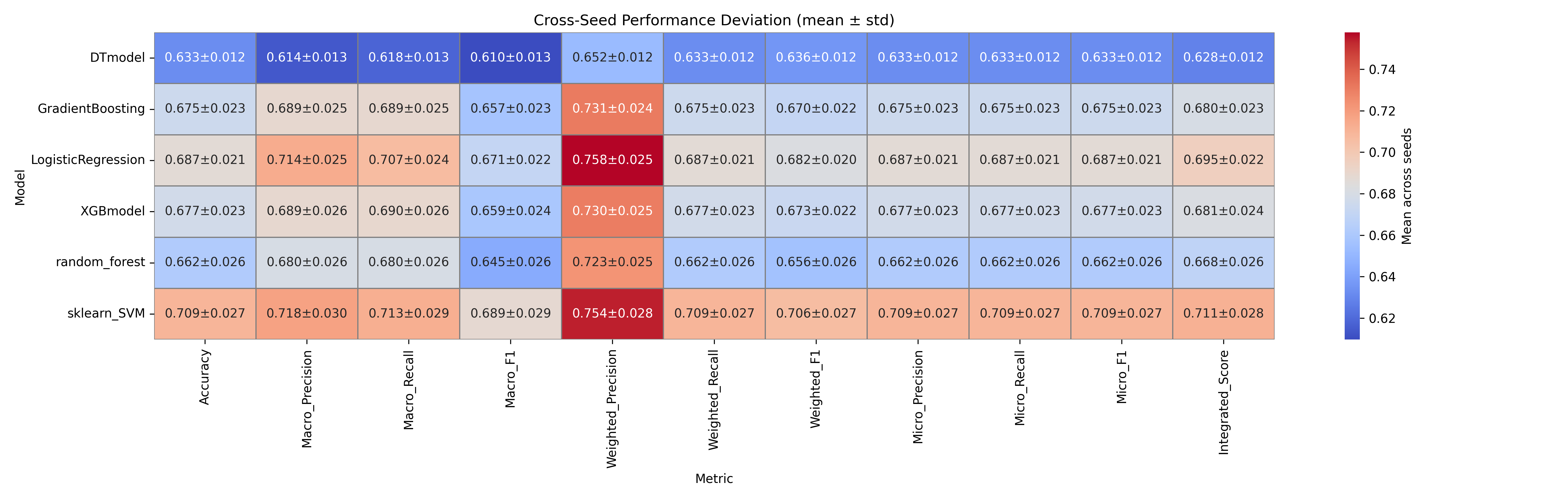}
\caption{Pima cross-seed mean performance and standard deviation across random seeds.}
\label{fig:pima_cross_seed_deviation}
\end{figure*}

\begin{table*}[t]
\centering
\caption{Friedman test results across evaluation metrics (computed over random seeds).}
\label{tab:friedman_results}
\begin{tabular}{lccc}
\toprule
Metric & $\chi^2$ & $p$-value & CD \\
\midrule
Accuracy            & 27.81 & $3.97\times10^{-5}$ & 2.53 \\
Macro Precision     & 27.90 & $3.80\times10^{-5}$ & 2.53 \\
Macro Recall        & 27.05 & $5.58\times10^{-5}$ & 2.53 \\
Macro F1            & 27.81 & $3.97\times10^{-5}$ & 2.53 \\
Weighted Precision  & 27.05 & $5.58\times10^{-5}$ & 2.53 \\
Weighted Recall     & 27.81 & $3.97\times10^{-5}$ & 2.53 \\
Weighted F1         & 27.81 & $3.97\times10^{-5}$ & 2.53 \\
Micro Precision     & 27.81 & $3.97\times10^{-5}$ & 2.53 \\
Micro Recall        & 27.81 & $3.97\times10^{-5}$ & 2.53 \\
Micro F1            & 27.81 & $3.97\times10^{-5}$ & 2.53 \\
Integrated Score    & 26.38 & $7.53\times10^{-5}$ & 2.53 \\
\bottomrule
\end{tabular}
\end{table*}

\subsubsection{Optimal Configuration Design Based on Experimental Analysis}
\label{sec:optimal_conf_design}

To investigate the contribution of individual pipeline components, a series of ablation experiments were conducted on the Pima Indians Diabetes dataset. Based on preliminary empirical analysis, both data augmentation and imbalance-handling strategies were found to have negligible or negative impact on Macro-F1 performance in this setting. Consequently, these components were excluded from the search space in subsequent experiments, and all pipelines were evaluated under a simplified configuration to ensure fair comparison.

Furthermore, redundancy in the feature selection space was analyzed. The original configuration set 
\texttt{config.set\_fs\_method(["biMeanInfgain", "biMaxInfgain", "infgain"])} 
was examined, and \texttt{biMeanInfgain} was removed due to its highly similar behavior to \texttt{biMaxInfgain} in terms of selected feature subsets and downstream predictive performance. The remaining feature selection methods were retained to preserve diversity while reducing computational redundancy.

In addition to feature selection ablation, we further investigated the impact of decision threshold and data splitting strategies across multiple experimental versions (V2--V6). The configurations are defined as follows:

\begin{itemize}
\item V2: $\{\text{threshold} \in \{0.5, 0.35\},\ \text{split ratio}  \in  \{0.1, 0.2\}\}$
\item V3: $\{\text{threshold} = 0.5,\ \text{split ratio} \in \{0.1, 0.2\}\}$
\item V4: $\{\text{threshold} \in \{0.5, 0.35\},\ \text{split ratio} = 0.2\}$
\item V5: $\{\text{threshold} = 0.5,\ \text{split ratio} = 0.2\}$
\item V6: $\{\text{threshold} = 0.5,\ \text{split ratio} \in \{0.1, 0.15, 0.2, 0.25, 0.3\}\}$
\end{itemize}

 The results  in Table~\ref{tab:pima_results}
 indicate that progressive simplification of the search space significantly reduces the number of candidate pipelines and overall computational cost. However, this reduction is accompanied by a general decline in predictive performance, as measured by Macro-F1.

In particular, the split ratio parameter exhibits a significant influence on model performance. Experimental results show that a split ratio of 0.1 consistently outperformed 0.2 in terms of Macro-F1 under the current configuration of the Pima Indians Diabetes dataset. This suggests that, in small-scale medical datasets, performance is highly sensitive to data partitioning strategies.

Furthermore, the experimental results reveal a strong interaction effect between the split ratio and decision threshold parameters. Specifically, the configuration combining a split ratio of 0.2 with a decision threshold of 0.5 consistently produced the lowest Macro-F1 among all evaluated settings.

\begin{table}[ht]
\centering
\small
\begin{tabular}{cccccccccccc}
\hline
expV&FN& FS & Scaler & Aug & Imbl & Model & TH &SR&  M-F1 \\
\hline
v2&5 & biMax& standard & noAug & noImbl & XGB & 0.35  &0.1&  0.8635  \\
v3&4 & biMax & standard & noAug & noImbl & XGB  & 0.5 & 0.1&  0.8548  \\
v4&7 & infgain & minmax & noAug & noImbl & XGB  & 0.35 & 0.2&  0.8127  \\
v5&8 & noSelect & minmax & noAug & noImbl & XGB  & 0.5 & 0.2&  0.7818  \\
v6&4 & biMax & standard & noAug & noImbl & XGB  & 0.5 & 0.1& 0.8548  \\

\hline
\end{tabular}
 
\caption{Experimental results on the Pima Indians Diabetes dataset showing the effect of configuration removal on Macro-F1 performance.}
\label{tab:pima_results}
\end{table}

 The experimental results on the Stroke   in Table~\ref{tab:stroke_results}
 Prediction dataset demonstrate that progressive refinement of the search space leads to a measurable improvement in Macro-F1 performance, increasing from 0.6560 in the baseline configuration to 0.6766 in the optimized configuration (V7).

Across all configurations, it is observed that imbalance handling plays a more significant role than data augmentation or feature selection alone. In particular, the introduction of RandomOverSampler (ROS) consistently improves Macro-F1 compared to configurations without explicit imbalance correction.

Furthermore, while feature selection and model variation contribute to performance fluctuations, the most stable improvements are associated with imbalance-aware configurations combined with moderate decision threshold tuning (e.g., 0.30).

The best-performing configuration (V7) achieves a Macro-F1 of 0.6766, representing a consistent improvement over earlier baselines. This indicates that, for highly imbalanced medical datasets such as stroke prediction, carefully designed imbalance handling strategies contribute more effectively to predictive performance than increasing model complexity or expanding feature selection space.

Detailed configuration settings and implementation specifics are provided in the accompanying source code file \texttt{stroke\_example\_v1.py}   in \url{https://github.com/yvsoucom/itekit-examples/releases/tag/v1.1}.

\begin{table}[ht]
\centering
\small
\begin{tabular}{ccccccccccc}
\hline
expV&FN & FS & Scaler & Aug & Imbl & Order&Model & TH &SR& M-F1   \\
\hline
v1& 9 & biMaxI & minmax   & GN & noImbl & last&DT & 0.50 & 0.10& 0.6556   \\
v2& 7 & infgain & standard   & GN & noImbl & last&RF & 0.25 & 0.10& 0.6594  \\
v3& 9 & infgain & minmax   & GN & noImbl & first&XGB & 0.35 & 0.20& 0.6472  \\
v4& 10 & noSelect & standard   & mixup & noImbl & last&XGB & 0.30 & 0.10& 0.6563 \\
v6& 7 & infgain & standard   & GN & ROS & first&RF & 0.20 & 0.10& 0.6710 \\
v7& 7 & infgain & standard   & GN & ROS & first&RF & 0.30 & 0.10& 0.6766\\

\hline
\end{tabular}
\caption{Experimental results on the Stroke Prediction dataset showing the effect of feature selection, augmentation, and model configuration on Macro-F1 performance.}
\label{tab:stroke_results}
\end{table}

\section{Discussion}
\label{sec:Discussion}

 The experimental results demonstrate that the proposed \texttt{yvsoucom-iterkit} framework enables systematic, large-scale, and reproducible benchmarking of tabular data prediction pipelines. By explicitly exposing the configuration space across preprocessing, feature selection, augmentation, and model design, the framework not only identifies which pipelines perform best, but also provides insights into the factors driving performance through interpretable, log-driven analysis.

 \subsection{Distinctive Features of the Proposed Framework}

A key contribution of this work is the design of the \texttt{yvsoucom-iterkit} framework, which fundamentally shifts the AutoML paradigm. Unlike conventional AutoML approaches that treat pipeline optimization as a black-box search problem, the proposed framework models each configuration as a deterministic, traceable log entity. This log-centric approach ensures full transparency and reproducibility by maintaining a complete record of every experiment, configuration, and outcome.

By embracing the full configuration space, the framework enables structured and exhaustive exploration across all components, from data preprocessing to model design. This comprehensive search spans every step, from raw data manipulation to final model outputs, offering insights into how different configurations interact at each stage of the pipeline. Instead of focusing solely on optimizing a single performance metric, \texttt{yvsoucom-iterkit} facilitates a full-space search, providing both performance metrics and detailed analysis of how various components contribute to the pipeline’s effectiveness. This approach empowers users to explore the entire configuration space, revealing the impact of choices in preprocessing techniques, feature representations, model choices  on performance.

The framework’s ability to trace results from raw data through to component-level outcomes is a critical feature. This capability allows users to conduct detailed, log-driven analysis, examining both the overall pipeline and individual components. By tracking how configuration settings impact each stage of the pipeline, users can gain valuable insights into the specific steps that drive improvements, ultimately leading to better model performance and more effective configurations.

Additionally, the framework supports multi-run and multi-seed experimentation, enabling a comprehensive evaluation of performance variability and stability. This approach shifts AutoML from a single-run optimization paradigm to a \textbf{statistically grounded experimental methodology}, where conclusions are derived from aggregated evidence rather than isolated results. By accounting for stochastic fluctuations, this rigorous methodology ensures that the findings are not biased, offering a more reliable basis for deploying models in real-world, dynamic environments.

A key advantage of this framework is its ability to explore the full configuration space, encompassing various components such as feature selection, augmentation, imbalance handling, and model choice. This flexibility enables users to adjust settings at both the high level (e.g., model selection) and the component level (e.g., preprocessing techniques), allowing them to observe the impact of these adjustments on performance. By experimenting with different configurations across the entire space, users gain deep insights into the specific components that drive optimal pipeline performance.

\subsection{Findings and Implications from Two Datasets: Insights Using the \texttt{yvsoucom-iterkit} Framework}

 The results provide critical insights into the key performance drivers, component importance, and robustness characteristics of pipelines constructed using the \texttt{yvsoucom-iterkit} framework.

\paragraph{Top Performance.}

Analysis of raw metric distributions and branch-level behavior reveals repetitive clustering patterns across pipelines, where multiple configurations produce near-identical outcomes, indicating a redundant configuration space with many-to-one mappings. The Pima dataset exhibits a more stable and low-sensitivity landscape, while the Stroke dataset demonstrates stronger boundary-constrained and imbalance-sensitive fluctuations, together suggesting functionally equivalent configurations and a reducible search space.

On the Pima dataset, top pipelines achieve Macro-F1 up to 0.8764, indicating relatively balanced class performance. XGBoost and SVM dominate top rankings, where SVM achieves strong peak performance but exhibits higher variability across configurations and random seeds, while ensemble methods show greater robustness.

In contrast, the Stroke dataset exhibits high Accuracy and Micro-F1 (0.9534–0.9537) and Weighted-F1 up to 0.9424, but significantly lower Macro-F1 (0.6511–0.6560), reflecting stronger class imbalance effects.
 
A notable finding is that configurations effective on the Pima dataset do not directly transfer to Stroke. Pima shows more stable and uniform behavior across experimental configurations, whereas Stroke requires adjusted configuration settings to achieve better performance, particularly due to higher sensitivity and class imbalance.

\paragraph{Component Importance and Correlations.}  
  
Random Forest-based importance analysis reveals clear component-level drivers of performance. On the Pima dataset, augmentation (\texttt{AugMethod},  Macro-F1  0.454), model choice (  Macro-F1 0.198), and imbalance handling ( Macro-F1 0.101) are the most influential factors, indicating that performance is primarily governed by data transformation and model selection. On the Stroke dataset, imbalance handling dominates (Macro-F1 0.406), followed by NormOrder , model choice and augmentation.
 
Fine-grained analysis confirms that augmentation strategies (e.g., Gaussian noise vs.\ mixup or no augmentation) induce substantial performance variation on the Pima dataset, with Gaussian noise consistently degrading performance.

Component similarity analysis on the Pima dataset further reveals strong redundancy in the configuration space. Feature selection variants (biMax–biMean)  exhibit low RMS distances ( 0.0252),  mixup     shows high similarity to no augmentation (RMS: 0.0279), and TomekLinks and no-imbalance are closely clustered (0.0325), indicating interchangeable behaviors.     In contrast, augmentation (Gaussian noise vs. no augmentation $\approx$ 0.10) and model families show higher divergence, reflecting their stronger functional impact.

Cross-component correlation analysis confirms substantial structural coupling across the pipeline space. On the Pima dataset, decision tree–based models (\texttt{DTModel}) exhibit moderate-to-high correlations ($\approx$ 0.4–0.9), with typical values around 0.68. For the Stroke dataset, \texttt{RandomUnderSampler} shows correlations ranging from 0.74 to 0.91, concentrated around 0.81, indicating relatively strong but more variable interactions compared to Pima.

These results indicate that performance is governed by a small subset of interacting components—primarily model choice, augmentation strategy, and imbalance handling—while many preprocessing configurations are functionally redundant, thereby guiding the design of future experiment configuration settings toward focused, interaction-aware optimization rather than independent tuning.

 \paragraph{Performance–Variability and Robustness.}  
 
Cross-seed analysis on the Pima dataset reveals a clear trade-off between predictive performance and robustness.  SVM achieves the highest mean Macro-F1 (0.689) but also exhibits the largest variability ($\sigma \approx 0.029$), indicating strong sensitivity to stochastic factors. In contrast, ensemble methods (XGBoost, Gradient Boosting, Random Forest) achieve slightly lower but competitive performance (0.645–0.659) with more stable variability ($\sigma \approx 0.023$–0.026), providing a better balance between accuracy and robustness. Decision Trees show the lowest variability ($\sigma \approx 0.012$) but reduced performance, while Logistic Regression occupies an intermediate regime between SVM and ensemble methods.

Friedman tests ($\chi^2 \approx 26$–28, $p \ll 0.05$) confirm statistically significant and consistent ranking differences across models, although robustness varies substantially. This observation reflects a trade-off between predictive performance and stability, where high-capacity models such as SVM achieve strong predictive performance but exhibit higher sensitivity to stochastic perturbations, while ensemble methods improve stability through aggregation. These results highlight the importance of multi-seed evaluation and indicate that reliable model selection should balance performance and stability rather than relying on single-seed outcomes.

 \paragraph{Optimal configuration design}

The experimental results reveal that optimal AutoML configuration design is strongly dataset-dependent, particularly in the context of medical prediction tasks.

For the Pima Indians Diabetes dataset, the removal of redundant components, including data augmentation and imbalance-handling strategies, reduces computational complexity while maintaining or slightly decreasing Macro-F1 performance in selected configurations. Furthermore, experimental analysis indicates that the split ratio parameter plays a critical role in model performance, with a split ratio of 0.1 consistently outperforming 0.2 under the same experimental setting. These results suggest that, for relatively low-dimensional and moderately balanced datasets, compact search spaces are sufficient to achieve competitive performance.

In contrast, the Stroke Prediction dataset benefits from progressive search-space enhancement. The inclusion of imbalance-handling techniques, particularly RandomOverSampler (ROS), leads to consistent improvements in Macro-F1 performance. The best-performing configuration (V7) achieves a Macro-F1 of 0.6766, improving from the baseline value of 0.6560. This demonstrates that, for highly imbalanced datasets, explicitly modeling class imbalance is more influential than increasing model or feature selection complexity.

 \paragraph{Search Space Structure, Sensitivity, and Implications for Future Experiment Design.}

This framework  search space exhibits two distinct regions: a redundant region with highly correlated, interchangeable configurations, and a sensitive region where key components (e.g., model choice and imbalance handling) significantly affect performance. This indicates that the effective dimensionality is much lower than the combinatorial space.

Practically, ensemble methods provide the best balance of performance and robustness, while imbalance handling is critical for minority-class prediction. Feature richness should be preserved, and excessive preprocessing offers limited gains compared to model and imbalance choices. Multi-seed evaluation is essential to ensure stability and generalizability.

Due to computational constraints, the current framework does not yet include neural networks, a wider range of augmentation and imbalance strategies, or more diverse feature selection methods. While strong results are achieved on the Pima dataset, the Stroke dataset remains challenging, with substantially lower Macro-F1 due to severe class imbalance and higher sensitivity.

 These findings provide clear guidance for future experiment design, emphasizing targeted expansion of high-impact components (e.g., imbalance handling and model diversity) and more efficient exploration of the reduced effective search space, thereby supporting more reliable deployment in dynamic real-world environments.

\subsection{Limitations and Future Work}

This study is limited to two datasets, and additional datasets are required to establish broader generalizability. Neural network models were not evaluated due to computational constraints, although future work may incorporate optimized architectures, particularly in data-rich settings.

Future work will focus on:
\begin{itemize}
    \item Extending the framework to multi-modal and unstructured data, including imaging, video, and text.
    \item Enabling multi-node parallel execution to scale pipeline evaluation across larger datasets.
    \item Exploring improved configuration strategies for challenging datasets (e.g., Stroke), with emphasis on imbalance handling.
    \item Conducting deployment-oriented evaluations to assess real-world clinical applicability.
\end{itemize}

These directions support the development of a scalable, reproducible, and structure-aware framework capable of efficiently identifying robust pipelines while maintaining transparency and interpretability across diverse biomedical datasets.

\section{Conclusion}
\label{sec:Conclusion}

This study introduces \texttt{yvsoucom-iterkit}, a deterministic and log-driven AutoML framework that fundamentally redefines pipeline optimization as a fully reproducible, configuration-level experimental system. Unlike conventional black-box AutoML approaches, the framework explicitly encodes each pipeline configuration as a traceable log entity, enabling complete transparency, fine-grained component-level analysis, and statistically grounded evaluation.

By exposing the full preprocessing–model configuration space, the framework enables exhaustive yet structured exploration of pipeline interactions across feature selection, augmentation, imbalance handling, and model design. This design transforms AutoML from a single-objective optimization process into a systematic experimental paradigm for analyzing component behavior and interactions.

Extensive evaluation over more than 18{,}000 pipeline configurations reveals a highly redundant search space, where many configurations converge to similar performance, indicating a substantially reduced effective dimensionality. Across both datasets, performance is primarily governed by a small subset of interacting components, particularly model selection, imbalance handling, and data augmentation, while most preprocessing operations contribute marginal gains.

Analysis of raw metric distributions and branch-level behavior reveals repetitive clustering patterns across pipelines, where multiple configurations produce near-identical outcomes, indicating a redundant configuration space with many-to-one mappings. The Pima dataset exhibits a more stable and low-sensitivity landscape, while the Stroke dataset demonstrates stronger boundary-constrained and imbalance-sensitive fluctuations, together suggesting functionally equivalent configurations and a reducible search space.

On the Pima dataset, top pipelines achieve Macro-F1 up to 0.8764, with ensemble models such as XGBoost showing strong and stable performance, while SVM exhibits higher variability across seeds ($\sigma \approx 0.029$). In contrast, the Stroke dataset achieves high Micro-F1 (0.9534–0.9537) and Weighted-F1 up to 0.9424 but substantially lower Macro-F1 (0.6511–0.6560), reflecting severe class imbalance effects.

Component similarity analysis on the Pima dataset reveals strong redundancy in the configuration space. Feature selection variants (biMax–biMean) exhibit low RMS distances (0.0252), mixup is highly similar to no augmentation (0.0279), and TomekLinks closely aligns with no imbalance handling (0.0325), indicating largely interchangeable behaviors. In contrast, augmentation strategies (e.g., Gaussian noise vs.\ no augmentation $\approx 0.10$) and model families show greater divergence, reflecting their stronger functional impact on performance.

Cross-seed analysis further confirms a clear performance–robustness trade-off, where ensemble methods achieve more stable variability ($\sigma \approx 0.023$–0.026) compared to SVM, which achieves higher peak performance but lower stability. Statistical testing (Friedman $\chi^2 \approx 26$–28, $p \ll 0.05$) validates significant and consistent model ranking differences across experimental runs.

Component-level analysis shows that augmentation (\texttt{AugMethod}), model choice, and imbalance handling are the dominant drivers of performance on the Pima dataset (Macro-F1 contributions 0.454, 0.198, and 0.101 respectively), while imbalance handling is most critical for the Stroke dataset (0.406). These results are directly enabled by the framework’s log-level traceability, which allows decomposition of performance into interpretable configuration components.

 The experimental results further demonstrate that optimal AutoML configuration design is strongly dataset-dependent. For the Pima Indians Diabetes dataset, computation performance is generally improved through search-space simplification, where redundant components such as data augmentation and imbalance-handling strategies are removed without significantly affecting Macro-F1. In particular, the split ratio parameter plays a dominant role, with a value of 0.1 consistently outperforming 0.2t.

In contrast, the Stroke Prediction dataset benefits from progressive search-space enhancement. The inclusion of imbalance-handling techniques, particularly RandomOverSampler (ROS), leads to consistent improvements in Macro-F1 performance, increasing from 0.6560 in the baseline configuration to 0.6766 in the optimized configuration (V7).

 Future work will extend the framework to neural architectures, multimodal datasets, and distributed execution, while enhancing imbalance-aware and interaction-aware optimization for real-world clinical applications. 
    
 \section*{Acknowledgements}

The authors declare that this work received no funding or financial support.  Hangzhou Domain Zones Technology Co. Ltd.   had no involvement in the study design, data collection, analysis, interpretation of results, or manuscript preparation and submission.

\section*{Data and Code Availability}

The datasets analyzed in this study are publicly available. 
 
The Pima Indians Diabetes dataset can be accessed from the UCI Machine Learning Repository via
\url{https://raw.githubusercontent.com/jbrownlee/Datasets/master/pima-indians-diabetes.data.csv}.
The dataset contains 768 records with 8 clinical predictor variables and one binary outcome variable indicating diabetes status. All instances correspond to female patients of Pima Indian heritage aged 21 years or older and include measurements such as glucose concentration, blood pressure, BMI, insulin level, diabetes pedigree function, and age.  

 The Stroke Prediction dataset is publicly available at 
\url{https://www.kaggle.com/datasets/fedesoriano/stroke-prediction-dataset}. 
The dataset file, \texttt{healthcare-dataset-stroke-data.csv}, contains 5,110 patient records with demographic, clinical, and lifestyle attributes, including age, hypertension status, heart disease history, average glucose level, BMI, smoking status, and stroke outcome labels.

Example scripts, including all pipeline configurations, sample data, and experiment logs required to reproduce the experiments, are publicly accessible at:
\url{https://github.com/yvsoucom/itekit-examples/releases/tag/v1.1}.

The full implementation of the \texttt{yvsoucom-iterkit} framework  will be made publicly available upon publication:
\url{https://github.com/yvsoucom/yvsoucom-iterkit}.

\section*{Statement for studies involving humans and animals}

This study does not involve any direct experimentation on humans or animals. The datasets used are publicly available and fully anonymized; therefore, no ethical approval and informed consent were required.

\section*{Declaration of Competing Interest}

One of the authors is affiliated with Hangzhou Domain Zones Technology Co. Ltd. The company had no role in the study design, data collection, analysis, interpretation of data, or in writing the manuscript. 

The authors declare that they have no other competing financial or non-financial interests, no additional support beyond their primary affiliations, and no other activities or relationships that could be perceived to have influenced the submitted work.

 \section*{Author Contributions}

\noindent
\textbf{Rui Huang:} Conceptualization, Methodology, Formal analysis, Validation, Investigation, Data curation, Visualization, Writing --- original draft, review, and editing.

\noindent
\textbf{Lican Huang:} Methodology, Software, Formal analysis, Investigation, Data curation, Writing --- original draft, review, and editing.

\noindent
All authors read and approved the final manuscript.

\section {Declaration of generative AI and AI-assisted technologies in the manuscript preparation process}
Statement: During the preparation of this work the authors used ChatGPT   in order to prepare and writing the draft paper. After using this tool, the authors reviewed and edited the content as needed and take full responsibility for the content of the published article.

\begin{appendices}

\onecolumn
\section*{Supplementary Materials Overview}
 
This document presents the \textbf{comprehensive set of experimental results, analyses, and visualizations} that support the findings reported in   the main manuscript.
It includes detailed performance metrics, branch-level trends, cross-component interactions, cross-seed variability, and value-level component similarity analyses for both the Pima and Stroke datasets.

The Supplementary Materials provide \textbf{tabular summaries, heatmaps, random-forest-based component importance analyses, correlation diagrams, and statistical comparisons}, offering a complete view of pipeline behavior, component contributions, and model robustness across multiple experimental configurations.

These results enable a \textbf{deeper understanding of pipeline design, cross-dataset comparisons, and the impact of preprocessing, data augmentation, imbalance handling, and model selection} on predictive performance.  
They also allow readers to trace the effects of individual pipeline components, examine branch-level variability, and evaluate model stability across different random seeds, providing a solid foundation for reproducible machine learning research.

All results are derived from the \textbf{logs generated by the \texttt{yvsoucom-iterkit} framework}, which can be traced and inspected on GitHub: \url{https://github.com/yvsoucom/itekit-examples}.  
The complete raw logs and experimental data are permanently archived on Zenodo: \url{https://zenodo.org/records/19476823}.

This PDF contains additional figures, tables, and analyses supporting the main manuscript.  Main figures and tables are reproduced here for completeness,.

\renewcommand{\thesubsection}{Supplementary \Alph{subsection}}   

\setcounter{figure}{0}
\setcounter{table}{0}

\renewcommand{\thefigure}{\thesubsection.\arabic{figure}}
\renewcommand{\thetable}{\thesubsection.\arabic{table}}

 \clearpage   
\subsection{Raw Metric Distributions and Branch-Level Trends}
\label{app:branch_trends}
\begin{figure*}[!htbp]
\centering
\includegraphics[width=0.85\linewidth]{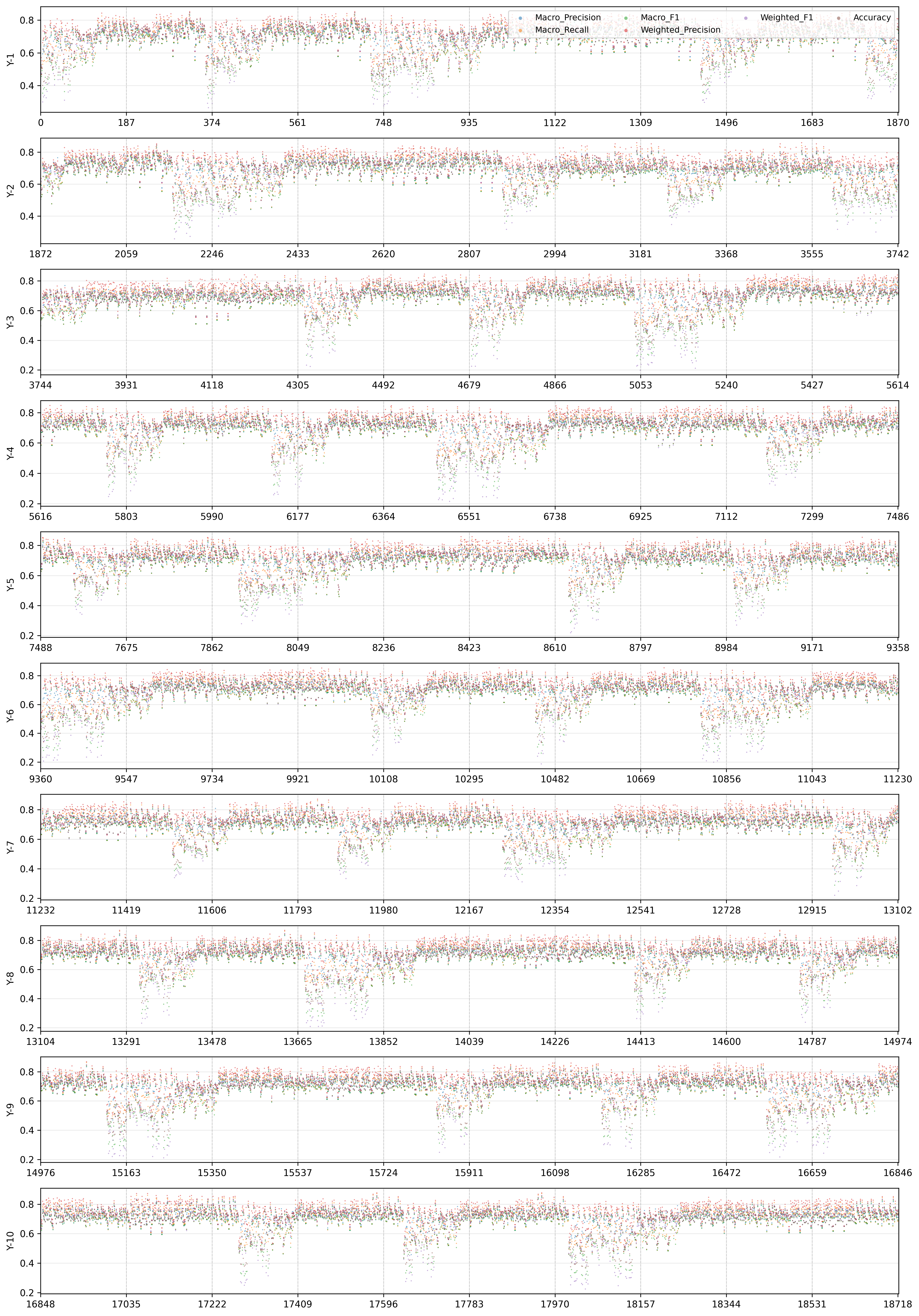}
\caption{Per-branch metric distribution for the Pima dataset.}
\label{app:pima_branch_trends}
\end{figure*}

\begin{figure*}[!htbp]
\centering
\includegraphics[width=0.85\linewidth]{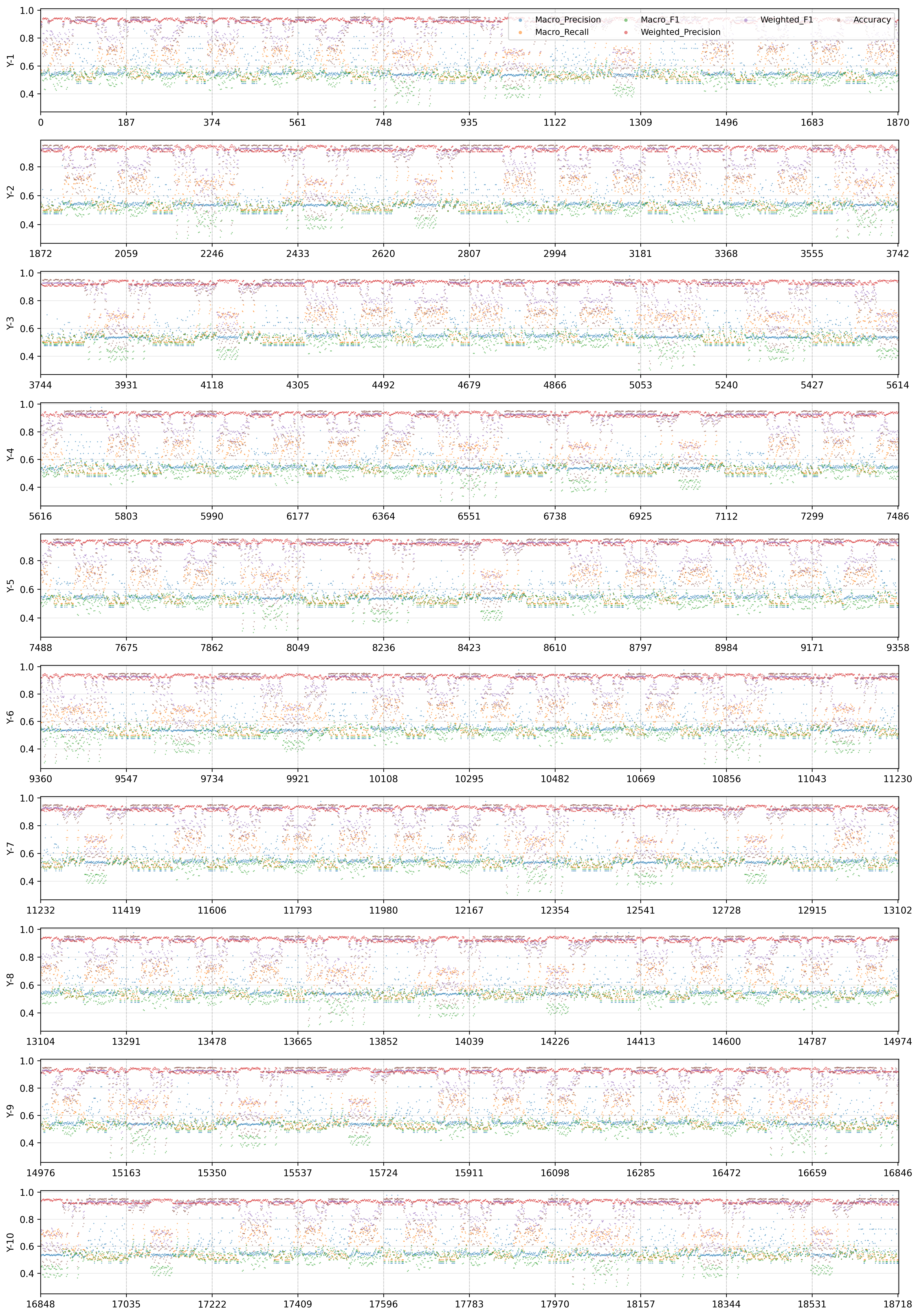}
\caption{Per-branch metric distribution for the Stroke dataset.}
\label{app:stroke_branch_trends}
\end{figure*}

 This subsection presents the \textbf{raw metric distributions for individual branches} of the Pima and Stroke datasets.  
It includes \textbf{top class-wise F1 scores} for the Pima dataset (Table~\ref{tab:pima_top5_class_wise_F1}) , \textbf{top class-wise F1 scores} for the Stroke dataset (Table~\ref{tab:top5_class_wise_F1_stroke}) and \textbf{visual distributions} of metrics across all branches (Figures~\ref{app:pima_branch_trends} and~\ref{app:stroke_branch_trends}).  

These results provide insight into branch-level performance variation, demonstrate the impact of different preprocessing, augmentation, and imbalance-handling configurations, and highlight the best-performing branches across multiple experimental setups.  
 
\paragraph{Abbreviations.}
{\small
\textit{C} = class label (0 = majority, 1 = minority);
\textit{Feat} = number of features;
\textit{FS} = feature selection (bMax = biMaxInfgain, bMean = biMeanInfgain, Inf = infgain);
\textit{Sc} = scaler (Std = Standard, MinMax = MinMax);
\textit{Aug} = augmentation (MX = mixup, GN = Gaussian noise, NA = none);
\textit{Imb} = imbalance handling (nIMB = none, Tomek = TomekLinks);
\textit{Norm} = normalization order (L = last, F = first);
\textit{Split} = train/test ratio;
\textit{Prob} = probability threshold;
\textit{Seed} = random seed;
\textit{Acc} = accuracy;
\textit{W-F1} = weighted F1;
\textit{M-F1} = macro F1.
}

\begin{table*}[htbp]
\centering
\small 
\caption{Top 5 Class-wise F1 Scores on Pima Dataset }
\centering
\footnotesize
\renewcommand{\arraystretch}{0.85}
\setlength{\tabcolsep}{2pt}
\begin{tabular}{l l l l l l l l l l l l l l l l}
\hline
F1 & C & RunID & Feat & FS & Sc & Aug & Imb & Model & Acc & W-F1 & M-F1 & Norm & Split & Prob & Seed \\
\hline
0.909 & 0 & 0754 & 4 & bMax & Std & MX & nIMB & SVM & 0.883 & 0.884 & 0.873 & L & 0.1 & 0.35 & 126 \\
0.909 & 0 & 0754 & 4 & bMean & Std & MX & nIMB & SVM & 0.883 & 0.884 & 0.873 & F & 0.1 & 0.35 & 126 \\
0.906 & 0 & 1919 & 5 & bMean & Std & MX & nIMB & RF & 0.870 & 0.866 & 0.849 & L & 0.1 & 0.50 & 7 \\
0.906 & 0 & 0754 & 4 & bMean & Std & MX & nIMB & GB & 0.870 & 0.866 & 0.849 & L & 0.1 & 0.50 & 126 \\
0.905 & 0 & 1411 & 6 & bMax & Std & NA & Tomek & XGB & 0.883 & 0.885 & 0.876 & L & 0.1 & 0.35 & 126 \\
\hline
\end{tabular}
\label{tab:pima_top5_class_wise_F1}
\end{table*}

\begin{table*}[htbp]
\centering
\small 
\caption{Top 5 Class-wise F1 Scores on Stroke Dataset }
\centering
\footnotesize
\renewcommand{\arraystretch}{0.85}
\setlength{\tabcolsep}{2pt}
\begin{tabular}{l l l l l l l l l l l l l l l}
\hline
F1 & C & RunID & Feat & FS & Sc & Aug & Imb & Model & Acc & W-F1 & M-F1 & Norm & Split & Prob \\
\hline
0.978 & 0 & 8f23 & 7 & Inf & Min & MX & Tomek & GB & 0.957 & 0.943 & 0.622 & T & 0.1 & 0.35 \\
0.977 & 0 & 8f23 & 9 & bMax & Min & GN & Tomek & RF & 0.955 & 0.936 & 0.563 & F & 0.1 & 0.50 \\
0.977 & 0 & 2736 & 8 & bMean & Std & GN & nIMB & LR & 0.955 & 0.936 & 0.563 & F & 0.1 & 0.35 \\
0.977 & 0 & 2736 & 9 & bMax & Min & GN & Tomek & XGB & 0.955 & 0.936 & 0.563 & T & 0.1 & 0.35 \\
0.977 & 0 & 8f23 & 8 & bMax & Std & MX & nIMB & XGB & 0.955 & 0.936 & 0.563 & F & 0.1 & 0.50 \\
\hline
\end{tabular}
\label{tab:top5_class_wise_F1_stroke}
\end{table*}

 \clearpage 

\subsection{Pipeline-Level and Aggregate Performance Analysis}
\label{app:ranking}

This subsection provides a detailed overview of the \textbf{performance of different pipeline configurations} evaluated on the Stroke and Pima datasets.  
We present both \textbf{top-ranked pipelines for each evaluation metric} (Table ~\ref{tab:pima_rank_metric_vertical})and ~\ref{tab:stroke_rank_metric_vertical}) and \textbf{aggregate performance statistics} across all pipelines and branches (Tables~\ref{tab:pima_aggregate} and~\ref{tab:stroke_aggregate}).  

The top-ranked pipelines highlight which combinations of preprocessing, feature selection, normalization, augmentation, imbalance-handling, and model selection yield the best performance under metrics such as Accuracy, Macro/Micro/Weighted Precision, Recall, F1, and the Integrated Score.  
By analyzing these pipelines, readers can identify \textbf{patterns in high-performing configurations}, including how normalization order, probability thresholds, and train-test splits influence predictive outcomes.  

The aggregate statistics complement the top-ranked pipelines by providing \textbf{overall trends and variability} across all configurations.  
This includes mean and standard deviation for each metric, allowing for assessment of pipeline robustness and the stability of performance across branches and experimental settings.  

Together, these results provide a comprehensive view of pipeline behavior, enabling a \textbf{deep understanding of design choices, cross-dataset comparisons, and the impact of different preprocessing, augmentation, and modeling strategies on predictive performance}.

\begin{table*}[htbp]
\caption{Top-ranked pipelines under different evaluation metrics on the Pima Indians Diabetes dataset. P\_th represents probability threshold. M\_ for Macro, m\_ for Micro, W\_ for Weighted metrics, , R\_S for Random Seed.}
\label{tab:pima_rank_metric_vertical}
\centering
 
\footnotesize
\renewcommand{\arraystretch}{0.85}
\setlength{\tabcolsep}{2pt}

\begin{tabular}{|c|l|c|>{\ttfamily\footnotesize}p{8.5cm}|c|c|c|}
\hline
Rank & Metric & Value & LogDir & Split\_Ratio & P\_th &R\_S\\
\hline
\multirow{10}{*}{1} & M\_P & 0.9032 & \texttt{4/infgain/gaussian\_noise\_\_noImbl\_\_standard/sklearn\_SVM} & 0.1 & 0.35  & 23 \\ 
 & M\_R & 0.8930 & \texttt{6/biMeanInfgain/noAug\_\_TomekLinks\_\_standard/XGBmodel} & 0.1 & 0.35  & 126 \\ 
 & M\_F1 & 0.8764 & \texttt{6/biMaxInfgain/noAug\_\_TomekLinks\_\_standard/XGBmodel} & 0.1 & 0.35  & 126 \\ 
 & m\_P & 0.8831 & \texttt{6/biMaxInfgain/noAug\_\_TomekLinks\_\_standard/XGBmodel} & 0.1 & 0.35  & 126 \\ 
 & m\_R & 0.8831 & \texttt{6/biMaxInfgain/noAug\_\_TomekLinks\_\_standard/XGBmodel} & 0.1 & 0.35  & 126 \\ 
 & m\_F1 & 0.8831 & \texttt{4/biMeanInfgain/standard\_\_mixup\_\_noImbl/sklearn\_SVM} & 0.1 & 0.35  & 126 \\ 
 & W\_P & 0.8944 & \texttt{6/biMaxInfgain/noAug\_\_TomekLinks\_\_standard/XGBmodel} & 0.1 & 0.35  & 126 \\ 
 & W\_R & 0.8831 & \texttt{6/biMaxInfgain/noAug\_\_TomekLinks\_\_standard/XGBmodel} & 0.1 & 0.35  & 126 \\ 
 & W\_F1 & 0.8850 & \texttt{6/biMeanInfgain/noAug\_\_TomekLinks\_\_standard/XGBmodel} & 0.1 & 0.35  & 126 \\ 
 & Acc & 0.8831 & \texttt{6/biMaxInfgain/noAug\_\_TomekLinks\_\_standard/XGBmodel} & 0.1 & 0.35  & 126 \\ 
\hline
\multirow{10}{*}{2} & M\_P & 0.8906 & \texttt{6/biMeanInfgain/standard\_\_gaussian\_noise\_\_noImbl/LogisticRegression} & 0.1 & 0.5  & 7 \\ 
 & M\_R & 0.8930 & \texttt{6/biMaxInfgain/noAug\_\_TomekLinks\_\_standard/XGBmodel} & 0.1 & 0.35  & 126 \\ 
 & M\_F1 & 0.8764 & \texttt{6/biMeanInfgain/noAug\_\_TomekLinks\_\_standard/XGBmodel} & 0.1 & 0.35  & 126 \\ 
 & m\_P & 0.8831 & \texttt{4/biMaxInfgain/mixup\_\_noImbl\_\_standard/sklearn\_SVM} & 0.1 & 0.35  & 126 \\ 
 & m\_R & 0.8831 & \texttt{4/biMaxInfgain/mixup\_\_noImbl\_\_standard/sklearn\_SVM} & 0.1 & 0.35  & 126 \\ 
 & m\_F1 & 0.8831 & \texttt{6/biMaxInfgain/noAug\_\_TomekLinks\_\_standard/XGBmodel} & 0.1 & 0.35  & 126 \\ 
 & W\_P & 0.8944 & \texttt{6/biMeanInfgain/noAug\_\_TomekLinks\_\_standard/XGBmodel} & 0.1 & 0.35  & 126 \\ 
 & W\_R & 0.8831 & \texttt{4/biMaxInfgain/mixup\_\_noImbl\_\_standard/sklearn\_SVM} & 0.1 & 0.35  & 126 \\ 
 & W\_F1 & 0.8850 & \texttt{6/biMaxInfgain/noAug\_\_TomekLinks\_\_standard/XGBmodel} & 0.1 & 0.35  & 126 \\ 
 & Acc & 0.8831 & \texttt{4/biMaxInfgain/mixup\_\_noImbl\_\_standard/sklearn\_SVM} & 0.1 & 0.35  & 126 \\ 
\hline
\multirow{10}{*}{3} & M\_P & 0.8906 & \texttt{5/biMeanInfgain/minmax\_\_gaussian\_noise\_\_noImbl/sklearn\_SVM} & 0.1 & 0.5  & 126 \\ 
 & M\_R & 0.8830 & \texttt{5/biMaxInfgain/noAug\_\_noImbl\_\_standard/XGBmodel} & 0.1 & 0.35  & 126 \\ 
 & M\_F1 & 0.8727 & \texttt{4/biMaxInfgain/mixup\_\_noImbl\_\_standard/sklearn\_SVM} & 0.1 & 0.35  & 126 \\ 
 & m\_P & 0.8831 & \texttt{6/biMeanInfgain/noAug\_\_TomekLinks\_\_standard/XGBmodel} & 0.1 & 0.35  & 126 \\ 
 & m\_R & 0.8831 & \texttt{6/biMeanInfgain/noAug\_\_TomekLinks\_\_standard/XGBmodel} & 0.1 & 0.35  & 126 \\ 
 & m\_F1 & 0.8831 & \texttt{4/biMaxInfgain/mixup\_\_noImbl\_\_standard/sklearn\_SVM} & 0.1 & 0.35  & 126 \\ 
 & W\_P & 0.8868 & \texttt{7/biMaxInfgain/standard\_\_mixup\_\_TomekLinks/random\_forest} & 0.1 & 0.35  & 126 \\ 
 & W\_R & 0.8831 & \texttt{6/biMeanInfgain/noAug\_\_TomekLinks\_\_standard/XGBmodel} & 0.1 & 0.35  & 126 \\ 
 & W\_F1 & 0.8836 & \texttt{4/biMeanInfgain/standard\_\_mixup\_\_noImbl/sklearn\_SVM} & 0.1 & 0.35  & 126 \\ 
 & Acc & 0.8831 & \texttt{6/biMeanInfgain/noAug\_\_TomekLinks\_\_standard/XGBmodel} & 0.1 & 0.35  & 126 \\ 
\hline
\multirow{10}{*}{4} & M\_P & 0.8846 & \texttt{6/infgain/standard\_\_gaussian\_noise\_\_noImbl/LogisticRegression} & 0.1 & 0.5  & 126 \\ 
 & M\_R & 0.8830 & \texttt{5/biMeanInfgain/noAug\_\_noImbl\_\_standard/XGBmodel} & 0.1 & 0.35  & 126 \\ 
 & M\_F1 & 0.8727 & \texttt{4/biMeanInfgain/standard\_\_mixup\_\_noImbl/sklearn\_SVM} & 0.1 & 0.35  & 126 \\ 
 & m\_P & 0.8831 & \texttt{4/biMeanInfgain/standard\_\_mixup\_\_noImbl/sklearn\_SVM} & 0.1 & 0.35  & 126 \\ 
 & m\_R & 0.8831 & \texttt{4/biMeanInfgain/standard\_\_mixup\_\_noImbl/sklearn\_SVM} & 0.1 & 0.35  & 126 \\ 
 & m\_F1 & 0.8831 & \texttt{6/biMeanInfgain/noAug\_\_TomekLinks\_\_standard/XGBmodel} & 0.1 & 0.35  & 126 \\ 
 & W\_P & 0.8868 & \texttt{7/infgain/standard\_\_noAug\_\_TomekLinks/XGBmodel} & 0.1 & 0.35  & 126 \\ 
 & W\_R & 0.8831 & \texttt{4/biMeanInfgain/standard\_\_mixup\_\_noImbl/sklearn\_SVM} & 0.1 & 0.35  & 126 \\ 
 & W\_F1 & 0.8836 & \texttt{4/biMaxInfgain/mixup\_\_noImbl\_\_standard/sklearn\_SVM} & 0.1 & 0.35  & 126 \\ 
 & Acc & 0.8831 & \texttt{4/biMeanInfgain/standard\_\_mixup\_\_noImbl/sklearn\_SVM} & 0.1 & 0.35  & 126 \\ 
\hline
\multirow{10}{*}{5} & M\_P & 0.8846 & \texttt{5/biMeanInfgain/gaussian\_noise\_\_noImbl\_\_standard/LogisticRegression} & 0.1 & 0.5  & 7 \\ 
 & M\_R & 0.8830 & \texttt{5/biMeanInfgain/standard\_\_noAug\_\_noImbl/XGBmodel} & 0.1 & 0.35  & 126 \\ 
 & M\_F1 & 0.8635 & \texttt{5/biMaxInfgain/noAug\_\_noImbl\_\_standard/XGBmodel} & 0.1 & 0.35  & 126 \\ 
 & m\_P & 0.8701 & \texttt{4/biMaxInfgain/standard\_\_noAug\_\_noImbl/XGBmodel} & 0.1 & 0.5  & 126 \\ 
 & m\_R & 0.8701 & \texttt{4/biMaxInfgain/standard\_\_noAug\_\_noImbl/XGBmodel} & 0.1 & 0.5  & 126 \\ 
 & m\_F1 & 0.8701 & \texttt{4/biMeanInfgain/minmax\_\_mixup\_\_noImbl/sklearn\_SVM} & 0.1 & 0.5  & 126 \\ 
 & W\_P & 0.8855 & \texttt{5/biMeanInfgain/noAug\_\_noImbl\_\_standard/XGBmodel} & 0.1 & 0.35  & 126 \\ 
 & W\_R & 0.8701 & \texttt{4/biMaxInfgain/standard\_\_noAug\_\_noImbl/XGBmodel} & 0.1 & 0.5  & 126 \\ 
 & W\_F1 & 0.8725 & \texttt{5/biMaxInfgain/standard\_\_noAug\_\_noImbl/XGBmodel} & 0.1 & 0.35  & 126 \\ 
 & Acc & 0.8701 & \texttt{4/biMaxInfgain/standard\_\_noAug\_\_noImbl/XGBmodel} & 0.1 & 0.5  & 126 \\ 
\hline
\end{tabular}
\end{table*}

\begin{table*}[htbp]
\caption{Top-ranked pipelines under different evaluation metrics on the Stroke prediction dataset. P\_th represents probability threshold. M\_ for Macro, m\_ for Micro, W\_ for Weighted metrics }
\label{tab:stroke_rank_metric_vertical}
\centering
 
\footnotesize
\renewcommand{\arraystretch}{0.85}
\setlength{\tabcolsep}{2pt}
 \begin{tabular}{|c|l|c|>{\ttfamily\footnotesize}p{8.2cm}|c|c|c|}
\hline
Rank & Metric & Value & LogDir &  Norm\_First & Split\_Ratio & P\_th  \\
\hline
\multirow{10}{*}{1} & M\_P & 0.9765 & \texttt{6/infgain/noAUG\_\_noIMB\_\_Standard/XGBmodel} & False & 0.1 & 0.5 \\
 & M\_R & 0.7800 & \texttt{8/infgain/Standard\_\_mixup\_\_SMOTE/LogisticRegression} & True & 0.2 & 0.5 \\
 & M\_F1 & 0.6560 & \texttt{9/biMaxInfgain/noAUG\_\_SMOTE\_\_MinMax/LogisticRegression} & False & 0.2 & 0.5 \\
 & m\_P & 0.9537 & \texttt{8/infgain/MinMax\_\_mixup\_\_noIMB/GradientBoosting} & True & 0.1 & 0.35 \\
 & m\_R & 0.9537 & \texttt{8/infgain/MinMax\_\_mixup\_\_noIMB/GradientBoosting} & True & 0.1 & 0.35 \\
 & m\_F1 & 0.9537 & \texttt{8/infgain/MinMax\_\_mixup\_\_noIMB/GradientBoosting} & True & 0.1 & 0.35 \\
 & W\_P & 0.9552 & \texttt{10/noSelect/noAUG\_\_TomekLinks\_\_Standard/XGBmodel} & False & 0.1 & 0.5 \\
 & W\_R & 0.9537 & \texttt{8/infgain/MinMax\_\_mixup\_\_noIMB/GradientBoosting} & True & 0.1 & 0.35 \\
 & W\_F1 & 0.9424 & \texttt{10/noSelect/noAUG\_\_noIMB\_\_MinMax/XGBmodel} & False & 0.1 & 0.35 \\
 & Acc & 0.9537 & \texttt{8/infgain/MinMax\_\_mixup\_\_noIMB/GradientBoosting} & True & 0.1 & 0.35 \\
\hline
\multirow{10}{*}{2} & M\_P & 0.9765 & \texttt{9/infgain/MinMax\_\_noAUG\_\_TomekLinks/LogisticRegression} & True & 0.2 & 0.5 \\
 & M\_R & 0.7792 & \texttt{7/infgain/Standard\_\_mixup\_\_RandomUnderSampler/LogisticRegression} & True & 0.2 & 0.5 \\
 & M\_F1 & 0.6560 & \texttt{9/biMaxInfgain/noAUG\_\_SMOTE\_\_Standard/LogisticRegression} & False & 0.2 & 0.5 \\
 & m\_P & 0.9534 & \texttt{9/infgain/MinMax\_\_mixup\_\_noIMB/LogisticRegression} & True & 0.2 & 0.35 \\
 & m\_R & 0.9534 & \texttt{9/infgain/MinMax\_\_mixup\_\_noIMB/LogisticRegression} & True & 0.2 & 0.35 \\
 & m\_F1 & 0.9534 & \texttt{7/infgain/mixup\_\_noIMB\_\_MinMax/LogisticRegression} & False & 0.2 & 0.35 \\
 & W\_P & 0.9552 & \texttt{9/infgain/MinMax\_\_noAUG\_\_TomekLinks/LogisticRegression} & True & 0.2 & 0.5 \\
 & W\_R & 0.9534 & \texttt{7/infgain/mixup\_\_noIMB\_\_MinMax/LogisticRegression} & False & 0.2 & 0.35 \\
 & W\_F1 & 0.9424 & \texttt{10/noSelect/MinMax\_\_noAUG\_\_noIMB/XGBmodel} & True & 0.1 & 0.35 \\
 & Acc & 0.9534 & \texttt{9/infgain/MinMax\_\_mixup\_\_noIMB/LogisticRegression} & True & 0.2 & 0.35 \\
\hline
\multirow{10}{*}{3} & M\_P & 0.9765 & \texttt{6/infgain/MinMax\_\_noAUG\_\_noIMB/XGBmodel} & True & 0.1 & 0.5 \\
 & M\_R & 0.7786 & \texttt{6/infgain/MinMax\_\_mixup\_\_SMOTE/LogisticRegression} & True & 0.2 & 0.5 \\
 & M\_F1 & 0.6560 & \texttt{9/biMeanInfgain/noAUG\_\_SMOTE\_\_MinMax/LogisticRegression} & False & 0.2 & 0.5 \\
 & m\_P & 0.9534 & \texttt{7/infgain/mixup\_\_noIMB\_\_MinMax/LogisticRegression} & False & 0.2 & 0.35 \\
 & m\_R & 0.9534 & \texttt{7/infgain/mixup\_\_noIMB\_\_MinMax/LogisticRegression} & False & 0.2 & 0.35 \\
 & m\_F1 & 0.9534 & \texttt{6/infgain/mixup\_\_TomekLinks\_\_Standard/LogisticRegression} & False & 0.2 & 0.35 \\
 & W\_P & 0.9552 & \texttt{6/infgain/noAUG\_\_noIMB\_\_Standard/XGBmodel} & False & 0.1 & 0.5 \\
 & W\_R & 0.9534 & \texttt{7/infgain/mixup\_\_TomekLinks\_\_Standard/LogisticRegression} & False & 0.2 & 0.35 \\
 & W\_F1 & 0.9424 & \texttt{10/noSelect/Standard\_\_noAUG\_\_noIMB/XGBmodel} & True & 0.1 & 0.35 \\
 & Acc & 0.9534 & \texttt{7/infgain/mixup\_\_noIMB\_\_MinMax/LogisticRegression} & False & 0.2 & 0.35 \\
\hline
\multirow{10}{*}{4} & M\_P & 0.9765 & \texttt{9/infgain/Standard\_\_noAUG\_\_TomekLinks/LogisticRegression} & True & 0.2 & 0.5 \\
 & M\_R & 0.7779 & \texttt{7/infgain/Standard\_\_mixup\_\_SMOTE/LogisticRegression} & True & 0.2 & 0.5 \\
 & M\_F1 & 0.6560 & \texttt{9/biMeanInfgain/noAUG\_\_SMOTE\_\_Standard/LogisticRegression} & False & 0.2 & 0.5 \\
 & m\_P & 0.9534 & \texttt{6/infgain/mixup\_\_TomekLinks\_\_Standard/LogisticRegression} & False & 0.2 & 0.35 \\
 & m\_R & 0.9534 & \texttt{6/infgain/mixup\_\_TomekLinks\_\_Standard/LogisticRegression} & False & 0.2 & 0.35 \\
 & m\_F1 & 0.9534 & \texttt{9/infgain/MinMax\_\_mixup\_\_noIMB/LogisticRegression} & True & 0.2 & 0.35 \\
 & W\_P & 0.9552 & \texttt{6/infgain/noAUG\_\_noIMB\_\_MinMax/XGBmodel} & False & 0.1 & 0.5 \\
 & W\_R & 0.9534 & \texttt{9/infgain/MinMax\_\_mixup\_\_noIMB/LogisticRegression} & True & 0.2 & 0.35 \\
 & W\_F1 & 0.9424 & \texttt{10/noSelect/noAUG\_\_noIMB\_\_Standard/XGBmodel} & False & 0.1 & 0.35 \\
 & Acc & 0.9534 & \texttt{6/infgain/mixup\_\_TomekLinks\_\_Standard/LogisticRegression} & False & 0.2 & 0.35 \\
\hline
\multirow{10}{*}{5} & M\_P & 0.9765 & \texttt{6/infgain/Standard\_\_noAUG\_\_noIMB/XGBmodel} & True & 0.1 & 0.5 \\
 & M\_R & 0.7772 & \texttt{6/infgain/MinMax\_\_mixup\_\_ADASYN/LogisticRegression} & True & 0.2 & 0.5 \\
 & M\_F1 & 0.6511 & \texttt{9/biMeanInfgain/noAUG\_\_ADASYN\_\_Standard/LogisticRegression} & False & 0.2 & 0.5 \\
 & m\_P & 0.9534 & \texttt{7/infgain/mixup\_\_TomekLinks\_\_Standard/LogisticRegression} & False & 0.2 & 0.35 \\
 & m\_R & 0.9534 & \texttt{7/infgain/mixup\_\_TomekLinks\_\_Standard/LogisticRegression} & False & 0.2 & 0.35 \\
 & m\_F1 & 0.9534 & \texttt{7/infgain/mixup\_\_TomekLinks\_\_Standard/LogisticRegression} & False & 0.2 & 0.35 \\
 & W\_P & 0.9552 & \texttt{6/infgain/MinMax\_\_noAUG\_\_noIMB/XGBmodel} & True & 0.1 & 0.5 \\
 & W\_R & 0.9534 & \texttt{6/infgain/mixup\_\_TomekLinks\_\_Standard/LogisticRegression} & False & 0.2 & 0.35 \\
 & W\_F1 & 0.9399 & \texttt{8/infgain/MinMax\_\_mixup\_\_noIMB/GradientBoosting} & True & 0.1 & 0.35 \\
 & Acc & 0.9534 & \texttt{7/infgain/mixup\_\_TomekLinks\_\_Standard/LogisticRegression} & False & 0.2 & 0.35 \\
\hline
\end{tabular}
\end{table*}

 \begin{table*}[htbp]
\centering
\caption{Aggregate performance statistics across all pipeline configurations on the Pima dataset. Results are reported as mean and standard deviation over all branches.}
\label{tab:pima_aggregate}
 
\small
\setlength{\tabcolsep}{4pt}
\begin{tabular}{|c|c|c|}
\hline
Metric & Mean & Std \\
\hline
Accuracy & 0.673649383024383 & 0.0897363795392836 \\
Macro\_Precision & 0.6838601341830406 & 0.0691823082366285 \\
Macro\_Recall & 0.6826892501253036 & 0.0755243216727676 \\
Macro\_F1 & 0.6550064165406746 & 0.0919794462164404 \\
Weighted\_Precision & 0.7246319526049205 & 0.0672222636465707 \\
Weighted\_Recall & 0.673649383024383 & 0.0897363795392836 \\
Weighted\_F1 & 0.6706759109693264 & 0.1011450473334413 \\
Micro\_Precision & 0.673649383024383 & 0.0897363795392836 \\
Micro\_Recall & 0.673649383024383 & 0.0897363795392836 \\
Micro\_F1 & 0.673649383024383 & 0.0897363795392836 \\
Integrated\_Score & 0.6774302360275797 & 0.0818401652216246 \\
\hline
\end{tabular}
\end{table*}

\begin{table*}[htbp]
\centering
\caption{Aggregate performance statistics across all pipeline configurations on the Stroke dataset. Results are reported as mean and standard deviation over all branches.}
\label{tab:stroke_aggregate}
\small
\setlength{\tabcolsep}{4pt}
\begin{tabular}{|c|c|c|}
\hline
Metric & Mean & Std \\
\hline
Accuracy & 0.818058674265866 & 0.1450322573703264 \\
Macro\_Precision & 0.5589756630812349 & 0.0599054344296001 \\
Macro\_Recall & 0.6028785619863295 & 0.0844300198038256 \\
Macro\_F1 & 0.5116298430269283 & 0.048158774525587 \\
Weighted\_Precision & 0.9255530374288182 & 0.0124443266608974 \\
Weighted\_Recall & 0.818058674265866 & 0.1450322573703265 \\
Weighted\_F1 & 0.8510809836508719 & 0.0963836572731272 \\
Micro\_Precision & 0.818058674265866 & 0.1450322573703264 \\
Micro\_Recall & 0.818058674265866 & 0.1450322573703264 \\
Micro\_F1 & 0.818058674265866 & 0.1450322573703264 \\
Integrated\_Score & 0.7403708199576546 & 0.0726826444494676 \\
\hline
\end{tabular}
\end{table*}

      \clearpage 
 \subsection{Pipeline Component Analysis: RF-Based Importance and Component-Specific Analysis}
\label{app:component}

This subsection presents an analysis of \textbf{individual pipeline components} and their contributions to predictive performance, using Random Forest-based importance scores and aggregated metric heatmaps.

Figure~\ref{fig:pima_unified_per_value_mega_heatmap} provides component-wise variability heatmaps, highlighting trends across branches, pipeline variations, and metric types for the Pima dataset.
Figure~\ref{fig:pima_rf_importance} summarizes component-level importance for the Pima dataset, showing which preprocessing, augmentation, and normalization steps most strongly influence predictive outcomes.
 
Figure~\ref{fig:pima_rf_importance_detail} shows the mean importance values of detailed component-level items for the Pima dataset .

\begin{figure*} [!t]
  \centering
 
    \includegraphics[width=\textwidth,height=0.95\textheight,keepaspectratio]{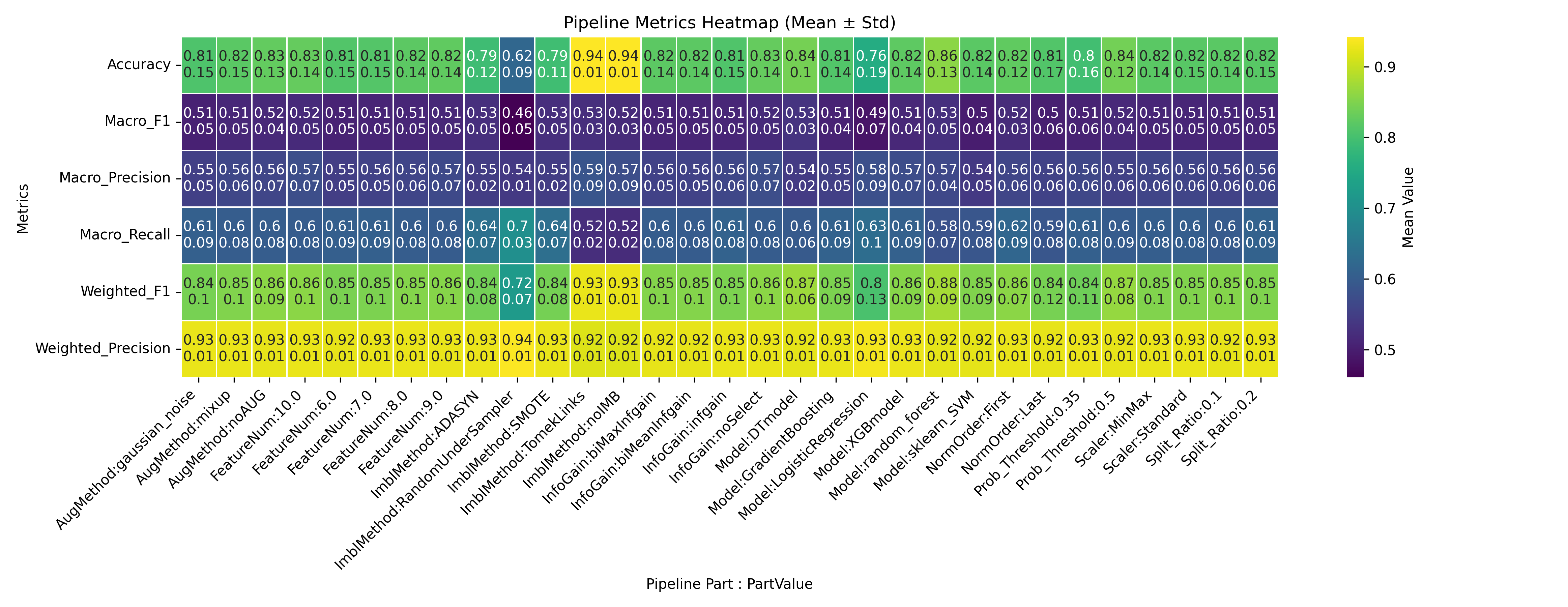}

    \caption{stroke unified metric mega heat map for branches}
              \label{fig:stroke_unified_per_value_mega_heatmap}

\end{figure*}

\begin{figure*}[!t]
\centering
\includegraphics[width=\textwidth]{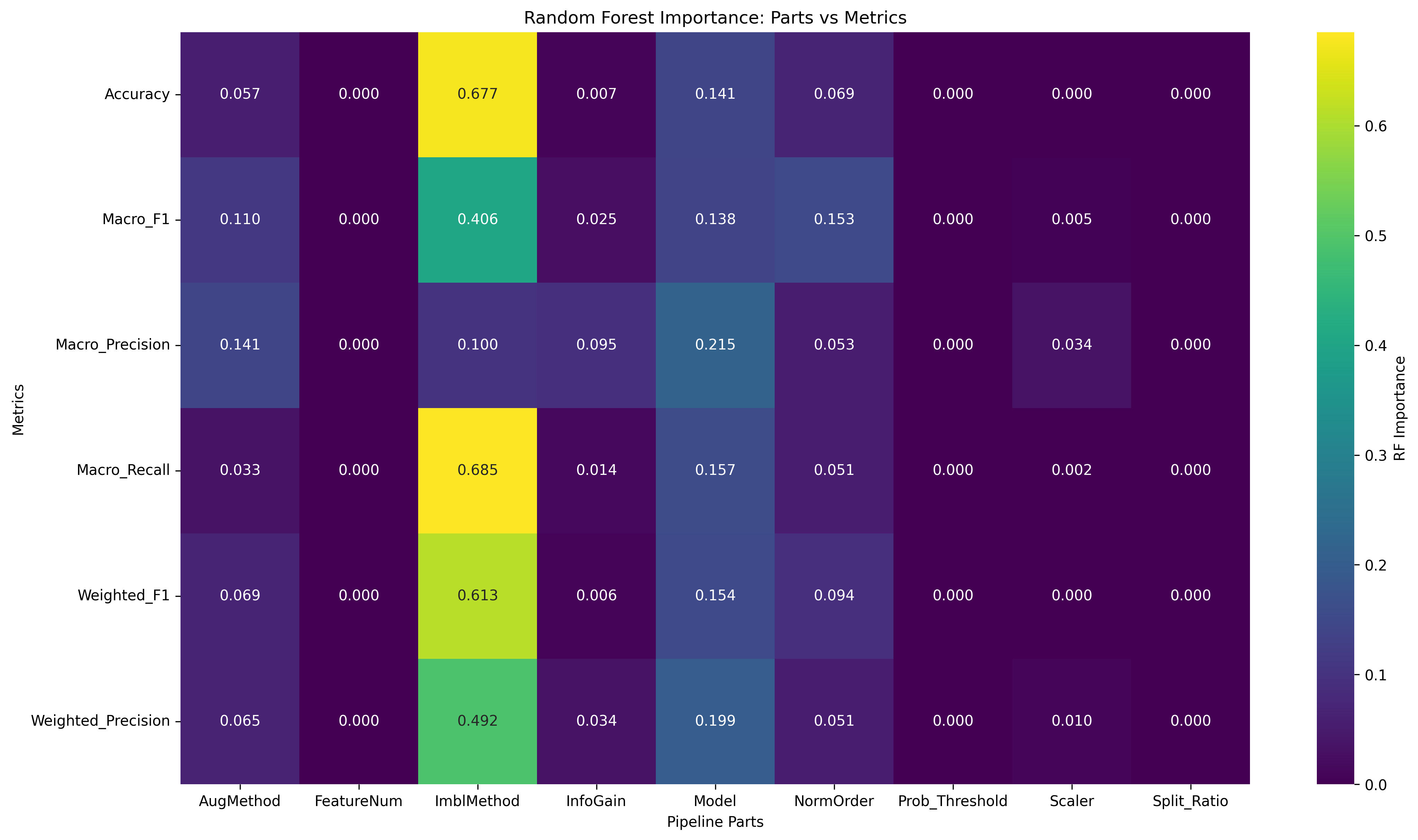}
\caption{Random Forest-based component importance for the Stroke dataset.}
\label{fig:stroke_rf_importance}
\end{figure*}

\begin{figure*}[!t]
\centering
\includegraphics[width=\textwidth]{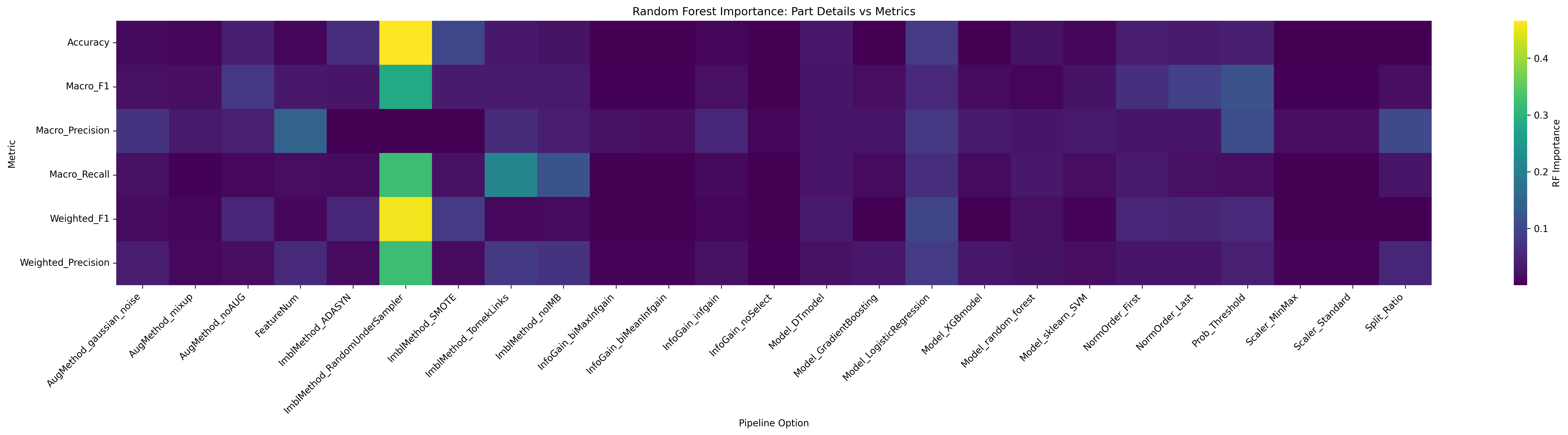}
\caption{Detailed Random Forest importance analysis for the Stroke dataset.}
\label{fig:stroke_rf_importance_detail}
\end{figure*}

Figure~\ref{fig:stroke_unified_per_value_mega_heatmap} provides component-wise variability heatmaps, highlighting trends across branches, pipeline variations, and metric types for the Stroke dataset.
Figure~\ref{fig:stroke_rf_importance} summarizes component-level importance for the Stroke dataset, showing which preprocessing, augmentation, and normalization steps most strongly influence predictive outcomes.
Figure~\ref{fig:stroke_rf_importance_detail} shows the mean importance values of detailed component-level items  for the Stroke dataset.

Together, these visualizations facilitate the identification of the most influential pipeline strategies and allow readers to observe how component choices interact with branch-specific performance, providing a comprehensive view of pipeline behavior and component contributions across datasets.

 \begin{figure*} [!t]
  \centering
 
    \includegraphics[width=\textwidth,height=0.95\textheight,keepaspectratio]{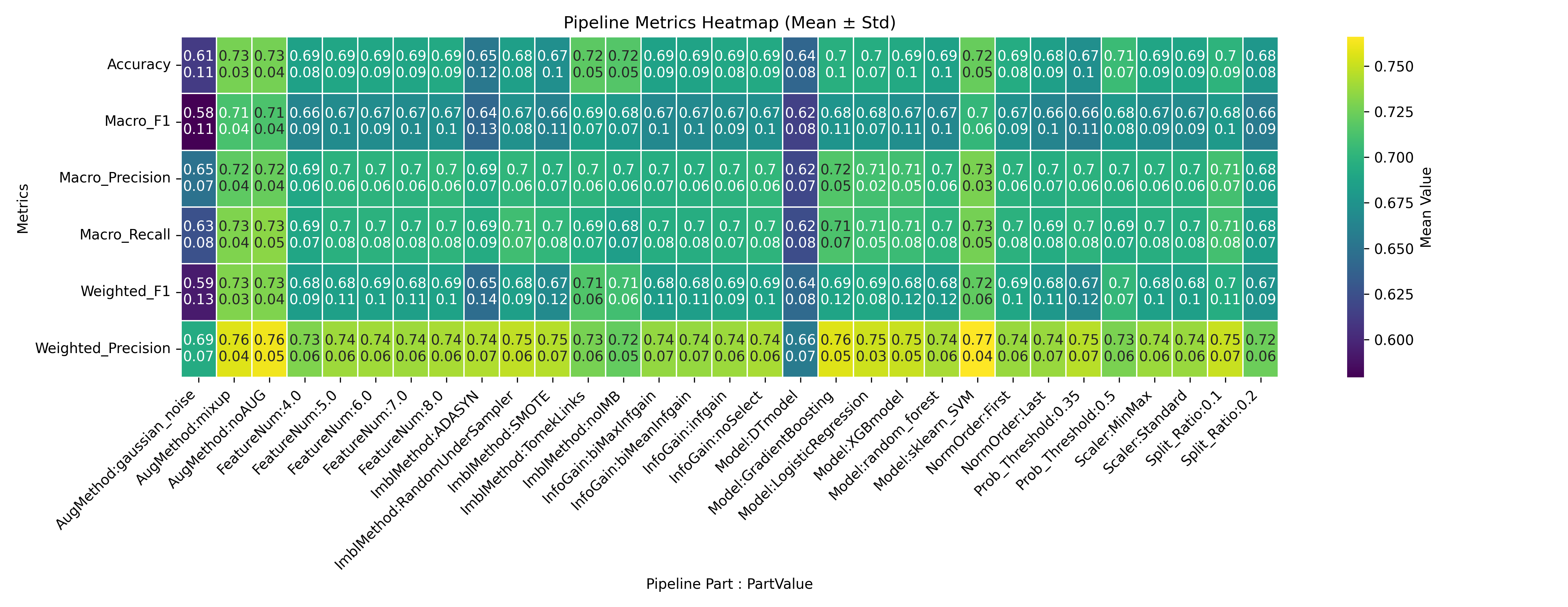}

    \caption{pima unified metric mega heat map for branches}
              \label{fig:pima_unified_per_value_mega_heatmap}

\end{figure*}

\begin{figure*}[!t]
\centering
\includegraphics[width=\textwidth]{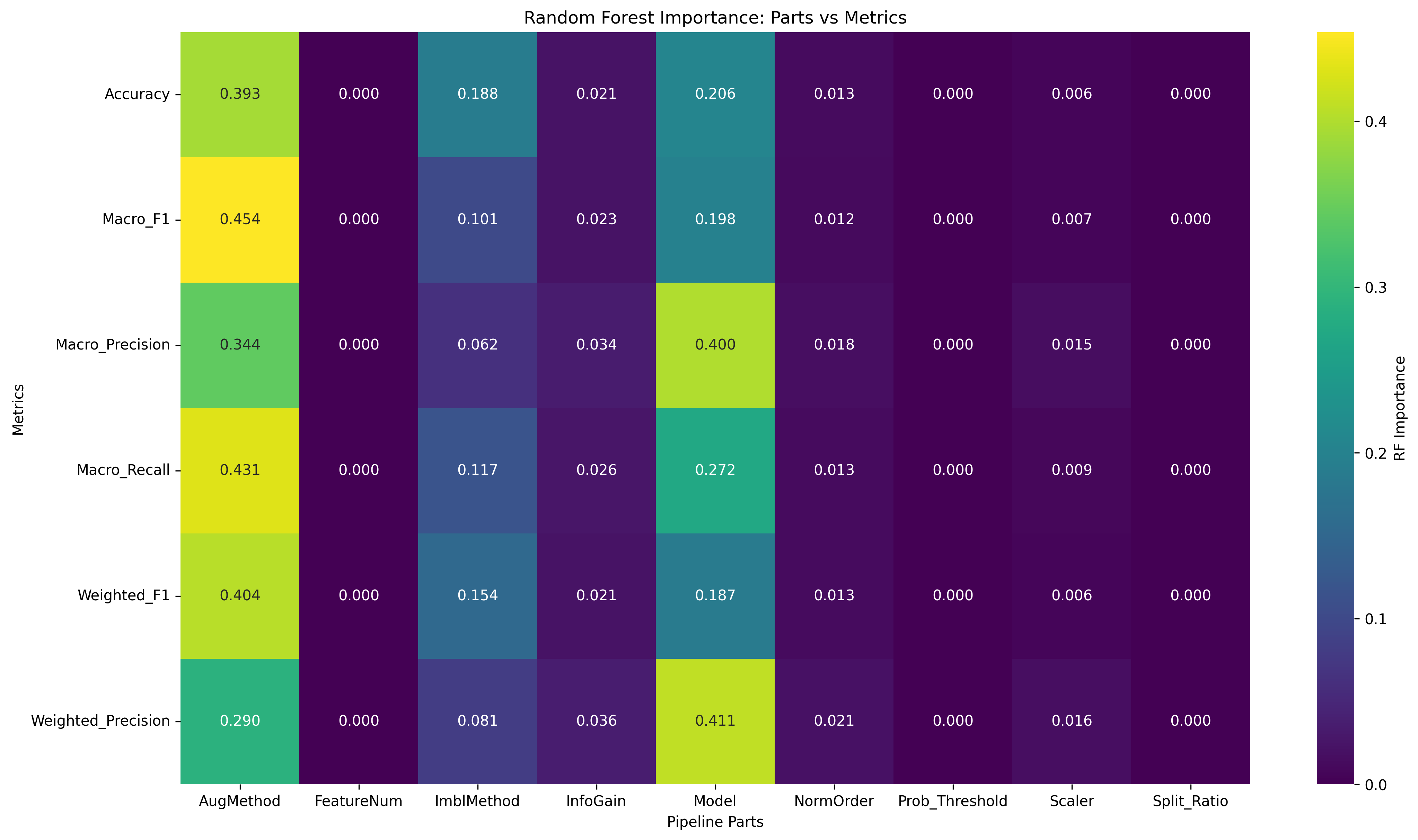}
\caption{Random Forest-based component importance for the Pima dataset.}
\label{fig:pima_rf_importance}
\end{figure*}

\begin{figure*}[!t]
\centering
\includegraphics[width=\textwidth]{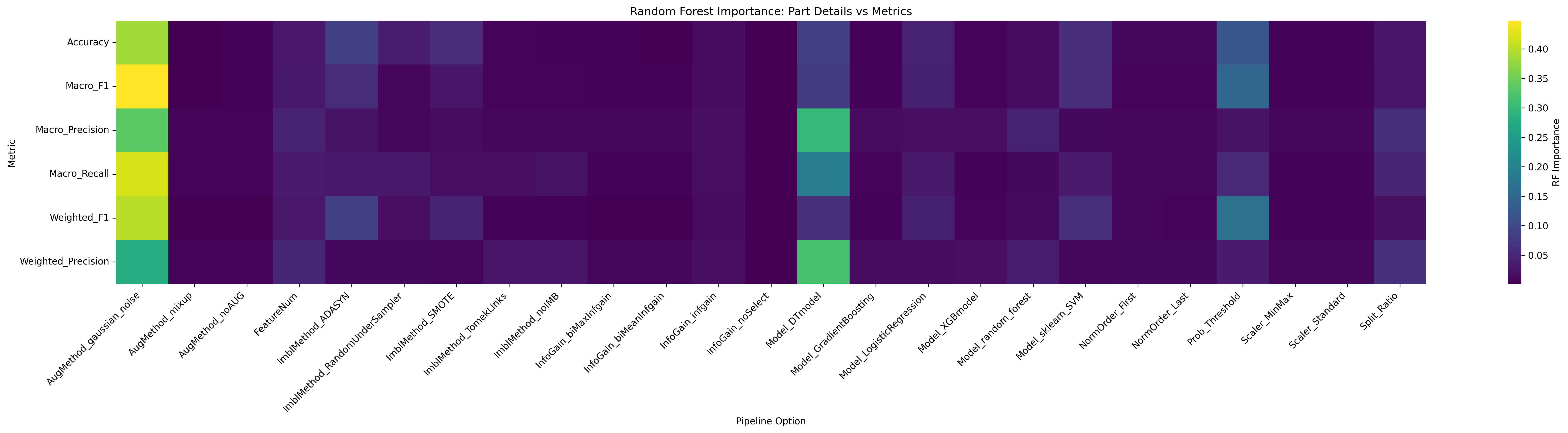}
\caption{Detailed Random Forest importance analysis for the Pima dataset.}
\label{fig:pima_rf_importance_detail}
\end{figure*}

\clearpage  
\subsection{Value-Level Component Similarity Analysis}    
\label{app:value_similarity}

This subsection evaluates \textbf{value-level component similarity} within the Pima dataset using Root Mean Square (RMS) distance, identifying configurations that yield \textbf{functionally similar predictive behavior}.

Tables~\ref{tab:pima_value_similarity_all} report pairwise similarities across different component categories for the Pima dataset, including:

Feature and Feature Selection components ,  Data Transformation components (e.g., preprocessing, augmentation) ,  and Model/Decision components (e.g., classifiers, thresholds)

These results highlight \textbf{redundant or interchangeable component values}, enabling pipeline simplification through pruning of low-impact configurations and supporting a more consistent, robust pipeline configuration.

Overall, this analysis provides a \textbf{fine-grained view of intra-component similarity, component equivalence, and interaction patterns} specific to the Pima dataset.

\begin{table*}[htbp]
\caption{Value-Level Similarity (RMS) across all pipeline components on Pima dataset}
\label{tab:pima_value_similarity_all}
\raggedright
\scriptsize

\centering
\footnotesize
\renewcommand{\arraystretch}{0.85}
\setlength{\tabcolsep}{2pt}
\textbf{(a) Feature and Feature Selection Components}

\begin{tabular}{lrrrrrrrrr}
\toprule
 & F:4 & F:5 & F:6 & F:7 & F:8 & FS:biMax & FS:biMean & FS:inf & FS:none \\
\midrule
F:4 &  & 0.0381 & 0.0397 & 0.0414 & 0.0405 \\
F:5 & 0.0381 &  & 0.0342 & 0.0364 & 0.0388 \\
F:6 & 0.0397 & 0.0342 &  & 0.0345 & 0.0373 \\
F:7 & 0.0414 & 0.0364 & 0.0345 &  & 0.0355 \\
F:8 & 0.0405 & 0.0388 & 0.0373 & 0.0355 &  \\

FS:biMax &  &  &  &  &  &  & 0.0252 & 0.0394 & 0.0405 \\
FS:biMean &  &  &  &  &  & 0.0252 &  & 0.0392 & 0.0411 \\
FS:inf &  &  &  &  &  & 0.0394 & 0.0392 &  & 0.0458 \\
FS:none &  &  &  &  &  & 0.0405 & 0.0411 & 0.0458 &  \\
\bottomrule
\end{tabular}

 \vspace{0.5em}

\textbf{(b) Data Transformation Components}

\begin{tabular}{lrrrrrrrrrrrr}
\toprule
 & Sc:minmax & Sc:std 
 & Aug:noise & Aug:mix & Aug:none 
 & Imb:ADA & Imb:RUS & Imb:SMOTE & Imb:Tomek & Imb:none \\
\midrule

Sc:minmax &  & 0.0438 \\
Sc:std & 0.0438 &  \\

Aug:noise &  &  &  & 0.1045 & 0.1028 \\
Aug:mix &  &  & 0.1045 &  & 0.0279 \\
Aug:none &  &  & 0.1028 & 0.0279 &  \\

Imb:ADA &  &  &  &  &  &  & 0.0435 & 0.0383 & 0.0648 & 0.0672 \\
Imb:RUS &  &  &  &  &  & 0.0435 &  & 0.0395 & 0.0567 & 0.0595 \\
Imb:SMOTE &  &  &  &  &  & 0.0383 & 0.0395 &  & 0.0548 & 0.0571 \\
Imb:Tomek &  &  &  &  &  & 0.0648 & 0.0567 & 0.0548 &  & 0.0325 \\
Imb:none &  &  &  &  &  & 0.0672 & 0.0595 & 0.0571 & 0.0325 &  \\

\bottomrule
\end{tabular}

\vspace{0.5em}

\textbf{(c) Model and Decision Components}

\begin{tabular}{lrrrrrrrrrrrr}
\toprule
 & M:DT & M:GB & M:LR & M:XGB & M:RF & M:SVM 
 & Norm:first & Norm:last 
 & Thr:.35 & Thr:.5 
 & Split:.1 & Split:.2 \\
\midrule

M:DT &  & 0.0799 & 0.0970 & 0.0792 & 0.0736 & 0.0992 \\
M:GB & 0.0799 &  & 0.0510 & 0.0319 & 0.0422 & 0.0510 \\
M:LR & 0.0970 & 0.0510 &  & 0.0523 & 0.0581 & 0.0384 \\
M:XGB & 0.0792 & 0.0319 & 0.0523 &  & 0.0419 & 0.0514 \\
M:RF & 0.0736 & 0.0422 & 0.0581 & 0.0419 &  & 0.0596 \\
M:SVM & 0.0992 & 0.0510 & 0.0384 & 0.0514 & 0.0596 &  \\

Norm:first &  &  &  &  &  &  &  & 0.0349 \\
Norm:last &  &  &  &  &  &  & 0.0349 &  \\

Thr:.35 &  &  &  &  &  &  &  &  &  & 0.0566 \\
Thr:.5 &  &  &  &  &  &  &  & 0.0566 &  \\

Split:.1 &  &  &  &  &  &  &  &  &  &  &  & 0.0472 \\
Split:.2 &  &  &  &  &  &  &  &  &  &  & 0.0472 &  \\

\bottomrule
\end{tabular}
\end{table*}

    \clearpage

\subsection{Cross-Component Interaction Analysis}
\label{app:interaction}

This subsection examines \textbf{interactions between pipeline components} by analyzing correlation patterns across branches and components in the evaluated pipelines. Given the large number of possible components, we focus on selected subsets to highlight the most informative interactions. Specifically, we analyze:

\begin{itemize}
    \item \textbf{Top-10 components by mean pairwise correlation}, representing consistently interacting components.
    \item \textbf{Top-10 components with the highest standard deviation of correlation}, representing components with highly variable interactions across configurations.
    \item \textbf{Clustered groups of components derived from the full correlation matrices}, indicating sets of components that tend to co-vary functionally.
\end{itemize}

\begin{figure*}[!t]
  \centering
  \includegraphics[width=\textwidth,height=0.95\textheight,keepaspectratio]{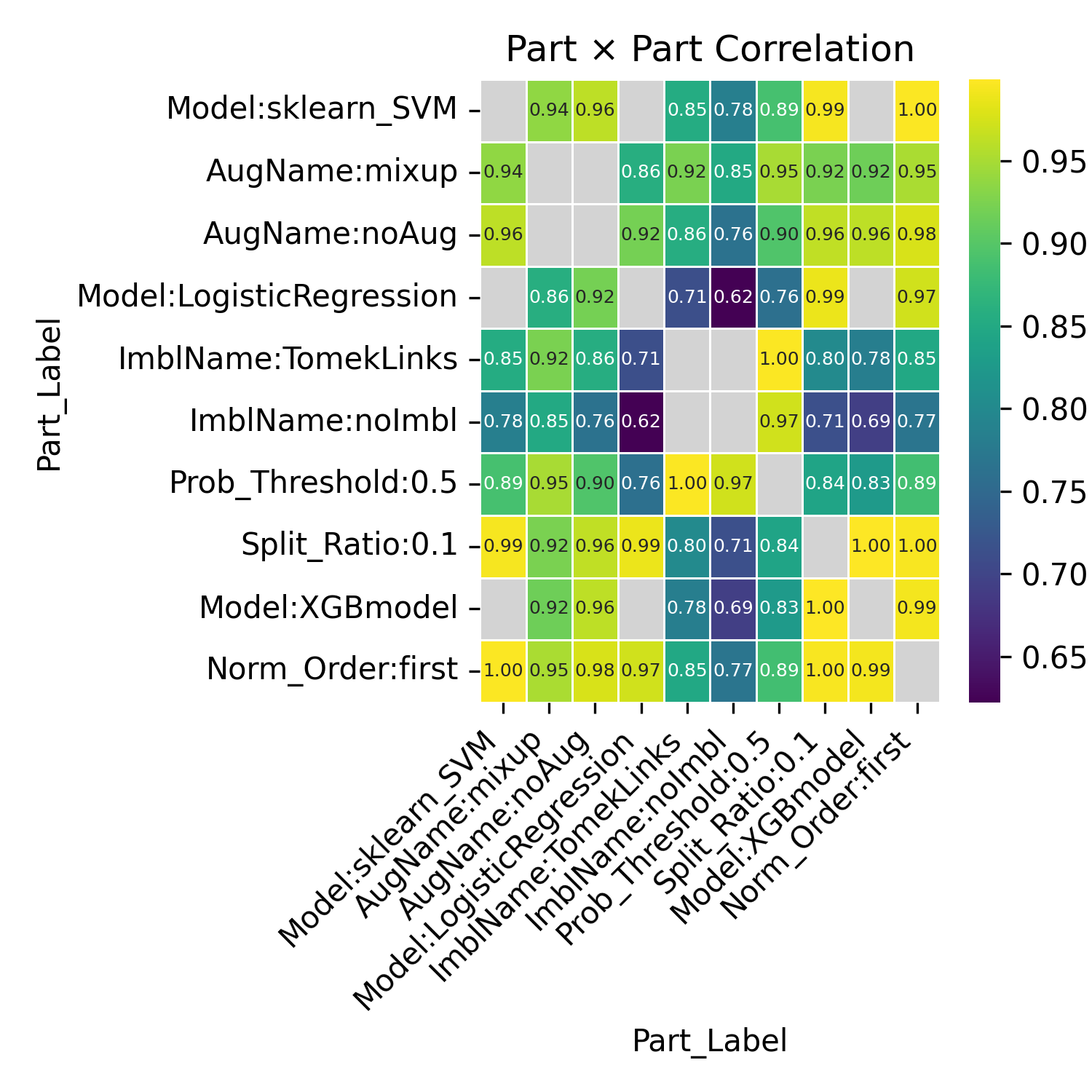}
  \caption{Pima dataset: Top-10   part-to-part correlation heat map.}
  \label{fig:pima_top10_part_correlation}
\end{figure*}

\begin{figure*}[!t]
  \centering
  \includegraphics[width=\textwidth,height=0.95\textheight,keepaspectratio ]{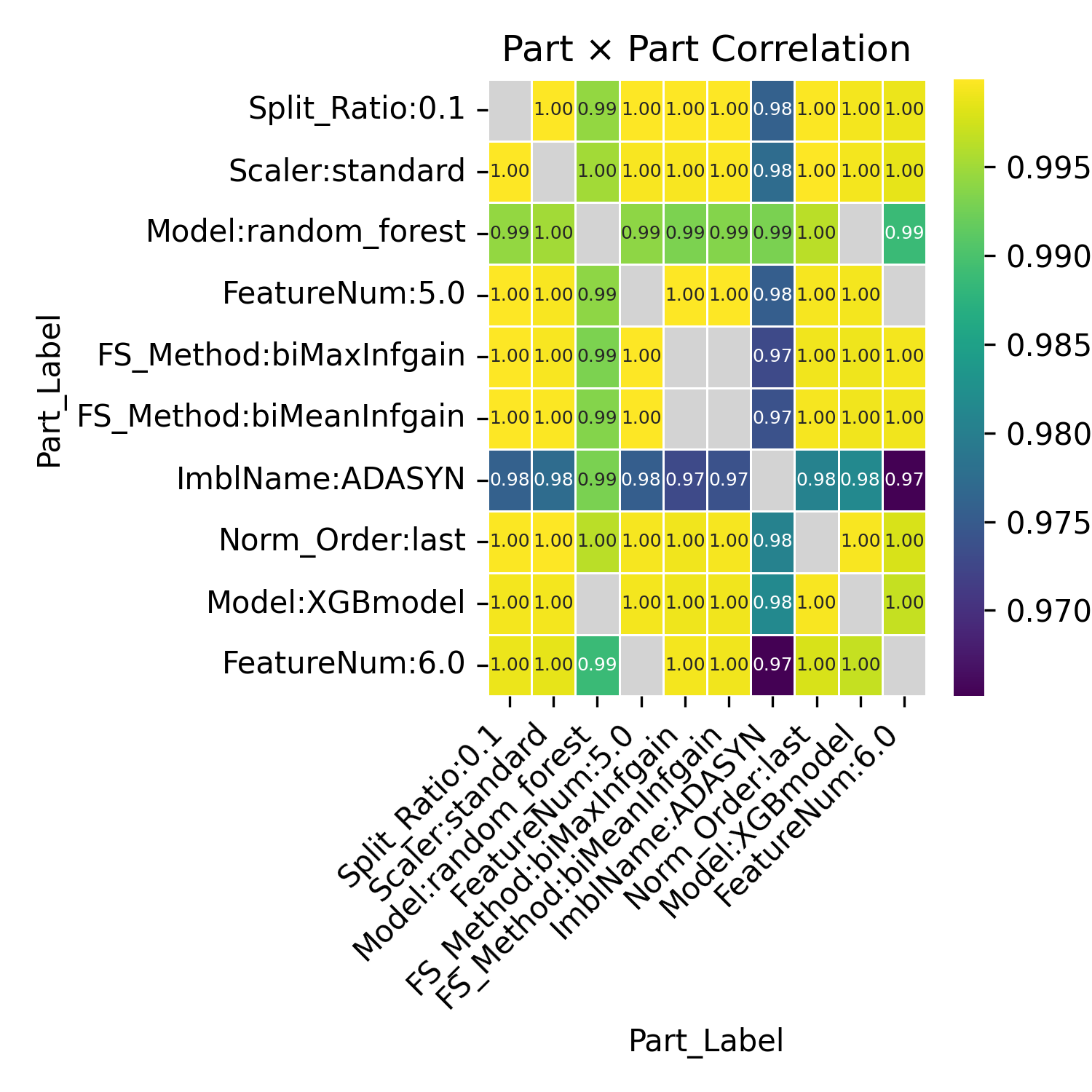}
  \caption{Pima dataset: Top-10 highest-standard-deviation part-to-part correlation heat map.}
  \label{fig:pima_top10_high_std_part_correlation}
\end{figure*}

\begin{figure*} [!t]
  \centering
 
    \includegraphics[width=\textwidth,height=0.95\textheight,keepaspectratio]{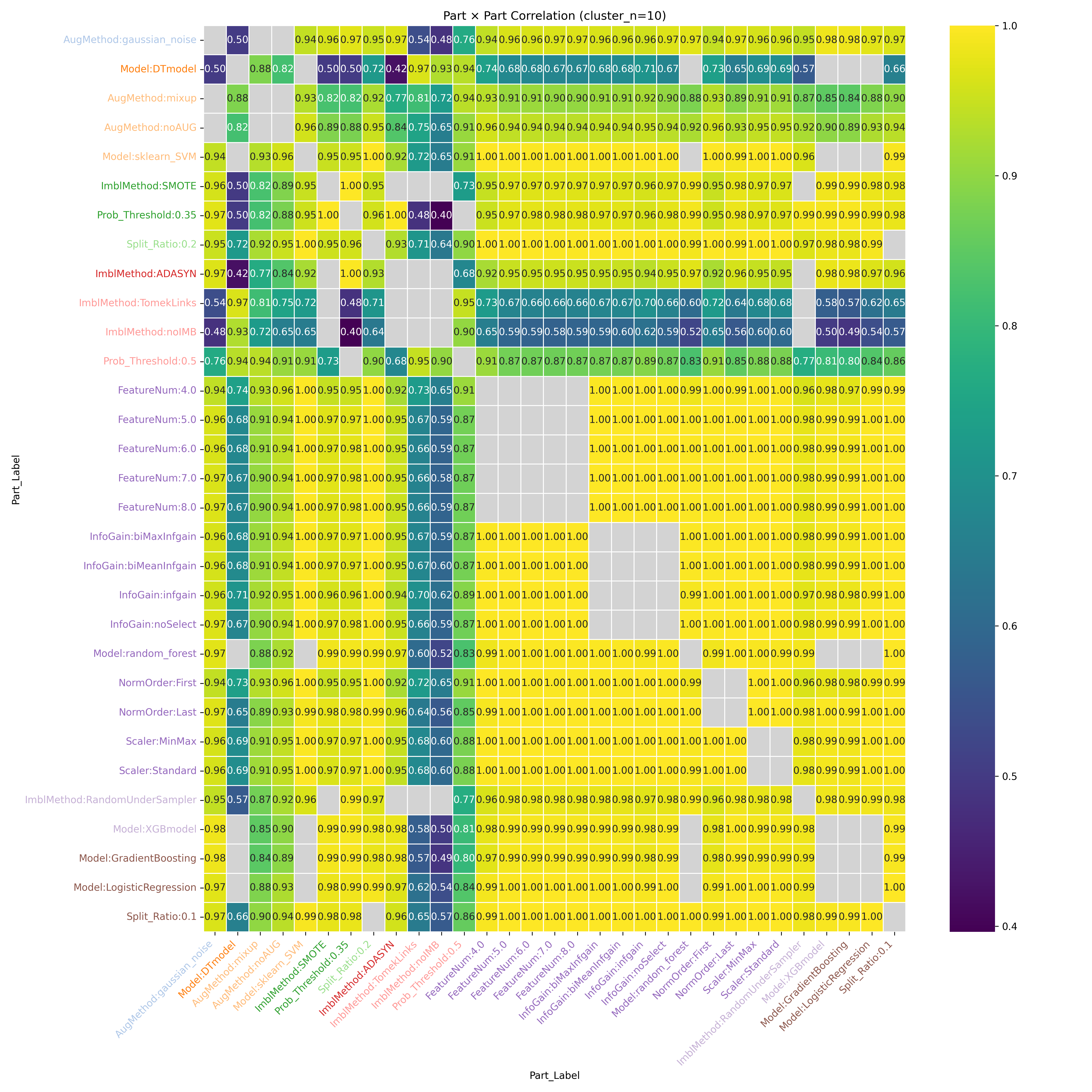}

    \caption{pima   cluster10  part vs part heat map}
              \label{fig:pimaclusteredPartxPartCorrelation}

\end{figure*}

For the Pima dataset, Figure~\ref{fig:pima_top10_part_correlation} shows general correlation patterns among key components, while Figure~\ref{fig:pima_top10_high_std_part_correlation} highlights the most variable interactions (highest standard deviation). Figure~\ref{fig:pimaclusteredPartxPartCorrelation} shows the clustered part-to-part correlations for the top-10 high-variance components in the Pima dataset.

\begin{figure*}[!t]
  \centering
  \includegraphics[width=\textwidth,height=0.95\textheight,keepaspectratio]{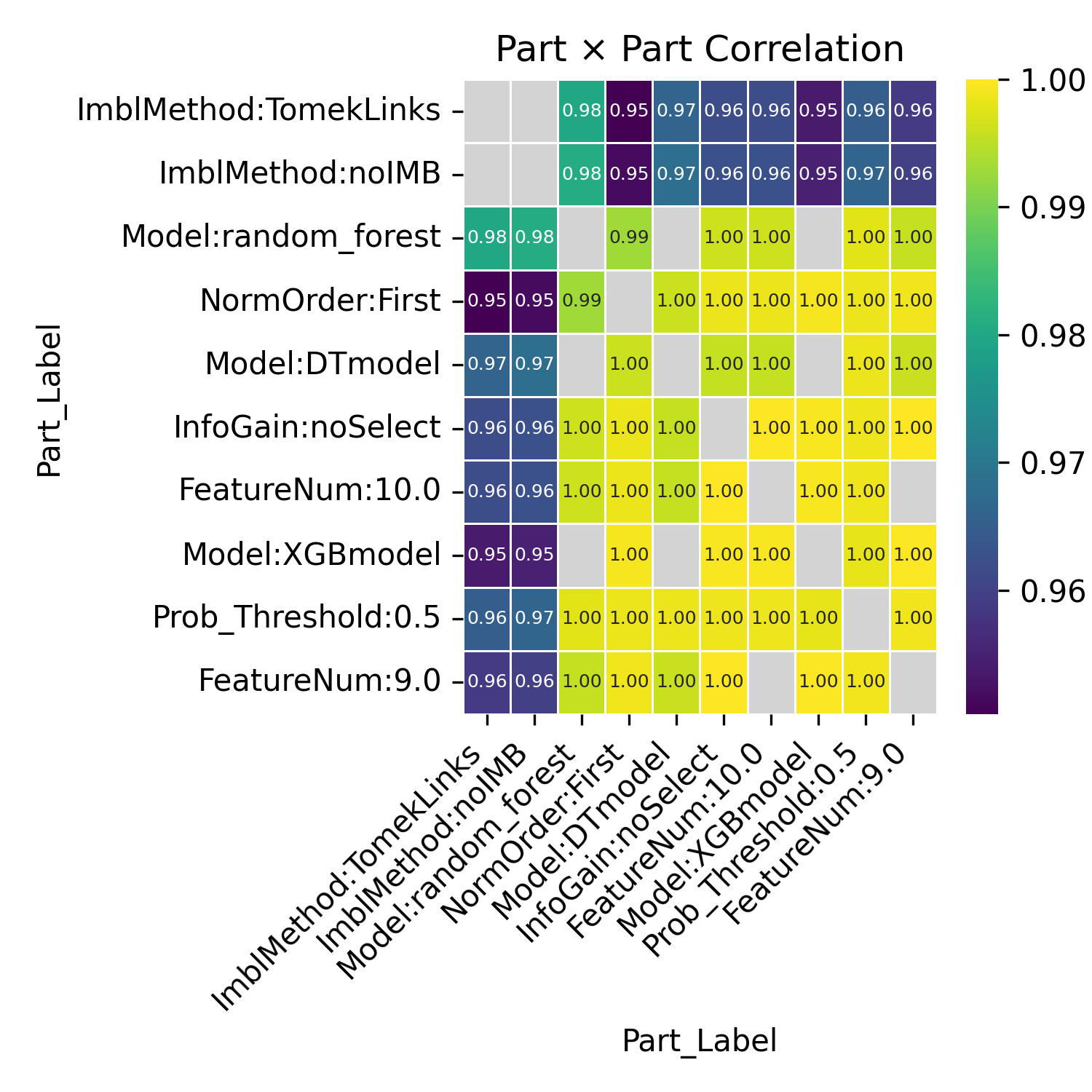}
  \caption{Stroke dataset: Top-10   part-to-part correlation heat map.}
  \label{fig:stroke_top10_part_correlation}
\end{figure*}

\begin{figure*}[!t]
  \centering
  \includegraphics[width=\textwidth,height=0.95\textheight,keepaspectratio]{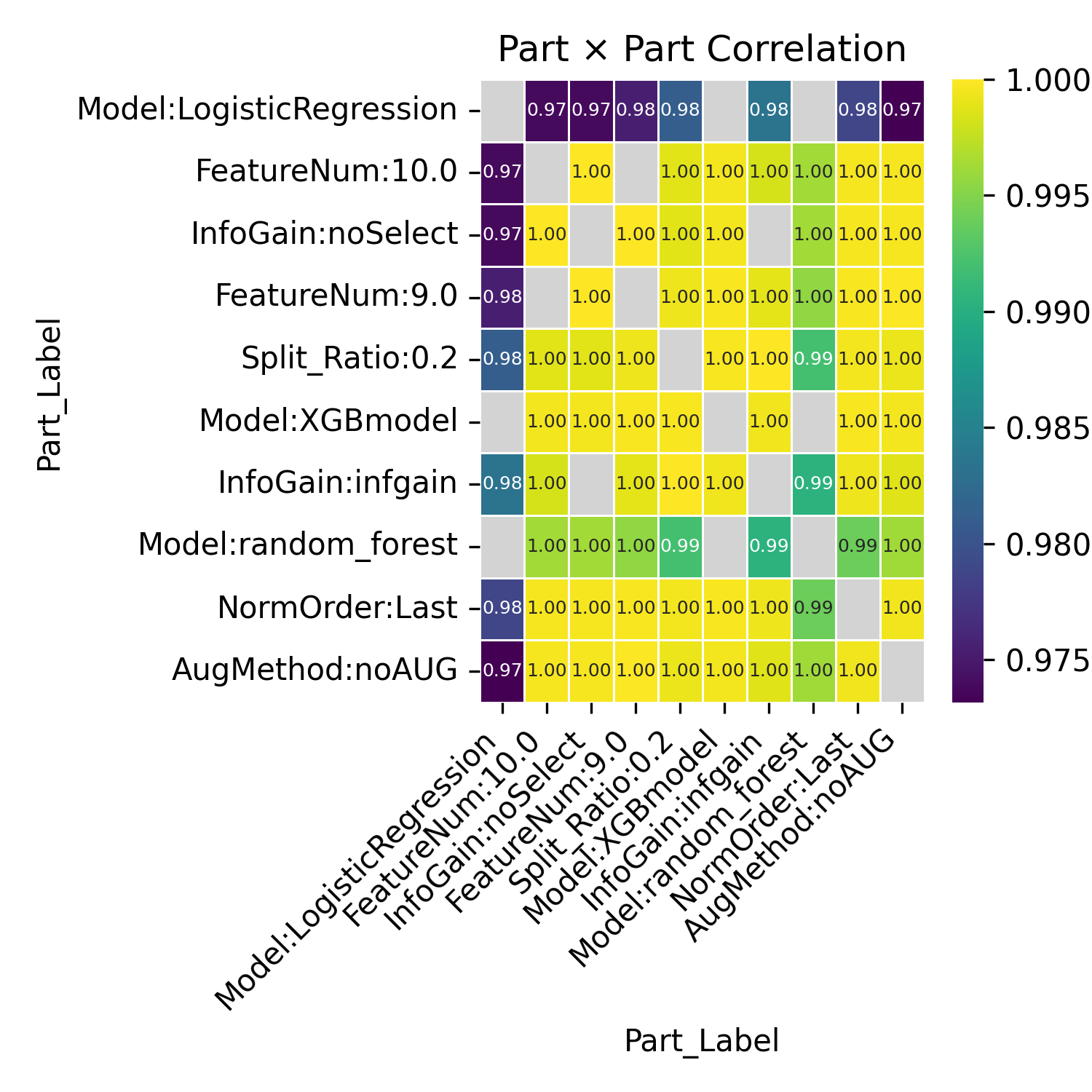}
  \caption{Stroke dataset: Top-10 highest-standard-deviation part-to-part correlation heat map.}
  \label{fig:stroke_top10_high_std_part_correlation}
\end{figure*}

\begin{figure*} [!t]
  \centering
 
    \includegraphics[width=\textwidth,height=0.95\textheight,keepaspectratio]{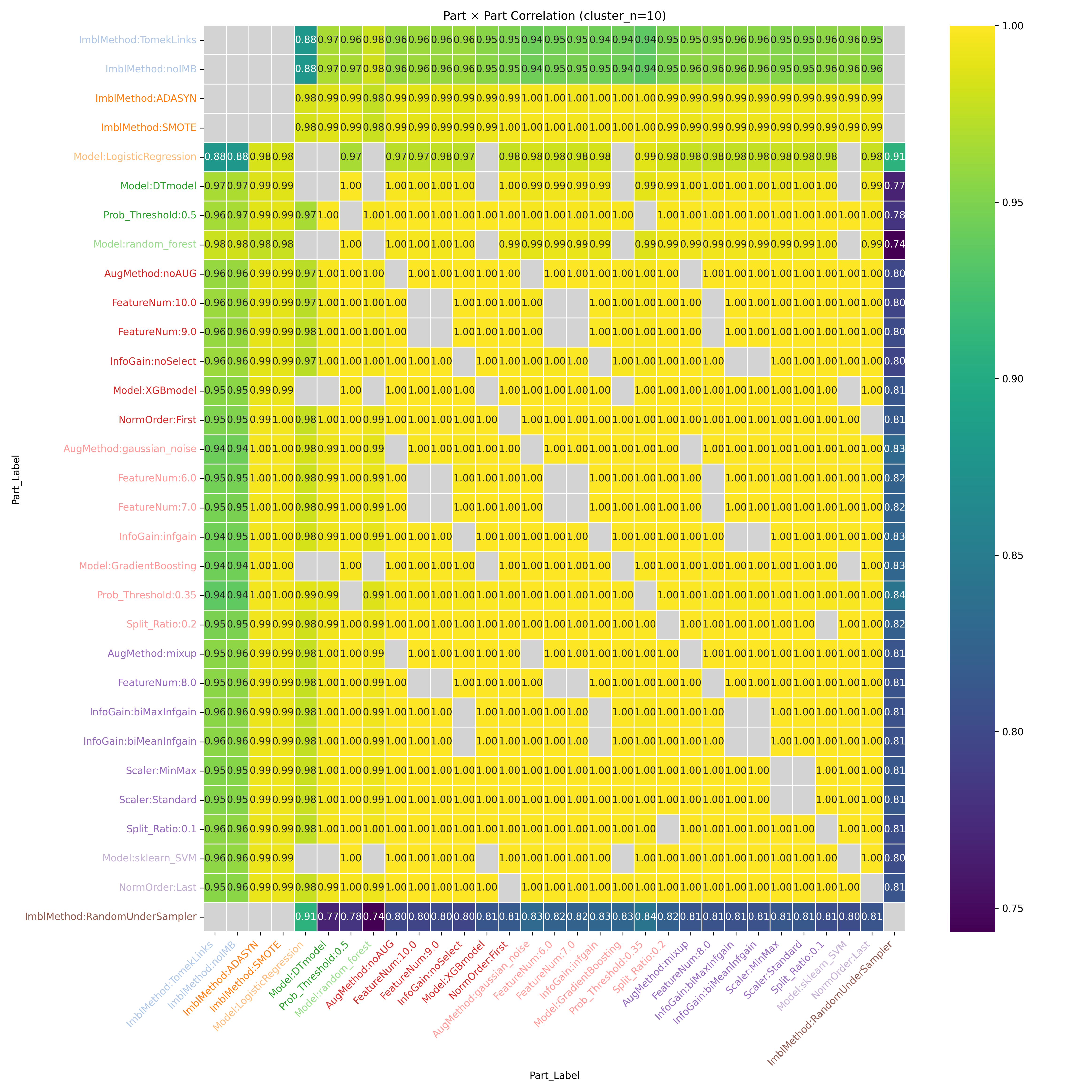}

    \caption{stroke  cluster part vs part heat map}
              \label{fig:strokeclusteredPartxPartCorrelation}

\end{figure*}

For the Stroke dataset, Figure~\ref{fig:stroke_top10_part_correlation} shows general correlation patterns among key components, while Figure~\ref{fig:stroke_top10_high_std_part_correlation} displays the top-10 highest-standard-deviation correlations. Figure~\ref{fig:strokeclusteredPartxPartCorrelation} shows the clustered part-to-part correlations for the top-10 high-variance components in the Stroke dataset.

These analyses reveal groups of preprocessing, augmentation, normalization, or modeling steps that tend to co-vary, providing insights into synergistic or redundant pipeline configurations. By focusing on the top or highly variable components, we reduce complexity while highlighting the most influential interactions. This allows readers to understand how combinations of components collectively impact predictive performance, guiding future pipeline optimization strategies.

   \clearpage

\subsection{Cross-Seed Robustness Analysis}
\label{app:cross_seed}

This subsection evaluates \textbf{cross-seed variability and model robustness} across multiple random initializations.

Table~\ref{tab:friedman_results} reports Friedman test results across evaluation metrics computed over random seeds.

\begin{figure*}[t]
\centering
\includegraphics[width=0.9\linewidth]{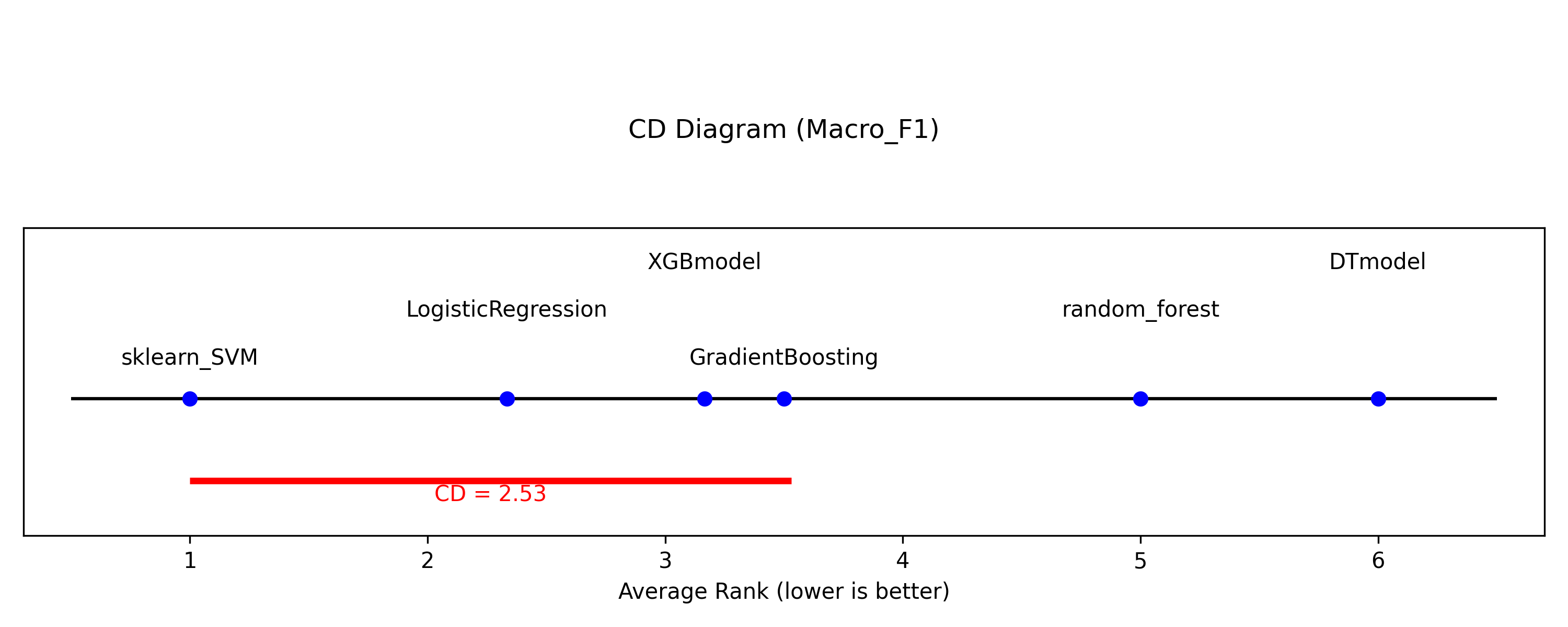}
 \caption{Critical Difference (CD) diagram for model ranks across random seeds based on Macro F1-score. Models connected by a horizontal bar are not significantly different according to the Friedman test and CD threshold.}

\label{fig:cd_macro_f1}
\end{figure*}

Figure~\ref{fig:cd_macro_f1} presents a \textbf{Critical Difference (CD) diagram} based on Macro F1-score, highlighting statistically significant differences between models under a class-imbalanced evaluation setting.

Figure~\ref{fig:pima_cross_seed_deviation} shows cross-seed mean performance and standard deviation across random seeds.

Overall, these results provide a concise view of \textbf{stability, robustness, and reproducibility}, supporting reliable pipeline selection.

\begin{table*}[t]
\centering
\caption{Friedman test results across evaluation metrics (computed over random seeds).}
\label{tab:friedman_results}
\begin{tabular}{lccc}
\toprule
Metric & $\chi^2$ & $p$-value & CD \\
\midrule
Accuracy            & 27.81 & $3.97\times10^{-5}$ & 2.53 \\
Macro Precision     & 27.90 & $3.80\times10^{-5}$ & 2.53 \\
Macro Recall        & 27.05 & $5.58\times10^{-5}$ & 2.53 \\
Macro F1            & 27.81 & $3.97\times10^{-5}$ & 2.53 \\
Weighted Precision  & 27.05 & $5.58\times10^{-5}$ & 2.53 \\
Weighted Recall     & 27.81 & $3.97\times10^{-5}$ & 2.53 \\
Weighted F1         & 27.81 & $3.97\times10^{-5}$ & 2.53 \\
Micro Precision     & 27.81 & $3.97\times10^{-5}$ & 2.53 \\
Micro Recall        & 27.81 & $3.97\times10^{-5}$ & 2.53 \\
Micro F1            & 27.81 & $3.97\times10^{-5}$ & 2.53 \\
Integrated Score    & 26.38 & $7.53\times10^{-5}$ & 2.53 \\
\bottomrule
\end{tabular}
\end{table*}

\begin{figure*}
\centering
\includegraphics[width=0.8\linewidth]{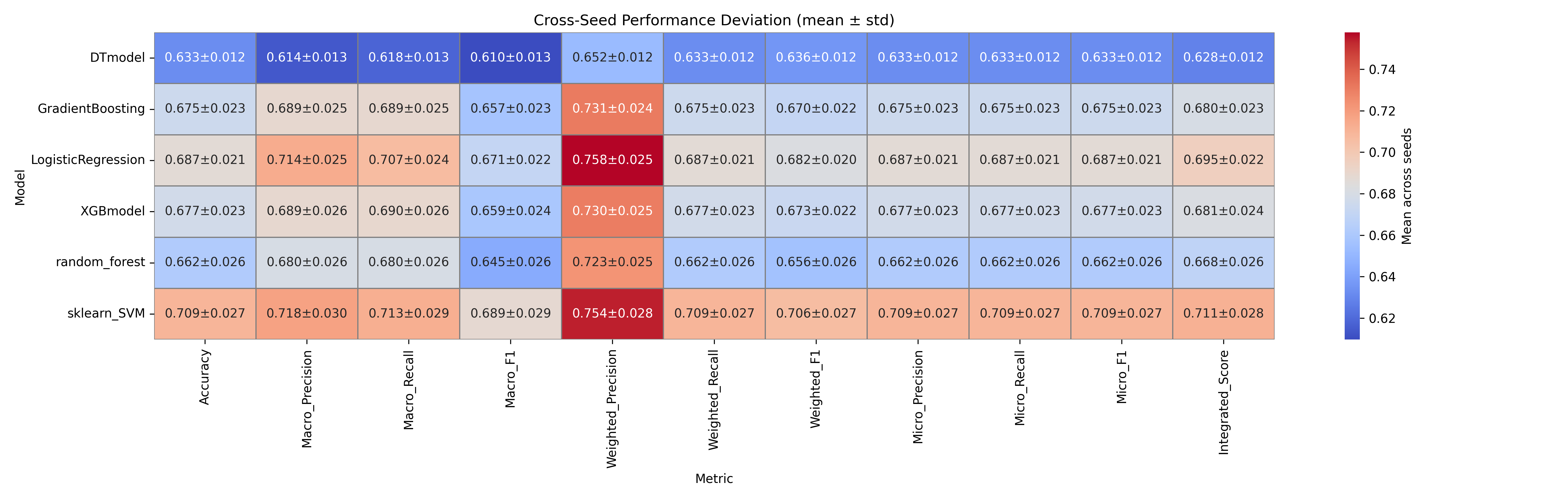}
 
\caption{Cross-seed mean performance and standard deviation across random seeds.}

\label{fig:pima_cross_seed_deviation}
\end{figure*}

\end{appendices}

\bibliography{huang-refs}
 

\end{document}